\documentclass[twoside,11pt]{article}

\usepackage{blindtext}

\usepackage{placeins}
\usepackage{amsmath, amsfonts}
\usepackage{subcaption}

\usepackage{algorithmic,algorithm}
\usepackage{xr}
\usepackage{mathtools}
\mathtoolsset{showonlyrefs}

\usepackage{enumitem}
\usepackage{adjustbox}
\usepackage{mathtools}
\usepackage{comment}

\usepackage[preprint]{jmlr2e}

\usepackage[utf8]{inputenc} 
\usepackage[T1]{fontenc}    
\usepackage{hyperref}       
\usepackage{url}            
\usepackage{booktabs}       
\usepackage{multirow}
\usepackage{amsfonts}       
\usepackage{nicefrac}       
\usepackage{microtype}      
\usepackage{xcolor}         

\graphicspath{{img/}}

\newcommand{\E}{\mathbb{E}}

\newcommand{\R}{\mathbb{R}}

\newcommand{\Y}{\mathcal{Y}}

\newcommand{\dobs}{y}
\newcommand{\ddz}{\mathbf{z}}
\newcommand{\ddtheta}{\boldsymbol{\theta}}
\newcommand{\ddsim}{\mathbf{x}}
\newcommand{\ddobs}{\mathbf{y}}

\newcommand{\Dobs}{Y}
\newcommand{\Ddsim}{\mathbf{X}}
\newcommand{\Ddobs}{\mathbf{Y}}
\newcommand{\Ddz}{\mathbf{Z}}

\newcommand{\Ddobsuptminus}{\mathbf{Y}_{t-k+1:t}}
\newcommand{\ddobsuptminus}{\mathbf{y}_{t-k+1:t}}
\newcommand{\Ddobsupt}{\mathbf{Y}_{1:t}}
\newcommand{\ddobsupt}{\mathbf{y}_{1:t}}

\newcommand{\ddobsupkplusl}{\mathbf{y}_{1:k+l-1}}
\newcommand{\DdobsupT}{\mathbf{Y}_{1:T}}

\newcommand{\Ddobsupkplusl}{\mathbf{Y}_{1:k+l-1}}
\newcommand{\DdobsupTfromk}{\mathbf{Y}_{k+l:T}}
\newcommand{\ddobsupT}{\mathbf{y}_{1:T}}
\newcommand{\ddobsupTfromk}{\mathbf{y}_{k+l:T}}

\newcommand{\sumjk}{\sum_{\substack{j,k=1\\k\neq j}}^m}

\newcommand{\Sv}{S_{\operatorname{v}}}
\newcommand{\SE}{S_{\operatorname{E}}}

\newcommand{\sumij}{\sum_{\substack{i,j=1\\i\neq j}}^m}
\newcommand{\probfor}{P^\phi(\cdot|\ddobs_{t-k+1:t})}

\DeclareMathOperator*{\argmin}{arg\,min}

\usepackage{lastpage}

\jmlrheading{25}{2024}{1-\pageref{LastPage}}{1/23; Revised
	12/23}{2/24}{23-0038}{Lorenzo Pacchiardi, Rilwan A. Adewoyin, Peter Dueben and Ritabrata Dutta}
\ShortHeadings{Probabilistic Forecasting with Generative Networks via Scoring Rule Minimization}{Pacchiardi, Adewoyin, Dueben and Dutta}

\firstpageno{1}

\begin{document}
	
	\title{Probabilistic Forecasting with Generative Networks\\via Scoring Rule Minimization}
	
	\author{\name Lorenzo Pacchiardi \email lorenzo.pacchiardi@gmail.com\\
		\addr Department of Statistics,
		University of Oxford \\
		Oxford, OX1 3LB \\
		United Kingdom
		\AND
		\name Rilwan A. Adewoyin\thanks{Also affiliated with Department of Computer Science and Engineering, Southern University of Science	and Technology, Shenzhen, China} \email rilwan.adewoyin@warwick.ac.uk \\
		\addr Department of Statistics,
		University of Warwick \\
		Coventry, CV4 7AL\\
		United Kingdom 
		\AND
		\name Peter Dueben \email peter.dueben@ecmwf.int \\
		\addr Earth System Modelling Section,
		European Centre for Medium-Range Weather Forecasts \\
		Reading, RG2 9AX\\
		United Kingdom 
		\AND
		\name Ritabrata Dutta \email ritabrata.dutta@warwick.ac.uk \\
		\addr Department of Statistics, 
		University of Warwick \\
		Coventry, CV4 7AL\\
		United Kingdom }

	\editor{Daniel Roy}
	
	\maketitle
	
	\begin{abstract}
		\looseness=-1
		Probabilistic forecasting relies on past observations to provide a probability distribution for a future outcome, which is often evaluated against the realization using a \textit{scoring rule}. Here, we perform probabilistic forecasting with generative neural networks, which parametrize distributions on high-dimensional spaces by transforming draws from a latent variable. Generative networks are typically trained in an \textit{adversarial} framework.
		In contrast, we propose to train generative networks to minimize a predictive-sequential (or \textit{prequential}) {scoring rule} on a recorded temporal sequence of the phenomenon of interest, which is appealing as it corresponds to the way forecasting systems are routinely evaluated.
		Adversarial-free minimization is possible for some scoring rules; hence, our framework avoids the cumbersome hyperparameter tuning and uncertainty underestimation due to unstable adversarial training, thus unlocking reliable use of generative networks in probabilistic forecasting.
		Further, we prove consistency of the minimizer of our objective with dependent data, while adversarial training assumes independence.
		We perform simulation studies on two chaotic dynamical models and a benchmark data set of global weather observations; for this last example, we define scoring rules for spatial data by drawing from the relevant literature. Our method outperforms state-of-the-art adversarial approaches, especially in probabilistic calibration, while requiring less hyperparameter tuning.
	\end{abstract}

	\section{Introduction}
	
	In many disciplines (for instance econometrics and meteorology), practitioners want to forecast the future state of a phenomenon. Providing prediction uncertainty (ideally by stating a full probability distribution) is often essential.
	This task is called \textit{probabilistic forecasting} \citep{gneiting2014probabilistic} and is commonplace in Numerical Weather Prediction (NWP, \citealp{palmer2012towards}), where physics-based models are run multiple times to obtain an ensemble of forecasts representing the possible evolution of the weather \citep{leutbecher2008ensemble}. To assess the performance of NWP systems, people commonly use Scoring Rules (SRs, \citealp{gneiting2007strictly}), functions quantifying the quality of a probabilistic forecast in relation to the observed outcome.

	\looseness=-1
	Here, we use \textit{generative (neural) networks} to provide probabilistic forecasts. In a generative network, a neural network maps a latent random variable to the required output space; hence, samples on the latter are obtained by transforming latent variable draws. As the density is inaccessible, the distribution is implicitly defined and specialized techniques are necessary to train generative networks.
	Among those, the popular Generative Adversarial Networks (GANs, \citealp{goodfellow2014generative, mirza2014conditional, nowozin2016f, arjovsky2017wasserstein}) framework trains a generative network by defining a min-max game against a competitor, termed \textit{critic}. However, adversarial training is unstable: it requires ad-hoc strategies \citep{gulrajani2017improved} and careful hyperparameter tuning \citep{salimans2016improved} but, even so, the trained generative network may not fully capture the data distribution, a phenomenon referred to as \textit{mode collapse} \citep{goodfellow2016nips,isola2017image, arora2017generalization,bellemare2017cramer, arora2017gans,richardson2018gans}. This prevents practitioners from reliably applying GANs to tasks where calibrated uncertainty quantification is paramount, such as probabilistic forecasting.
	Additionally, it is unclear how to extend the GAN training objective to the temporal data considered in probabilistic forecasting. Indeed, the adversarial framework is derived from divergences between probability distributions and considers data as independent and identically distributed samples from one of those distributions. 
	
	\looseness=-1
	Therefore, motivated by the use of scoring rules to evaluate traditional forecasting systems, we propose to train generative networks to minimize scoring rule values. Given a recorded temporal sequence of the phenomenon of interest, we use the generative network to forecast all steps of the sequence conditioned on the past. Then, our objective is the average over steps of the scoring rule between forecasts and realizations.
	In contrast to the adversarial framework, this so-called \textit{prequential} (predictive-sequential, \citealp{dawid1984present}) scoring rule captures the temporal structure of the data. Additionally, the minimizer of the prequential scoring rule enjoys consistency under mild conditions on the temporal sequence.
	Furthermore,
	our proposal allows adversarial-free training through a reparametrization trick \citep{kingma2013auto} for SRs defined as expectations over the generative distribution. Training with our objective is therefore drastically easier than with GAN, requires less hyperparameter tuning and easily avoids mode collapse.
	More in detail, our contributions are:
	\begin{itemize}
		\item We introduce a novel training objective for probabilistic forecasting based on a {prequential} scoring rule. 	
		\item Under stationarity and mixing conditions of the time series, we prove that the minimizer of the prequential scoring rule coincides asymptotically with that of the expected prequential scoring rule. Importantly, the latter corresponds to the true parameter value if the distribution induced by the generative network is well-specified.
		\item We leverage previous works in meteorology \citep{gneiting2007strictly, scheuerer2015variogram} and design training objectives for high-dimensional spatio-temporal data, enabling good performance with no need for a learnable data transformation.
		\item We test our method and state-of-the-art adversarial approaches on two chaotic models and a spatio-temporal weather data set. We find our method to be more stable and perform better, particularly in terms of uncertainty quantification of the forecast.
	\end{itemize}

	The rest of the paper is organized as follows. In Sec.~\ref{sec:background}, we discuss how the adversarial framework is obtained from a divergence minimization setup and overview the scoring rules training formulation for independent data, which was considered in previous works. In Sec.~\ref{sec:main}, which contains the main contributions of our work, we give our training objective for probabilistic forecasting, show its consistency and discuss SRs for spatial data. We discuss some related works in Sec.~\ref{sec:related_works} and show simulation results in Sec.~\ref{sec:simulations}. We conclude in Sec.~\ref{sec:conclusions}.

	\textit{Notation:} We use upper case $ X,Y$ and ${Z} $ to denote random variables, and their lower-case counterpart to denote observed values. Bold symbols denote vectors, and subscripts to bold symbols denote sample index (for instance, $ \ddobs_t $). Instead, subscripts to normal symbols denote component indices (for instance, $ y_i $ is the $ i $-th component of $ \ddobs $, and $ y_{t,j} $ is the $ j $-th component of $ \ddobs_t $). Finally, we use notation $\ddobs_{j:k} = (\ddobs_j, \ddobs_{j+1}, \ldots, \ddobs_{k-1}, \ddobs_k)$, for $j\le k$.

	\section{Background}\label{sec:background}

	\subsection{Generative networks via divergence minimization}\label{sec:GAN_via_div}
	
	A generative network represents a distribution $ P^\phi $ on a space $ \mathcal{Y} $ via a map $h_\phi: \mathcal {Z} \to \mathcal Y$ transforming samples from a probability distribution $ Q $ over the space $ \mathcal{Z}$; the map is parametrized by a Neural Network (NN) with weights $ \phi $. 
	Samples from $ P^\phi $ are obtained by generating $\mathbf{z}\sim Q $ and computing $ h_\phi(\mathbf{z}) \in \mathcal{Y}$; therefore, for any function $ g $ on $ \mathcal Y $, the expectation $ \E_{ \Ddobs \sim P^\phi} [g(\Ddobs)] $ can be computed by $ \E_{ \Ddz \sim Q} [g(h_\phi(\Ddz))] $. However, in general, the probability density of $ P^\phi $ cannot be evaluated.

	Assume now we observe data from a distribution $ P^\star $ on $ \mathcal Y $ and want to tune $ \phi $ so that $ P^\phi $ approximates $P^\star $. A divergence $ D(P^\star||P^\phi) $ is a function of two distributions such that $ D(P^\star||P^\phi)\ge0 $ and $D(P^\star||P^\phi) =0\iff P^\star=P^\phi$. Therefore, for a given $ D $, we can attempt solving
	\begin{equation}\label{Eq:div_min}
	\argmin_\phi D(P^\star||P^\phi).
	\end{equation}
	Various proposed approaches differ according to (i) their choice of divergence $ D $ and (ii) how they estimate the optimal solution in Eq.~\eqref{Eq:div_min} using samples from $ P^\star $ and $ P^\phi $. 
	A popular strategy is choosing $ D $ to be an $ f $-divergence (termed $ f $-GAN, \citealp{nowozin2016f}), in which case
	a variational lower bound can be obtained
	\begin{equation}\label{Eq:f-div-bound}
	D_f(P^\star||P^\phi)	\ge \sup_{c \in \mathcal{C}} \left(\mathbb{E}_{\Ddobs \sim P^\star}c(\Ddobs)-\mathbb{E}_{\Ddsim \sim P^\phi}f^{*}(c(\Ddsim))\right),
	\end{equation}
	where $ f^* $ is the Fenchel conjugate of the function $ f $ (see Appendix~\ref{app:f-GAN}) and $ \mathcal{C} $ is any set of functions
	from $ \mathcal{Y} $ to the domain of $ f^* $. By representing the set $ \mathcal{C} $ by a neural network $ c_\psi $ (termed \textit{critic} or \textit{discriminator}) with parameters $ \psi \in \Psi $, 
	an equivalent problem to Eq.~\eqref{Eq:div_min} when $ D $ is an \textit{f}-divergence is
	\begin{equation}\label{Eq:f-div-obj}
	\arg \min_\phi \max_\psi  \left(\mathbb{E}_{\Ddobs \sim P^\star}c_\psi(\Ddobs)-\mathbb{E}_{\Ddsim \sim P^\phi}f^{*}(c_\psi(\Ddsim))\right).
	\end{equation}
	The WGAN of \citet{arjovsky2017wasserstein}, which uses the 1-Wasserstein distance as $ D $, has a similar objective to Eq.~\eqref{Eq:f-div-obj}, differing mainly in taking $ \mathcal{C} $ to be the set of 1-Lipschitz functions.
	Details are given in Appendix~\ref{app:WGAN}.
	
	Typically, the problem in Eq.~\eqref{Eq:f-div-obj} is tackled by
	alternating gradient optimization steps over $ \psi $ and $ \phi $; the expectations are estimated via samples from both $ P^\star $ (i.e., a 	minibatch of observations) and from $ P^\phi $ (draws from the generative network). This approach is termed \textit{adversarial} as $ P^\phi $ and $ c_\psi$ respectively aim to minimize and maximize the same objective.

	Adversarial training of generative networks is however unstable and difficult.
	A well-known consequence of unstable adversarial training is mode collapse \citep{goodfellow2016nips,isola2017image, arora2017generalization,bellemare2017cramer, arora2017gans,richardson2018gans}, in which the generative distribution underestimates uncertainty and, in extreme cases, can collapse to a single point.
	Mode collapse has been related to the approximations involved in adversarial training: \cite{arora2017generalization} showed that mode collapse can arise due to finite capacity of the critic $ c_\psi $, while \cite{bellemare2017cramer} and \cite{binkowski2018demystifying} respectively linked it to using finite data and a finite number of steps in optimizing the $ c_\psi  $ network and subsequently using it to obtain gradient estimates for $ \phi $, which are thus biased.
	
	To avoid adversarial training altogether and bypass the above issues, Moment Matching Networks \citep{li2015generative, dziugaite2015training} are trained by considering $ D $ to be 
	the squared Maximum Mean Discrepancy (MMD) induced by a positive definite kernel $ k $
	\begin{equation}\label{Eq:MMD}
	D_k \left({P}^{\star}|| {P}^{\phi}\right)
	:= \E\left[k\left(\Ddsim, \Ddsim^{\prime}\right)-2 k(\Ddsim, \Ddobs)+k\left(\Ddobs, \Ddobs^{\prime}\right)\right], \quad  \Ddsim, \Ddsim^{\prime}\sim  P^\phi, \quad \Ddobs, \Ddobs^{\prime}\sim P^\star
	\end{equation}
	From Eq.~\eqref{Eq:MMD}, we can obtain an empirical unbiased estimate of $ D_k $ and its gradients without introducing a critic network. 
	However, using a fixed kernel on raw data can yield small discriminative power (as in the case of images, where numerical values have little meaning), leading to a poor fit of $ P^\phi $ to $P^\star$. Hence, \citet{li2017mmd} suggested applying a learnable transformation before computing the kernel, with parameters trained to maximize the MMD. This approach, termed MMD-GAN, again leads to an adversarial setting and to the issues mentioned above. Details in Appendix~\ref{app:MMDGAN}.

	\subsubsection{Conditional setting}
	To represent a conditional distribution $ P^\phi(\cdot|\ddtheta) $, for $ \ddtheta\in\Theta  $, 
	a map $h_\phi: \mathcal {Z} \times \Theta \to \mathcal Y$ can be used; similarly to above, samples from $ P^\phi(\cdot|\ddtheta) $ for fixed $ \ddtheta $ can be obtained via $ h_\phi(\mathbf{z};\ddtheta) $, $\mathbf{z}\sim Q $.
	In this way,
	$ f $-GAN, WGAN and MMD-GAN can all be easily extended to the setting in which we have data
	\begin{equation}\label{Eq:joint_data}
	(\ddtheta_i, \ddobs_i)_{i=1}^n\text{, where } \ddtheta_i\sim\Pi \text{ and } \ddobs_i \sim P^\star(\cdot|\ddtheta_i),
	\end{equation} 
	and want
	$ P^\phi(\cdot|\ddtheta) =P^\star(\cdot|\ddtheta) $ $ \Pi $-almost everywhere. For instance, the $ f$-GAN objective in Eq.~\eqref{Eq:f-div-obj} becomes
	\begin{equation*}\label{Eq:f-gan-cond}
	\min_\phi \max_\psi  \E_{ \ddtheta \sim \Pi}\big(\mathbb{E}_{\Ddobs \sim P^\star(\cdot|\ddtheta)}c_\psi(\Ddobs; \ddtheta)
	-\mathbb{E}_{\Ddobs \sim P^\phi(\cdot|\ddtheta)}f^{*}(c_\psi(\Ddobs;\ddtheta))\big),
	\end{equation*}
	where now $ c_\psi: \mathcal {Y} \times \Theta \to \operatorname{dom}_{f^*}$. More details can be found in Appendix~\ref{app:GAN}.

	\subsection{Generative networks via scoring rules minimization}\label{sec:scores}

	Here, we review scoring rules and a formulation for training generative networks 
	based on them which, for some choices, is intrinsically  adversarial-free. 

	\subsubsection{Scoring rules}

	A Scoring Rule (SR) $ S $ is a function of a distribution and an observation; see \cite{gneiting2007strictly, dawid2014theory} for an overview of their properties and usage. Generally, $ S(P^\phi, \ddobs) $ represents a \textit{penalty} assigned to the distribution $ P^\phi $ when $ \ddobs $ is observed. If $\ddobs$ is the realization of a random variable $ \Ddobs \sim P^\star $, the expected SR is
	$ S(P^\phi,P^\star) := \E_{\Ddobs \sim P^\star} S(P^\phi, \Ddobs).$
	$ S $ is said to be \textit{proper} relative to a set of distributions $ \mathcal{P}$ if the expected Scoring Rule is minimized in $ P^\phi $ when $ P^\phi=P^\star $ $$ S(P^\star, P^\star) \le S(P^\phi,P^\star) \ \forall \ P^\phi,P^\star \in \mathcal{P}.$$ Moreover, $ S $ is \textit{strictly proper} relative to $ \mathcal{P} $ if $ P^\phi = P^\star $ is the unique minimum. 
	In practice, assuming that $ \exists\ \phi^\star  : P^{\phi^\star} = P^\star $, $ P^\phi \neq P^\star $ can still minimize an expected proper SR $ S(P^\phi,P^\star) $, which in turn implies there may be multiple minima (still, the different minima can be thought of as more ``similar'' to $P^\star$ than other distributions, in some way); instead, if $ S $ is strictly proper, $ P^\star $ and $ P^\phi $ coincide if and only if $ P^\phi $ is the (unique) minimum of the expected SR. In case where $ \nexists \ \phi : P^\phi = P^\star$, then the expected proper SR  decreases as $ P^{\phi} $ becomes more similar to the data distribution $ P^\star $; however, nothing can be said on the number of minima without more information on $ \mathcal P $, even if $ S $ is strictly proper.
	
	
	For a strictly proper SR $ S $, the quantity $ D(P^\star||P^\phi) := S(P^\phi,P^\star) - S(P^\star,P^\star) $
	is a statistical divergence, as in fact $  D(P^\star||P^\phi)\ge 0 $ and
	$ D(P^\star||P^\phi) = 0 \iff P^\phi=P^\star $.
	
	A strictly proper SR which we will employ in the following is the \textit{Kernel Score} \citep{gneiting2007strictly}\begin{equation}\label{Eq:kernel_score}
	S_k(P^\phi, \ddobs) := \E[k(\Ddsim,\Ddsim')] - 2\cdot\E [k(\Ddsim, \ddobs)], \quad  \Ddsim,  \Ddsim' \sim P^\phi,
	\end{equation}
	where $ k $ is a positive-definite kernel. This choice is due to the expectation form of the kernel score, which, as explained in Sec.~\ref{sec:adv_free_training}, is required by our method. The kernel score is associated with the MMD in Eq.~\eqref{Eq:MMD}; see more details in Appendix~\ref{app:scores}.
	
	\subsubsection{Adversarial-free training of generative networks}\label{sec:adv_free_training}
	
	SRs have been previously used to train conditional generative networks in \cite{bouchacourt2016disco} and \cite{gritsenko2020spectral}, where the authors considered
	\begin{equation}\label{Eq:SR_obj_cond}
	\min_\phi \E_{\ddtheta\sim\Pi} \E_{\Ddobs\sim P^\star(\cdot|\ddtheta)} S(P^\phi(\cdot|\ddtheta),\Ddobs);
	\end{equation}
	for strictly proper $ S $, the solution is $ P^\phi(\cdot|\ddtheta) =P^\star(\cdot|\ddtheta) $ $ \Pi $-almost everywhere. With $ (\ddtheta_i, \ddobs_i)_{i=1}^n $ as in Eq.~\eqref{Eq:joint_data}, an unbiased estimate of the argument of $ \min_\phi $ in Eq.~\eqref{Eq:SR_obj_cond} is
	\begin{equation}\label{Eq:SR_obj_cond_emp}
	\frac{1}{n} \sum_{i=1}^n S(P^\phi(\cdot|\ddtheta_i),\ddobs_i).
	\end{equation}
	\looseness=-1
	Thus, to optimize Eq.~\eqref{Eq:SR_obj_cond} via Stochastic Gradient Descent (SGD), it is enough to obtain unbiased estimates of $\nabla_\phi S(P^\phi(\cdot|\ddtheta_i),\ddobs_i) $. That is possible whenever $ S $ is defined via a (possibly repeated) expectation over $ P^\phi $ (as for the kernel score), which can be estimated unbiasedly by generating samples $\ddsim_j\sim P_\phi, j=1,\ldots,m $, $ m>1 $ at each SGD step. Additionally, by recalling that samples $ \ddsim_j\sim P_\phi $ are obtained as $ \ddsim_j = h_\phi(\ddz), \ddz\sim Q $, automatic-differentiation libraries \citep{pytorch} can be exploited to compute gradients. 
	Hence, considering the kernel score as an example, at each SGD step, $ \phi $ will be updated by
	\begin{equation}\label{Eq:grad_update_k}
	\phi \leftarrow \phi - \gamma \cdot \frac{1}{|\mathcal B|} \sum_{i \in \mathcal{B}} \nabla_\phi  \left[\frac{1}{m(m-1)} \sum_{j\neq k} k(\ddsim_j, \ddsim_k) -  \frac{2}{m} \sum_{j} k(\ddsim_j, \ddobs_i)   \right],
	\end{equation}
	where $ \gamma  $ is the learning rate. More details are given in Appendix~\ref{app:unbiased_estimates}.
	This algorithm is equivalent to Moment Matching Networks (which use the objective in Eq.~\ref{Eq:MMD}).
	
	The Energy Score used in \cite{bouchacourt2016disco} and\cite{gritsenko2020spectral} can be obtained from $S_k $ by choosing
	$ k(\ddsim, \ddobs) =  - || \ddobs -\ddsim ||^\beta  $
	for $ \beta\in(0,2) $ \citep{gneiting2007strictly}. As such, the Energy Score also takes the expectation form necessary for our method and leads to a gradient descent update similar to Eq.~\eqref{Eq:grad_update_k}. See more details in Appendix~\ref{app:scores}.

	\section{Generative networks for spatio-temporal models via SR minimization}\label{sec:main}

	We will now extend the SR formulation to a training objective for probabilistic forecasting (Sec.~\ref{sec:preq_SR}) which is intuitive for temporal data and enjoys some consistency (Sec.~\ref{sec:consistency}). Later (Sec.~\ref{sec:SR_spatial}), we will discuss how to exploit the SR formulation to tackle high dimensional spatial data, by relying on a previously studied score from the probabilistic forecasting and meteorology literature \citep{scheuerer2015variogram} and by introducing \textit{patched} scores. 
	The resulting objectives can be minimized without resorting to adversarial training.

	\subsection{Time-series probabilistic forecasting via the prequential SR}\label{sec:preq_SR}

	Consider a discrete-time stochastic process $(\Ddobs_1, \Ddobs_2, \ldots, \Ddobs_t, \ldots ) = (\Ddobs_t)_t \sim P^{\star}$, where $\Ddobs_t\in \mathcal Y$; in general, $ \Ddobs_t $'s are \textit{not} independent. For a generic distribution $ P$ for $ (\Ddobs_t)_t $, we denote by $ P_t $ the marginal distribution for $ \Ddobs_t$, and by $ P_{r:s} $ the marginal distribution for $ \Ddobs_{r:s} $; the conditional distribution for $ \Ddobs_{t}|\ddobs_{u:v} $ will be denoted by 
	$ P_{t}(\cdot|\ddobs_{u:v}) $ and similar for $ \Ddobs_{r:s} $.

\begin{figure}
		\centering
		\includegraphics[width=0.5\columnwidth]{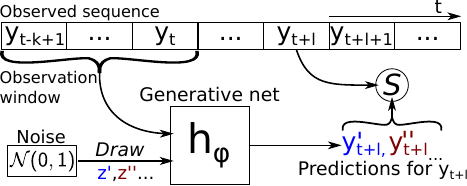}
		\caption{Estimation of the SR evaluating the forecast $ P^\phi_{t+l}(\cdot|\ddobs_{t-k+1:t})$ for the realization $\ddobs_{t+l} $. The prequential SR is obtained by repeating this procedure for all $ t $'s and summing the scores.}
		\label{fig:preq_SR}
	\end{figure}

	Having observed $ \ddobs_{1:t}$, we produce a \textit{probabilistic forecast} for $ \Ddobs_{t+l} $ for a given lead time $ l $ via a generative network conditioned on the last $ k $ observations,
	$ P^\phi_{t+l}(\cdot|\ddobs_{{t-k+1}:t}) $. We then repeat this procedure for all $ t $'s in a recorded window of length $ T $ and evaluate the forecast performance via $ S(P^\phi_{t+l}(\cdot|\ddobs_{t-k+1:t}),\ddobs_{t+l})$ for a SR $ S $ (Fig.~\ref{fig:preq_SR}); we then propose setting $ \phi $ to
	\begin{equation}\label{Eq:preq_SR}
	\hat \phi_T(\ddobsupT) :=\argmin_\phi  \sum\limits_{t=k}^{T-l}S(P^\phi_{t+l}(\cdot|\ddobs_{t-k+1:t}),\ddobs_{t+l}),
	\end{equation}
	which selects the value of $ \phi $ for which the average $ l $-steps ahead forecast in the training data is optimal according to $ S$. Operationally, Eq.~\eqref{Eq:preq_SR} can be tackled in the same way as Eq.~\eqref{Eq:SR_obj_cond_emp}, i.e., by simulating from $ P^\phi $ for each observation window $ \ddobs_{t-k+1:t} $ in a training batch, unbiasedly estimating the SR $ S $ and descending the gradient. 
	
	\looseness=-1
	The objective in Eq.~\eqref{Eq:preq_SR} evaluates sequential predictions obtained from the generative network; as such, we term it the \textit{prequential} (or \textit{predictive-sequential}) score \citep{dawid1984present, dawid2015bayesian}. This reflects what is usually done in evaluating traditional (physics-based) probabilistic forecasting systems \citep{leutbecher2008ensemble, gneiting2014probabilistic}.
	
	\subsubsection{Consistency of prequential SR minimization}\label{sec:consistency}

	Contrary to the independent-data setting of Eq.~\eqref{Eq:SR_obj_cond_emp}, Eq.~\eqref{Eq:preq_SR} cannot be seen as the empirical estimate of an expected SR. 
	Still, under some stationarity and mixing conditions of $ (\Ddobs_t)_t $, we prove below that the empirical minimizer $ \hat \phi_T(\DdobsupT) $ converges to the minimizer of the expected prequential SR. The reader uninterested in theoretical guarantees may skip this section, as it does not contain necessary information for understanding the remained of the paper.

	First, the objective in Eq.~\eqref{Eq:preq_SR} involves $ P^\phi_{t+l}(\cdot|\ddobs_{t-k+1:t}) $ for $ t\in\{k, k+1, \ldots, T-l-1, T-l\} $ and evaluates them against $ \ddobs_{k+l:T} $. In contrast, the initial part of the recorded sequence $ \ddobs_{1:k+l-1} $ only enters as conditioning values (indeed, the generative network cannot provide a forecast for the first $ k+l-1 $ elements of the sequence). Formally, we can define the joint distribution on $ \Ddobs_{k+l:T} $ induced by the generative network as $P^\phi_{k+l:T}(\cdot|\ddobs_{1:k+l-1}) $ and interpret the objective in Eq.~\eqref{Eq:preq_SR} as a SR evaluating $P^\phi_{k+l:T}(\cdot|\ddobs_{1:k+l-1}) $ against $ \ddobs_{k+l:T} $
	\begin{equation}\label{Eq:ST_def}
	S_T (P^\phi_{k+l:T}(\cdot|\ddobs_{1:k+l-1}), \ddobs_{k+l:T}):= \sum\limits_{t=k}^{T-l}S(P^\phi_{t+l}(\cdot|\ddobs_{t-k+1:t}),\ddobs_{t+l}).
	\end{equation}
	The above only makes sense as $ P^\phi_{t+l}(\cdot|\ddobs_{t-k+1:t}) $ can be obtained from $ P^\phi_{k+l:T}(\cdot|\ddobs_{1:k+l-1}) $, thanks to the marginal distribution for $ \Ddobs_{t+l} $ in $ P^\phi_{k+l:T}(\cdot|\ddobs_{1:k+l-1}) $ 
	being independent on $ \ddobs_{1:k+l-1} $ conditionally on $ \ddobs_{t-k+1:t} $. If that was not the case, $ \ddobs_{1:k+l-1} $ would also appear explicitly in the conditioning of $ P^\phi_{t+l} $. Indeed, $ P^\phi_{k+l:T}(\cdot|\ddobs_{1:k+l-1}) $ satisfies the following property (which generalizes the standard $ k $-Markov property):
	\begin{definition}
		A probability distribution $ P_{1:T} $ is $ k $-Markovian with lag $ l $ if, assuming it has density $ p_{1:T} $ with respect to some base measure, it can be decomposed as: 
		$ 	p_{1:T}(\ddobs_{1:T}) = p_{1:k+l-1}(\ddobs_{1:k+l-1})\prod_{t=k}^{T-l}p_{t+l}(\ddobs_{t+l}|\ddobs_{{t-k+1}:t}).
		$
	\end{definition}
	
	Therefore, $ S_T $ defined in Eq.~\eqref{Eq:ST_def} is a SR for distributions over $ \Ddobs_{k+l:T}| \ddobs_{1:k+l-1}$ which are $ k $-Markovian with lag $ l $.
	The following result (proved in Appendix~\ref{app:proper_preq_l}) establishes that $ S_T $ meaningfully evaluates $ P^\phi_{k+l:T}(\cdot|\ddobs_{1:k+l-1}) $ although it only employs $P^\phi_{t+l}(\cdot|\ddobs_{t-k+1:t}) $ explicitly:
	
	\begin{theorem}\label{Th:main_proper}
		If $ S $ is (strictly) proper, then $ S_T $ is (strictly) proper for distributions over $ \Ddobs_{k+l:T}| \ddobs_{1:k+l-1} $
		which are $ k $-Markovian with lag $ l $.
		%
	\end{theorem}

	Next, we introduce two quantities:
	\begin{equation*}\label{Eq:exp_preq_SR_min}
	\begin{aligned}
	\tilde \phi_T(\ddobs_{1:k+l-1}) &:= 
	\argmin_\phi \overbrace{\E_{\Ddobs_{k+l:T}| \ddobs_{1:k+l-1}} S_T(P^\phi_{k+l:T}(\cdot|\ddobs_{1:k+l-1}), \Ddobs_{k+l:T})}^{:=\tilde S_T(P^\phi_{k+l:T}(\cdot|\ddobs_{1:k+l-1}))},\\
	\phi^\star_T &:= \argmin_\phi \underbrace{\E S_T(P^\phi_{k+l:T}(\cdot|\Ddobs_{1:k+l-1}), \Ddobs_{k+l:T})}_{:= S_T^\star(P^\phi_{k+l:T})}.
	\end{aligned}
	\end{equation*}
	$ \tilde \phi_T(\ddobs_{1:k+l-1}) $ minimizes the expected prequential SR with respect to $ \Ddobs_{k+l:T}| \ddobs_{1:k+l-1} $, for which we introduced the short-hand notation $ \tilde S_T(P^\phi_{k+l:T}(\cdot|\ddobs_{1:k+l-1})) $; by Theorem~\ref{Th:main_proper}, if $ S $ is strictly proper and the distribution of $ \Ddobs_{k+l:T}| \ddobs_{1:k+l-1} $ is \textit{k}-Markovian with lag \textit{l}, $ \tilde \phi_T(\ddobs_{1:k+l-1})  $ parametrizes the true distribution.
	$ \phi^\star_T $ instead minimizes the expectation of $ S_T $ with respect to the full sequence $ \Ddobs_{1:T} $, which we shorten to $ S_T^\star(P^\phi_{k+l:T}) $. 
	
	Each term in the sum defining $ S_T $ depends on a finite number of observations; therefore, if $ (\Ddobs_t)_t $ satisfies some mixing and stationarity properties, we expect $  \tilde \phi_T(\ddobs_{1:k+l-1}) $ to not depend on $ \ddobs_{1:k+l-1} $ for large $ T $; similarly, we expect the empirical estimator $ \hat \phi_T(\ddobsupT) $ to converge to a fixed quantity. The following Theorem proves such consistency of $ \hat \phi_T(\ddobsupT) $ and $ \tilde \phi_T(\ddobs_{1:k+l-1}) $ to $ \phi^\star_T $.

	\begin{theorem}\label{Th:main_consistency}
		Let the following assumptions hold almost surely for $ \Ddobsupkplusl \sim P^\star_{1:k+l-1}$:
		\begin{enumerate}
			\item \label{ass:1} $ \Phi $ is compact.
			\item \label{ass:2} $ \phi^\star_T $ and $ 	\tilde\phi_T(\Ddobsupkplusl) $ are unique; additionally, there exist a metric $ d $ on $ \Phi $ such that, for all $ \epsilon>0 $,
			\begin{equation}\label{}
			\liminf_{T\to+\infty} \left\{ \min_{\phi: d(\phi, {\phi^\star_T})\ge\epsilon} S_T^\star(P_{k+l:T}^\phi) - S_T^\star(P_{k+l:T}^{\phi^\star_T})    \right\} >0 \qquad \text{ and }
			\end{equation}
			\begin{equation}\label{}
			\liminf_{T\to+\infty} \left\{ \min_{\phi: d(\phi, {	\tilde\phi_T(\Ddobsupkplusl)})\ge\epsilon} \tilde S_T(P^\phi_{k+l:T}(\cdot|\ddobs_{1:k+l-1})) - \tilde S_T(P^{\tilde\phi_T(\Ddobsupkplusl)}_{k+l:T}(\cdot|\Ddobs_{1:k+l-1}))    \right\} >0.
			\end{equation}
			\item \label{ass:3} (Asymptotic stationarity) Let $ G_t $ be the marginal distribution of $ \Ddobs_{t-k+1:t+l} $ and $\tilde G_t $ be the marginal distribution of $ \Ddobs_{t-k+1:t+l}|\Ddobsupkplusl $ for $ t\ge k $. Then, $ (T-l-k+1)^{-1} \sum_{t=k}^{T-l} G_t $ and $ (T-l-k+1)^{-1} \sum_{t=k}^{T-l} \tilde G_t $ both converge weakly to some probability measures on $ \mathcal{Y}^{k+l} $ as $ T \to\infty$.
			\item \label{ass:4} Both conditions below are satisfied:
			\begin{enumerate}
				\item (Mixing)\footnote{Roughly speaking, both mixing properties imply that $ \Ddobs_{t-m}$  and  $\Ddobs_{t} $ become independent as $ m\to\infty $.} Both $ (\Ddobs_t)_t \sim P^\star $ and $ (\Ddsim_t)_t \sim P^\star(\cdot|\Ddobsupkplusl)$ satisfy either one of these mixing properties (defined in Appendix~\ref{app:mixing}; $ (\Ddsim_t)_t $ and $ (\Ddobs_t)_t$ can satisfy different ones): 
				\begin{enumerate}
					\item $ \alpha $-mixing with mixing coefficient of size $ r/(2r-1) $, with $ r \ge 1 $, or
					\item $ \varphi $-mixing with mixing coefficient of size $ r/(r-1) $ with $ r > 1 $.
				\end{enumerate}
				\item (Moment boundedness) Define $ H(\ddobs_{t-k+1:t+l}) = \sup_{\phi \in \Phi } |S(P^\phi(\cdot|\ddobs_{t-k+1:t}), \ddobs_{t+l})| $; then, $$ \sup_{t\ge k} \E\left[H(\Ddobs_{t-k+1:t+l})^{r+\delta}\right]  \text{ and } \sup_{t\ge k} \E_{\Ddobs_{t-k+1:t+l}|\ddobsupkplusl}\left[H(\Ddobs_{t-k+1:t+l})^{r+\delta}\right] $$ are finite for some $ \delta>0 $, for the value of $ r $ corresponding to the condition above which is satisfied.
			\end{enumerate}
		\end{enumerate}
		Then, $ d(\phi^\star_T , \hat \phi_T(\Ddobs_{1:T})) \to 0$ and  $d(\tilde\phi_T(\Ddobsupkplusl) , \hat \phi_T(\Ddobs_{1:T})) \to 0 $ when $ T\to\infty $ almost surely with respect to $ (\Ddobs_t)_t \sim P^\star $. 
		It also follows that $ d(\tilde\phi_T(\Ddobsupkplusl) , \phi^\star_T) \to 0 $.
	\end{theorem}
	
	Under the assumptions of Theorem~\ref{Th:main_consistency}, with large enough $ T $, $ \hat{\phi}_T(\ddobsupT) $ and $ \tilde \phi_T(\ddobs_{1:k+l-1}) $ will be independent of the observed sequence $ \ddobsupT $ and will converge to $ \phi^\star_T $. Therefore, minimizing the prequential SR in Eq.~\eqref{Eq:preq_SR} asymptotically recovers the minimizer of an expected proper SR, which does not depend on the initial conditions of the sequence $ \ddobsupkplusl $.

	Proof of Theorem~\ref{Th:main_consistency} is given in Appendix~\ref{app:consistency}. The proof holds when $P^\phi_{t+l}$ depends on $ t $ only through the value of the past observations, which is our case of interest as we use the same generative network for all $ t $'s. The proof relies on the following steps: first, Assumptions~\ref{ass:1}, \ref{ass:2} and \ref{ass:4} are used to obtain a uniform law of large numbers using Theorem 2 in \cite{potscher1989uniform} (Appendix~\ref{app:ULLN}); then, this is combined with Assumption~\ref{ass:2} to obtain the results thanks to Theorem 5.1 in \cite{skouras1998optimal} (Appendix~\ref{app:generic_consistency}). As such, Theorem~\ref{Th:main_consistency} is a consequence of classical results in empirical process theory, adapted to our specific objective function in Eq.~\eqref{Eq:ST_def}. To make the intermediary results more easily usable and the proof easier to follow, Appendix~\ref{app:consistency} separately states and proves convergence of $ \hat \phi_T $ to $ \phi^\star_T $ (Appendix~\ref{app:consistency_statement_part_1}) and that of $ \hat \phi_T $ to $ \tilde \phi_T $ (Appendix~\ref{app:consistency_statement_part_2}), by splitting the assumptions in two sets.

	The assumptions in Theorem~\ref{Th:main_consistency} may be hard to verify. To make this easier, in Appendix~\ref{app:boundedness_SRs}, we show that Assumption~\ref{ass:2} is satisfied if $ S $ is strictly proper and $ P^\phi $ is a well-specified model with $ \phi $ being identifiable (Lemma~\ref{lemma:unique}). Moreover, we also provide simple sufficient conditions under which the moment boundedness condition in Assumption~\ref{ass:4} holds for the Energy and Kernel score (Lemmas~\ref{lemma:bounded_energy} and \ref{lemma:bounded_kernel}); the simplest of these conditions require the kernel or the space to be bounded for the Kernel score and the Energy score respectively.

	\subsection{Scoring rules for spatial data}\label{sec:SR_spatial}

	In contrast to multivariate data, spatial data is structured: the relation between different entries depends on their spatial distance. Computing, say, the Kernel SR in Eq.~\eqref{Eq:kernel_score} would discard this structure; we discuss here SRs which instead capture it, and which we will use for a spatio-temporal data set (Sec.~\ref{sec:weatherbench}).

	\subsubsection{Variogram Score}
	Say now $ \mathcal{Y}\subseteq \R^d $. For any $ p>0 $, the Variogram Score \cite{scheuerer2015variogram} is defined as
	\begin{equation}\label{Eq:var_score}
	\Sv^{(p)}(P^\phi, \ddobs):=\sum_{i, j=1}^{d} w_{i j}\left(\left|y_{i}-y_{j}\right|^{p}-\E_{\Ddsim\sim P^\phi}\left|X_{i}-X_{j}\right|^{p}\right)^{2}, \end{equation}
	where 
	$w_{ij}>0$ are fixed scalars. If $ \mathcal Y $ has a spatial structure $ w_{ij} $ can be set to be inversely proportional to the spatial distance of locations $ i $ and $ j $ \citep{scheuerer2015variogram}.
	However, $ \Sv^{(p)} $ is proper but not strictly so: it is invariant to change of sign and shift of all entries of $ \Ddsim $ by a constant, and only depends on the moments of $ P^\phi $ up to order $ 2p $ \citep{scheuerer2015variogram}. We will fix $p=1$ in the rest of our work.
	
	\subsubsection{Patched SR}
	
	\looseness=-1
	To convey the spatial structure of the data, we can compute a SR on
	a localized \textit{patch} of the data. In this way, the resulting score only considers the correlation between nearby components. 
	We can then shift the patch across the map and cumulate the resulting score (see Fig.~\ref{fig:patched} in Appendix). However, this SR is non-strictly proper as it does not evaluate long-range dependencies. A similar approach was suggested for an adversarial setting in \citet{isola2017image}, where the critic outputs separate numerical values for different patches of an input image.

	\subsubsection{Sum of SRs}
	
	Both SRs introduced above are non-strictly proper; 
	we can however obtain a strictly proper SR by adding a strictly proper SR to a proper one,
	as stated by the lemma below (proof in Appendix~\ref{app:proof_lemma}).
	\begin{lemma}\label{lemma:additive_SR}
		Consider two proper SRs $S_1$ and $S_2$, and let $ \alpha_1, \alpha_2 > 0 $; the quantity
		$$S_+(P,\ddobs) = \alpha_1 \cdot S_1(P,\ddobs) + \alpha_2 \cdot S_2(P,\ddobs)$$
		is a proper SR. If at least one of $S_1$ and $S_2$ is also strictly proper, then $S_+$ is strictly proper.\end{lemma}

	\subsubsection{Probabilistic forecasting for spatial data}
	
	Inserting the spatial SRs discussed above in the prequential score in Eq.~\eqref{Eq:preq_SR} enables probabilistic forecasting for spatial data using generative networks.
	For the Variogram Score, unbiased gradient estimates can be computed by simulating from $ P^\phi $; same holds for the patched SR if the underlying SR admits unbiased gradient estimates (Appendix~\ref{app:unbiased_estimates}).

	\section{Related works}\label{sec:related_works}
	
	Scoring rules have long been used in statistics: early characterisations are given in \cite{mccarthy1956measures} and \cite{savage1971elicitation}. Their usage for parameter estimation is also commonplace, see \cite{gneiting2007strictly} for an overview and \cite{dawid2016minimum} for theoretical properties.
	Closer to our method, \citet{dawid2013estimation} used SRs to infer parameters for spatial models, considering the conditional distribution in each location given all the others to be available; instead, \citet{dawid2015bayesian} considered model selection based on SRs and studied a prequential application. 
	
	Prior works proved theoretical results related to our consistency result in Sec.~\ref{sec:consistency}: Theorem~\ref{Th:main_consistency} combines Theorem 5.1 in \cite{skouras1998optimal}, which proves parameter consistency under uniform law of large numbers, with Theorem 2 in \cite{potscher1989uniform}, which is a classical result in empirical process theory obtaining a uniform law of large numbers for dependent data. \citet{skouras1998optimal} also discusses other properties of prequential losses for forecasting systems, such as our Eq.~\eqref{Eq:preq_SR}. Analogous results to our Theorem~\ref{Th:main_consistency} in similar settings were also shown: for instance, \cite{dziugaite2015training} showed consistency of the minimizer of an unbiased MMD empirical estimate to minimizer of the population MMD; they also rely on uniform convergence arguments, but, in contrast to our  Theorem~\ref{Th:main_consistency}, their result applied to i.i.d. data.
	
	As mentioned before, SR minimization for generative networks had been previously sparsely employed; however, a rigorous formulation such as the one we provide here was missing; moreover, no work specifically applied SR minimization to forecasting. Specifically, \cite{bouchacourt2016disco} used a formulation corresponding to SR minimization with the Energy Score, but obtained it using different arguments. Similarly to the latter, \cite{gritsenko2020spectral} trained a generative network via a generalized Energy Distance for a speech synthesis task, again considering independent samples. 
	More recent works use SR minimization for simulation-based Bayesian inference \citep{pacchiardi2022likelihood}, Neural SDEs \citep{issa2023non} and self-supervised representation learning \citep{vahidi2023probabilistic}.

	A research niche focuses on generating full time series with GANs \citep{brophy2021generative} often by using Recurrent NNs (RNN) as both discriminator and generator.
	In contrast, we focus on forecasting a single time step by conditioning on previous elements of the time-series. Some work aiming at generating full time-series can however be adapted for forecasting: for instance, the trained generator of \cite{yoon2019} can be conditioned on past data; still, our training method is more convenient if forecasting is the task at hand, as we do not require a temporal discriminator nor multiple independent time-series as training data.

	Some works instead directly used GANs for probabilistic forecasting, such as  \cite{kwon2019predicting, koochali2021, bihlo2021generative, ravuri2021skillful}. However, they considered the training samples as independent and did not study theoretically the consequence of using dependent data. \citet{bihlo2021generative} tested their method on a similar data set to ours (which we privileged as it is a standardized benchmark) and found the GAN to underestimate uncertainty, so they considered a GANs ensemble to mitigate uncertainty underestimation. Instead, \citet{ravuri2021skillful} exploited GANs for a precipitation {nowcasting} task (i.e., predicting for small lead time), achieving good deterministic and probabilistic performance. 
	\cite{rasul2021multivariate} instead performed probabilistic forecasting with a normalizing flow \citep{papamakarios2019normalizing}, by conditioning it on the output of a RNN or a Transformed network \citep{vaswani2017attention} to which the past elements of the time series were input. While their method is adversarial-free, the use of a normalizing flow reduces its flexibility, possibly inhibiting its capacity to efficiently represent spatial data, which is instead straightforward with generative networks (Sec.~\ref{sec:weatherbench}).

	Deterministic forecasting with NNs for the WeatherBench data set (Sec.~\ref{sec:weatherbench}) was studied extensively \cite{dueben2018challenges,scher2018toward, scher2019weather, weyn2019can}. Fewer studies tackled probabilistic forecasting: \citet{scher2021ensemble} combined deterministic NNs with ad-hoc strategies, not guaranteed to lead to the correct distribution. \citet{clare2021combining} binned instead the data, thus mapping the problem to that of estimating a categorical distribution.

	\section{Simulation study}\label{sec:simulations}

	We first study two low-dimensional time-series models which allow exhaustive hyperparameter tuning and architecture comparison but still present challenging dynamics due to their chaotic nature. We then move to a high-dimensional spatio-temporal meteorology data set.
	For all examples, we train generative models with the Energy and the Kernel Scores (Appendix~\ref{app:scores}) and their sum, termed \textit{Energy-Kernel} Score (a strictly proper SR due to Lemma~\ref{lemma:additive_SR}). As discussed in Sec.~\ref{sec:adv_free_training}, we choose these scores as they can be written via an expectation, which makes our method applicable. Other scores (such as, for instance, the log score, \citealp{gneiting2007strictly}), do not enjoy this property and are therefore unsuitable to our method. Additional SRs, discussed later in Sec.~\ref{sec:weatherbench}, are used for the meteorology example.
	For the Kernel Score, we use the Gaussian kernel (Appendix~\ref{app:scores}) with bandwidth $ \gamma $ tuned from the validation set (Appendix~\ref{app:gamma}). For all SR methods, we use 10 forecasts from the generator for each observation window to estimate SR values during training; however, performance does not degrade when using as few as 3 simulations (Appendix~\ref{app:n_outputs}), which lowers the computational cost (Appendix~\ref{app:weatherbench_comp_cost}).
	We compare with the original GAN \citep{goodfellow2014generative} and WGAN with gradient penalties (WGAN-GP, \citealp{gulrajani2017improved}). The latent variable $ \mathbf{Z} $ has independent components with standard normal distribution. To have a reference for the deterministic performance of the probabilistic methods, we compare them with deterministic networks trained to minimize the standard regression loss.

	All data sets consist of a long time series, which we split into training, validation and test set.
	We use the validation set for early stopping and hyperparameter tuning and report the final performance on the test set. The adversarial methods do not allow early stopping or hyperparameter selection using the training objective, as the generator loss depends on the critic state. For these methods, 
	therefore, we use other metrics to pick the best hyperparameters (see below).

	On the test set, we assess the calibration of the probabilistic forecasts by the \textit{calibration error} (the discrepancy between credible intervals in the forecast distribution and the actual frequencies). We also evaluate how close the means of the forecast distributions are to the observation by the \textit{Normalized Root Mean-Square Error} (NRMSE) and the \textit{coefficient of determination} R$^2  $; we detail all these metrics in Appendix~\ref{app:metric}. As all these metrics are for scalar variables, we compute their values independently for each component and report their average (standard deviation in Appendix~\ref{app:exp_res}).

	Our simulations show how the SR methods are easier to train and provide better uncertainty quantification. The adversarial methods require more hyperparameter tuning. We find the original GAN to be unstable and very poor at quantifying uncertainty due to mode collapse; WGAN-GP performs better but still has inferior performance than the SR approaches. Likely, ad-hoc adversarial training strategies could lead to better performance; however, the possibility of effortlessly training with off-the-shelf methods is an advantage of the SR approaches. Code for reproducing results is available \href{https://github.com/LoryPack/GenerativeNetworksScoringRulesProbabilisticForecasting}{here}.

	\begin{figure*}[t]
		\centering
		\begin{subfigure}[t]{0.5\textwidth}
			\begin{center}
				\includegraphics[width=\columnwidth]{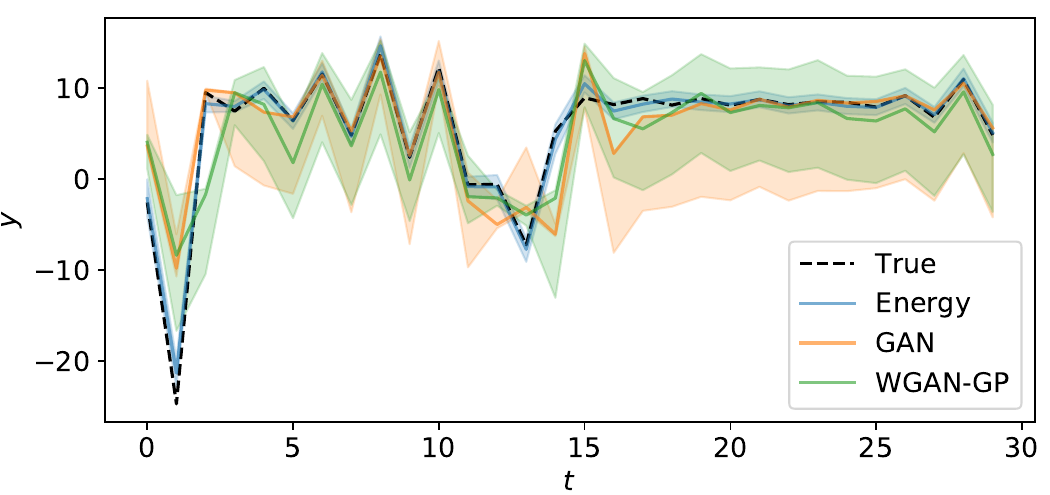}
			\end{center}
			\caption{Lorenz63}\label{fig:lorenz63}
		\end{subfigure}~
		\begin{subfigure}[t]{0.5\textwidth}
			\centering
			\includegraphics[width=\textwidth]{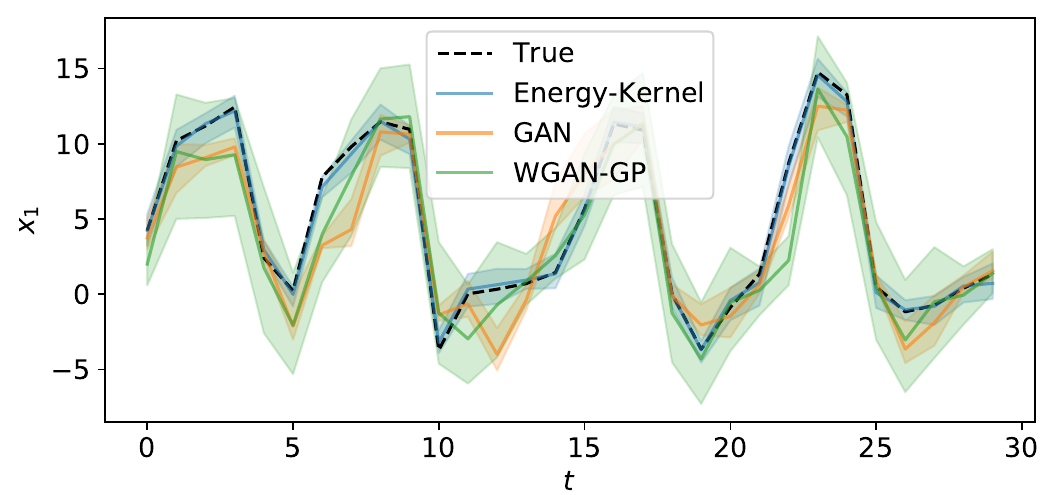}
			\caption{Lorenz96 (first data component).}\label{fig:lorenz96}
		\end{subfigure}											\caption{Results for selected methods for 
			Lorenz63 and Lorenz96 (first data component): median forecasts (solid line) and 99\% credible area (shaded area) for a part of the test set. For each $ t $, forecasts are obtained using the previous observation window. Credible regions for GAN and WGAN-GP are broader but contain the truth less frequently.}
		
		\label{fig:lorenz_both}
	\end{figure*}
	
	\subsection{Time-series models}\label{sec:lorenz63}\label{sec:lorenz96}
	
	We consider the Lorenz63 \citep{lorenz1963deterministic} and Lorenz96 \citep{lorenz1996predictability} chaotic models (Appendices~\ref{app:Lorenz63_model} and \ref{app:Lorenz_model}).  The former is defined on a 3-dimensional variable, a single component of which we assume to observe. The latter contains two sets of variables; we observe only one of them, which is 8-dimensional. In both cases, we generate an observed trajectory from a long model integration, from which we take the first 60\% as training set, the following $ 20\% $ as validation and the remaining $ 20\% $ as test set.

	We train the generative networks to forecast the next time step ($ l=1 $) from an observation window of size $ k=10 $. We use recurrent NNs based on Gated Recurrent Units (GRU, \citealp{cho2014properties}; Appendices~\ref{app:Lorenz63_nets} and \ref{app:Lorenz_nets}); we also tested fully connected networks but they had worse performance, so we do not report them here. For the SR methods, we select the best learning rate among 6 values according to the validation loss. For the adversarial methods, we consider instead 14 learning rates for both generator and critic; we also try two hidden dimensions for the GRU layers and four numbers of critic training steps for WGAN-GP; overall, we run 392 experiments for GAN and 1568 for WGAN-GP. As the validation loss is not a meaningful metric for adversarial approaches, we report results for 3 different configurations for GAN and WGAN-GP, maximizing either deterministic performance (1) or calibration (2), or striking the best balance between these two (3). More details are in Appendix~\ref{app:Lorenz63_hyperparameters} and \ref{app:Lorenz_hyperparameters}). These experiments are run on CPU machines and take at most a few minutes to complete.

	In Table~\ref{tab:res_lorez63_lorenz96}, we report performance metrics on the test set.
	The Kernel Score excels in deterministic forecasts, getting close to or outperforming the regression loss; however, all SR methods lead to combined great deterministic and probabilistic performance. On the other hand, adversarial methods are capable of good deterministic performance (1) or calibration (2) independently; but either of these two is at the expense of the other; the configuration with the best trade-off (3) is much worse than the SR methods (with WGAN-GP better than GAN).
	In Fig.~\ref{fig:lorenz_both}, we show observation and forecast for a part of the test set, for GAN and WGAN-GP in configuration (3), the Energy Score for Lorenz63 and the Energy-Kernel Score for Lorenz96.
	For the two SR methods, the median forecast is close to the observation and the credible region contains the true observation for most time steps. For GAN and WGAN-GP, the match with the observation is worse and credible regions generally contain the truth less frequently albeit being wider.
	Additional results are given in Appendices~\ref{app:res_lorenz63} and \ref{app:res_lorenz96}.

	\begin{table}[tb]
		\begin{center}
			\begin{tabular}{lcccccc}
				\toprule
				& \multicolumn{3}{c}{Lorenz63}  &  \multicolumn{3}{c} {Lorenz96} \\ 
				\cmidrule(r){2-4} \cmidrule(r){5-7}
				& Cal. error $ \downarrow $ & NRMSE $ \downarrow $ & R$ ^2 $ $ \uparrow $ & Cal. error $ \downarrow $ & NRMSE $ \downarrow $ & R$ ^2 $ $ \uparrow $ \\
				\midrule
				Regression & -  & 0.0079  &  0.9977  &  	-  & 0.0198  &  0.9905   \\
				\midrule
				Energy & {0.0380} &  0.0105  & 0.9960   & {0.0205}  & 0.0166  &  0.9933  \\
				Kernel & 0.0910 & \textbf{0.0083} & \textbf{0.9975} &    0.2196  & \textbf{0.0164}  &  \textbf{0.9935} \\
				Energy-Kernel & 0.1000 & {0.0114} & {0.9953} & \textbf{0.0104}  & 0.0173  &  0.9928   \\
				GAN (1) & 0.4830 & 0.0274 & 0.9729 & 0.4644  & 0.0354  &  0.9696  \\
				GAN (2) &0.0860  & 0.2425  &  -1.1166  & 0.2671  & 0.1500  &  0.4537 \\ 
				GAN (3) & 0.3590 & 0.0698 & 0.8245 &  0.3700 & 0.0763 & 0.8590  \\
				WGAN-GP (1) & 0.4710 & 0.0398 & 0.9429 & 0.4134  & 0.0330  &  0.9736  \\
				WGAN-GP (2) & \textbf{0.0270} & 0.1243 & 0.4440 & 0.0565  & 0.1081  &  0.7165  \\
				WGAN-GP (3) & 0.2100 & 0.0914 & 0.6996 & 0.1648 & 0.0786 & 0.8502  \\
				\bottomrule
			\end{tabular}
		\end{center}
		\caption{Performance on test set for the different methods, on the Lorenz63 and Lorenz96 models. Results with three hyperparameter configurations are reported for GAN and WGAN-GP, see text. Overall, SR methods perform well on both calibration and deterministic forecast metrics (NMRSE and R$ ^2 $), while adversarial approaches are incapable of doing so. }	\label{tab:res_lorez63_lorenz96}
	\end{table}

	\subsection{Meteorological data set}\label{sec:weatherbench}

	\begin{figure}[htb]
		\centering
		\includegraphics[width=0.6\textwidth]{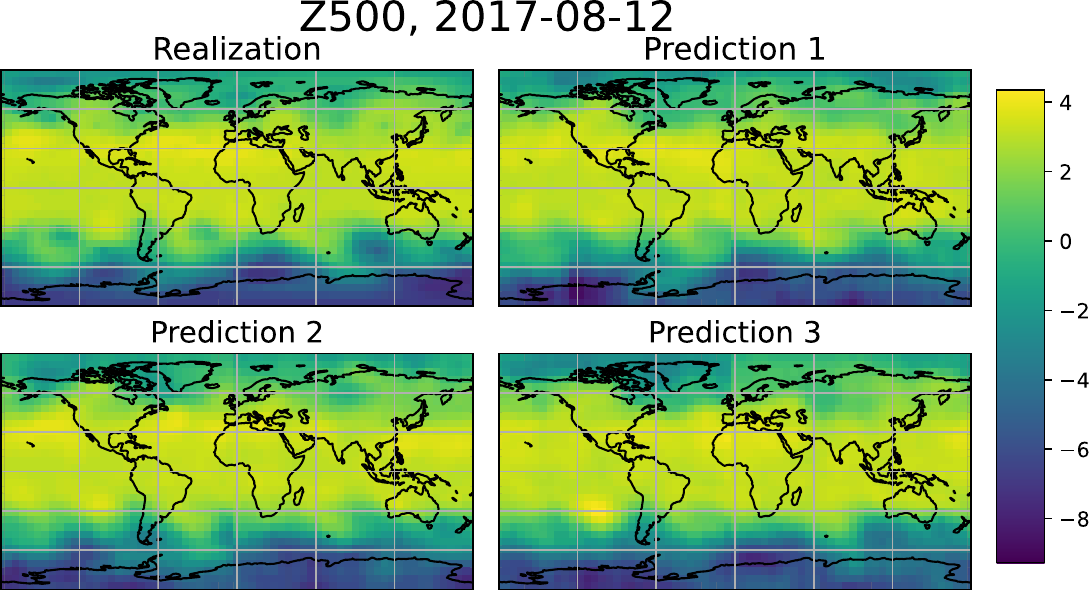}
		\caption{Realization and example of predictions obtained with the patched Energy Score (patch size 16) for a specific date in the test set for the WeatherBench data set. The predictions capture the main features but are slightly different from each other.}
		\label{fig:weatherbench}
	\end{figure}
	
	The WeatherBench data set\footnote{Released under MIT license, see \href{https://github.com/pangeo-data/WeatherBench/blob/master/LICENSE}{here}.} for data-driven weather forecasting \citep{rasp2020weatherbench}
	contains hourly values of several atmospheric fields from  1979 to 2018 at different resolutions;
	we choose here a resolution of $ 5.625^\circ $ over both longitude and latitude, corresponding to a 32$ \times $64 grid. We consider a single observation per day (12:00 UTC) and the 500 hPa geopotential (Z500) variable. 
	We forecast with a lead of 3 days ($ l=3 $) from a single observation ($ k=1 $). We use the years from 1979 to 2006 as training set, 2007 to 2016 as validation test and 2017 to 2018 as test set. 
	
	\looseness=-1
	In addition to the Energy, Kernel and Energy-Kernel Scores, we test the spatial SRs introduced in Sec~\ref{sec:SR_spatial}. 
	Specifically, we consider the Variogram Score with weights $ w $ inversely proportional to the distance on the globe (Appendix~\ref{app:weatherbench_Variogram}) and sum it to the Energy (\textit{Energy-Variogram}) or the Kernel (\textit{Kernel-Variogram}) Scores.
	We also consider the Patched Energy Score with patch sizes 8 and 16; to ensure the score is strictly proper, we add the overall Energy Score (summation weights in Appendix~\ref{app:weatherbench_weights}). We also consider patched regression loss.
	
	We employ a U-NET architecture \citep{ronneberger2015unet} for the generative network and a PatchGAN discriminator \citep{isola2017image} for the critic (Appendix~\ref{app:weatherbench_nets}). For the SR methods, we select the best learning rate among 6 values according to the validation loss; for the adversarial ones, we consider instead 7 values for both generator and critic, resulting in 49 experiments. We then pick the setups optimizing deterministic or calibration performance. For WGAN-GP, a single configuration optimizes both; for GAN, that did not happen. As for the time-series models, we report therefore results for setups maximizing either deterministic performance (1) or calibration (2), or striking the best balance between these two (3). All training is run on a single Tesla V100 GPU; computing times are reported in Appendix~\ref{app:weatherbench_comp_cost}. 
	
	Table~\ref{tab:res_weatherbench} reports performance on the test set. According to the calibration error, NRMSE and R$ ^2 $, the Patched Energy Scores perform best, with deterministic skill only slightly worse than the regression loss. 
	In Fig.~\ref{fig:weatherbench} we show observation and three different predictions obtained with the Patched Energy Score for a date in the test set. 
	More results in Appendix~\ref{app:weatherbench_res}.

	\begin{table}
		\begin{center}
			\begin{tabular}{lccc}
				\toprule
				& Cal. error $ \downarrow $ & NRMSE $ \downarrow $ & R$ ^2 $ $ \uparrow $ \\
				\midrule
				Regression & - & 0.1162  &  0.5300    \\
				Patched Regression, 8 & - & 0.1147  &  0.5459    \\
				Patched Regression, 16 & - & 0.1144  &  0.5509    \\
				\midrule
				Energy & 0.0863  & 0.1208   &  0.4968  \\
				Kernel & 0.0797  & 0.1200   &  0.5097  \\
				Energy-Kernel &0.0794  & 0.1194   &  0.5150   \\
				Energy-Variogram & 0.0899  & 0.1192   &  0.5177   \\
				Kernel-Variogram &0.1704  & 0.1203   &  0.5050   \\
				Patched Energy, 8 & \textbf{0.0550}  & 0.1189   &  0.5217  \\
				Patched Energy, 16 & 0.0690  & \textbf{0.1186}   &  \textbf{0.5248}   \\
				GAN (1) & 0.4845 & 0.1573 & 0.1418   \\
				GAN (2) & 	0.3130  & 0.2487   &  -2.7970  \\ 
				GAN (3) & 0.3625  & 0.1693   &  -0.0117   \\ 
				WGAN-GP & 0.1009  & 0.1302   &  0.4340   \\
				\bottomrule
			\end{tabular}
		\end{center}
		\caption{Performance on WeatherBench test set for different methods. Results with three hyperparameter configurations are reported for GAN, see text. SR methods perform well on both calibration and deterministic forecast metrics (NMRSE and R$ ^2 $). WGAN-GP is worse and GAN is drastically worse.}	\label{tab:res_weatherbench}
	\end{table}

	\section{Conclusions}\label{sec:conclusions}
	We proposed a method to train generative networks for probabilistic forecasting by minimizing a prequential scoring rule. 
	Compared to the standard adversarial framework, the advantages of the Scoring Rule formulation are: (i) it provides a principled objective for probabilistic forecasting; (ii) it yields adversarial-free training, with which better uncertainty quantification is possible, as we show empirically; (iii) it enables leveraging the literature on SRs to define objectives for spatio-temporal data sets. 
	The resulting training method is easier to use and requires less hyperparameter tuning than adversarial methods.
	
	We highlight the following limitations of our work: first, our Theorem~\ref{Th:main_consistency} relies on assumptions which are hard to verify, although, for some assumptions, we provide sufficient conditions applicable to the Kernel and Energy Scores in Appendix~\ref{app:boundedness_SRs}. However, we believe similar consistency properties hold provided the temporal process satisfies some generic stationarity and memory-less properties. Secondly, we do not experiment with forecasting multiple time-steps at once as we preferred focusing on single time-step forecast tasks for  analytical simplicity while developing our framework.
	Doing so would be a useful extension of our work; in practice, SRs assessing temporal coherence analogous to what is done with temporal discriminators in \cite{ravuri2021skillful} in the adversarial setting could be developed. Finally, we presented adversarial training and SR minimization as alternative approaches, but it is plausible that combining them would be beneficial. We leave this for future work.

	\newpage
	\acks{LP was supported by the EPSRC and MRC through the OxWaSP CDT programme\\(EP/L016710/1),	which also funded the computational resources used to perform this work. RD was funded by EPSRC (grant nos. EP/V025899/1, EP/T017112/1) and NERC (grant no. NE/T00973X/1). PD gratefully acknowledges funding from the Royal Society for his University Research Fellowship, as well as from the ESiWACE Horizon 2020 project (\#823988) and the MAELSTROM EuroHPC Joint Undertaking project (\#955513).
		We thank Geoff Nicholls, Christian Robert, Peter Watson, Matthew Chantry, Mihai Alexe and Eugenio Clerico for valuable feedback and suggestions.}
	
	
	
	\FloatBarrier
	\appendix

	\section{Proofs of theoretical results}

	\subsection{Proof of Lemma~\ref{lemma:additive_SR}}\label{app:proof_lemma}
	
	\begin{proof}
		By the definition of proper SR, we have that
		$$ \alpha_1\cdot S_1(Q, Q) \le  \alpha_1\cdot  S_1(P,Q) \ \forall \ P,Q \in \mathcal{P},$$
		and similar for $ S_2 $. By adding the two inequalities, we have therefore that
		\begin{equation}\label{}
		\alpha_1\cdot S_1(Q, Q) + \alpha_2\cdot S_2(Q, Q) \le  \alpha_1\cdot  S_1(P,Q) +\alpha_2\cdot  S_2(P,Q) \ \forall \ P,Q \in \mathcal{P},
		\end{equation}
		which implies that $ S_+ $ is a proper SR. 
		
		Assume now additionally that $ S_1 $, without loss of generality, is strictly proper, i.e.
		$$ \alpha_1\cdot S_1(Q, Q) <  \alpha_1\cdot  S_1(P,Q) \ \forall \ P,Q \in \mathcal{P}: P\neq Q;$$
		then, summing the above with the corresponding inequality for $ S_2 $ gives that
		\begin{equation}\label{}
		\alpha_1\cdot S_1(Q, Q) + \alpha_2\cdot S_2(Q, Q) <  \alpha_1\cdot  S_1(P,Q) +\alpha_2\cdot  S_2(P,Q) \ \forall \ P,Q \in \mathcal{P}: P\neq Q,
		\end{equation}
		which implies that $ S_+ $ is a strictly proper SR.\end{proof}

	\subsection{Propriety of the prequential SR}\label{app:proof_preqSR}
	
	In this Section, let $ P^\star $ denote the data generating distribution for $ (\Ddobs_1, \Ddobs_2, \ldots, \Ddobs_t, \ldots) = (\Ddobs_t)_t $, and let $ P $ denote a generic distribution assigned to $ (\Ddobs_t)_t $. From the distribution on the full sequence $ P $, conditional and marginals can be obtained, and denoted as follows: $ P_{t+1}(\cdot|\ddobs_{1:t}) $ denotes the conditional distribution for $ \Ddobs_{t+1} $ given $ \ddobsupt $, and $ P_{1:t} $ the (marginal) distribution for $ (\Ddobs_1, \Ddobs_2, \ldots, \Ddobs_t) $. Similar notation will be used for the conditional and marginals induced by $ P^\star $.

	\subsubsection{Generic 1-step ahead prequential SR}
	We first consider a simplified case in which we can access the marginal for $ \Ddobs_1 $ and all subsequent conditionals from $ P $.
	Given $ \ddobsupt $, we use the distribution $ P $ to construct a forecast distribution for $ \Ddobs_{t+1} $, namely $ P_{t+1}(\cdot|\ddobsupt) $; we penalize the forecast, against the verifying observation $ \ddobs_{t+1} $, via a SR $ S $
	\begin{equation}\label{Eq:app_SR}
	S(P_{t+1}(\cdot|\ddobsupt),\ddobs_{t+1}   ).
	\end{equation}
	From the above, we construct the \textit{prequential} SR for the forecast $ P_{1:T} $ as follows
	\begin{equation}\label{Eq:app_preq_SR}
	S_T(P_{1:T}, \ddobsupT) = \frac{1}{T} \left[\sum_{t=1}^{T-1}	S(P_{t+1}(\cdot|\ddobsupt),\ddobs_{t+1}) + S(P_{1},\ddobs_{1}) \right]; 
	\end{equation}
	the above assumes that at each time instant we obtain a probabilistic forecast $ P_{t+1}(\cdot|\ddobsupt) $ from the distribution $ P $ and we verify it against the next observed element of the sequence $ \ddobs_{t+1} $. Additionally, at the first time step, we have not yet received any observation, so our forecast $ P_1 $ is unconditional.
	Also, let us define the expected prequential score as
	\begin{equation}\label{Eq:app_preq_exp_SR}
	S_T(P_{1:T}, P_{1:T}^\star) := \E_{\DdobsupT\sim P^\star_{1:T}} S_T(P_{1:T}, \DdobsupT),
	\end{equation}

	\begin{theorem}\label{th:propriety_generic}
		If the scoring rule $ S $ is proper, then the prequential score $ S_T$ in Eq.~\eqref{Eq:app_preq_SR} is proper for distributions over $ \mathcal{Y}^T $, i.e.
		\begin{equation}\label{}
		S_T(P_{1:T}^\star, P_{1:T}^\star)\le 	S_T(P_{1:T}, P_{1:T}^\star).
		\end{equation}
		Similarly, if $ S $ is strictly proper, the prequential score $ S_T $ is strictly proper, i.e. the equality only holds if $ P_{1:T}=P_{1:T}^\star $.
	\end{theorem}
	\begin{proof}
		By definition of proper SR, we have that
		\begin{equation}\label{}
		\E_{\Ddobs_{t+1}\sim P_{t+1}^\star(\cdot|\ddobsupt)} S(P^\star_{t+1}(\cdot|\ddobsupt) ,\Ddobs_{t+1}) \le \E_{\Ddobs_{t+1}\sim P_{t+1}^\star(\cdot|\ddobsupt)} S(P_{t+1}(\cdot|\ddobsupt), \Ddobs_{t+1})
		\end{equation}
		for any conditional distribution $ P_{t+1}(\cdot|\ddobsupt) $ and for any values $ \ddobsupt $.
		
		Similarly, it holds
		\begin{equation}\label{Eq:preq_SR_proof1}
		\E_{\Ddobs_{1}\sim P_{1}^\star} S(P^\star_{1} ,\Ddobs_{1}) \le \E_{\Ddobs_{1}\sim P_{1}^\star} S(P_{1}, \Ddobs_{1}),
		\end{equation}
		for any distribution $ P_1 $.

		For the expected prequential SR, it holds that:
		\begin{equation}\label{key}
		\begin{aligned}
		S_T(P_{1:T}, P_{1:T}^\star) &= \E_{\DdobsupT\sim P^\star_{1:T}} S_T(P_{1:T}, \DdobsupT) \\
		&= \frac{1}{T} \left[\sum_{t=1}^{T-1} \E_{\DdobsupT\sim P^\star_{1:T}} S(P_{t+1}(\cdot|\Ddobsupt), \Ddobs_{t+1}) + \E_{\DdobsupT\sim P^\star_{1:T}} S(P_{1}, \Ddobs_{1}) \right] \\
		&=\frac{1}{T}\left[\sum_{t=1}^{T-1} \E_{\Ddobs_{1:t+1}\sim P^\star_{1:t+1}} S(P_{t+1}(\cdot|\Ddobsupt), \Ddobs_{t+1}) + \E_{\Ddobs_1\sim P^\star_{1}} S(P_{1}, \Ddobs_{1}) \right] ;
		\end{aligned}
		\end{equation}
		but now
		\begin{equation}\label{Eq:preq_SR_proof2}
		\begin{aligned}
		\E_{\Ddobs_{1:t+1}\sim P^\star_{1:t+1}} S(P_{t+1}(\cdot|\Ddobsupt), \Ddobs_{t+1}) &= \E_{\Ddobs_{1:t}\sim P^\star_{1:t}}\left[\E_{\Ddobs_{t+1}\sim P^\star_{t+1}(\cdot|\Ddobs_{1:t})}  S(P_{t+1}(\cdot|\Ddobsupt), \Ddobs_{t+1}) \right] \\
		&\ge \E_{\Ddobs_{1:t}\sim P^\star_{1:t}}\left[\E_{\Ddobs_{t+1}\sim P^\star_{t+1}(\cdot|\Ddobs_{1:t})}  S(P^\star_{t+1}(\cdot|\Ddobsupt), \Ddobs_{t+1}) \right],
		\end{aligned}
		\end{equation}
		so that
		\begin{equation}\label{Eq:preq_SR_proof}
		\begin{aligned}
		S_T(P_{1:T}, P_{1:T}^\star) &\ge \frac{1}{T}\left[\sum_{t=1}^{T-1} \E_{\Ddobs_{1:t+1}\sim P^\star_{1:t+1}} S(P_{t+1}^\star(\cdot|\Ddobsupt), \Ddobs_{t+1}) + \E_{\Ddobs_{1}\sim P_{1}^\star} S(P^\star_{1} ,\Ddobs_{1})\right]  \\
		&=\frac{1}{T} \left[\sum_{t=1}^{T-1}\E_{\Ddobs_{1:T}\sim P^\star_{1:T}} S(P_{t+1}^\star(\cdot|\Ddobsupt), \Ddobs_{t+1}) + \E_{\DdobsupT\sim P_{1:T}^\star} S(P^\star_{1} ,\Ddobs_{1})\right]  \\
		&= 	S_T(P^\star_{1:T}, P^\star_{1:T}),
		\end{aligned}
		\end{equation}
		which proves that $ S_T $ is proper. 
		
		To show that $ S_T $ is strictly proper if $ S $ is,
		we first notice that $ P_{1:T} $ is fully determined by the marginal $ P_1 $ and by the conditionals $ P_{t+1}(\cdot|\ddobs_{1:t}) $ for all possible values of $ \ddobs_{1:t} $, $ 1\le t \le T-1 $. In fact, if $ P_{1:T} $ and its conditional marginals have densities, you can write $$ p_{1:T}(\ddobs_{1:T}) = p_1(\ddobs_1)p_2(\ddobs_2|\ddobs_1)p_3(\ddobs_3|\ddobs_{1:2})\ldots p_{T-1}(\ddobs_{T-1}|\ddobs_{1:T-2})p_{T}(\ddobs_{T}|\ddobs_{1:T-1}).$$
		
		Next, notice that the $ \ge $ sign in Eq.~\eqref{Eq:preq_SR_proof} is an equality if and only if the $ \le $ sign in Eq.~\eqref{Eq:preq_SR_proof1} is an equality and the $ \ge $ sign in \eqref{Eq:preq_SR_proof2} is an equality for all $ 1\le t\le T $. As $ S $ is proper, the latter being true requires $$ \E_{\Ddobs_{t+1}\sim P^\star_{t+1}(\cdot|\ddobs_{1:t})}  S(P_{t+1}(\cdot|\ddobsupt), \Ddobs_{t+1}) = \E_{\Ddobs_{t+1}\sim P^\star_{t+1}(\cdot|\ddobs_{1:t})}  S(P^\star_{t+1}(\cdot|\ddobsupt), \Ddobs_{t+1})$$ for all values of $ \ddobs_{1:t} $ in the support of $ P^\star_{1:t} $. If $ S $ is strictly proper, however, the above conditions require that $ P_1=P_1^\star $ and $ P_{t+1}(\cdot|\ddobsupt)=P^\star_{t+1}(\cdot|\ddobsupt) \ \forall\ \ddobsupt $ in the support of $ P^\star_{1:t} $ and for $  1\le t\le T-1  $, which implies that $ P_{1:T} = P^\star_{1:T} $ due to distributions on $ \Ddobs_{1:T} $ being determined by the marginal for $ \Ddobs_1 $ and the conditional on $ \Ddobs_{t+1}|\ddobs_{1:t} $ for all values of $ \ddobs_{1:t} $ in the support of $ P^\star_{1:t} $. \end{proof}

	\subsubsection{$ l $-steps ahead prequential SR (Theorem~\ref{Th:main_proper})}\label{app:proper_preq_l}
	We now go back to the specific setting considered in the main body of the paper. 
	By discarding the model parameter $ \phi $ in the notation for simplicity, the generative network induces conditional distributions $ P_{t+l}(\cdot|\ddobs_{1:t}) $ for $ \Ddobs_{t+l} $ which only depend on the last $ k $ observations, i.e. $ P_{t+l}(\cdot|\ddobsupt) =P_{t+l}(\cdot|\ddobsuptminus) $. Therefore, the joint distribution for $ \Ddobs_{k+l:T} $ induced by the generative network satisfies the following property:
	\begin{definition}\label{def:k-markov_lag_l}
		A probability distribution $ P_{1:T} $ is $ k $-Markovian with lag $ l $ if it can be decomposed as follows, assuming it has density $ p_{1:T} $ with respect to some base measure: 
		\begin{equation}\label{Eq:kMarkov-lag-l}
		p_{1:T}(\ddobs_{1:T}) = p_{1:k+l-1}(\ddobs_{1:k+l-1})\prod_{t=k}^{T-l}p_{t+l}(\ddobs_{t+l}|\ddobs_{{t-k+1}:t}).
		\end{equation}
	\end{definition}
	Setting $ l=1 $ recovers the standard definition of $ k $-Markovian models.	
	
	Notice also that the set of distributions which are $ k $-Markovian with lag $ l $ is a subset of $ (k+l-1) $-Markovian distributions, for which in fact
	\begin{equation}\label{}
	\begin{aligned}
	p_{1:T}(\ddobs_{1:T}) &= p_{1:k+l-1}(\ddobs_{1:k+l-1})\prod_{t=k+l}^{T}p_{t}(\ddobs_{t}|\ddobs_{{t-k-l+1}:{t-1}})\\
	&= p_{1:k+l-1}(\ddobs_{1:k+l-1})\prod_{t=k}^{T-l}p_{t+l}(\ddobs_{t+l}|\ddobs_{{t-k+1}:{t+l-1}});
	\end{aligned}
	\end{equation}
	the additional assumption in Definition~\ref{def:k-markov_lag_l} with respect to $ (k+l-1) $-Markovian is that the conditional distribution for $ \Ddobs_t $ is \textit{not} influenced by the last $ l-1 $ elements.

	In our setting, we can only access $ P_{t+l}(\cdot|\ddobsuptminus) $ for $ k\le t\le T-l $; the marginals $ P_{1:k+l-1} $ are not available. Therefore, we consider the following quantity
	\begin{equation}\label{Eq:app_preq_SR_2}
	S_T^{k,l} (P_{k+l:T}(\cdot|\ddobs_{1:k+l-1}), \ddobs_{k+l:T}):= \frac{1}{T-l-k+1}\sum\limits_{t=k}^{T-l}S(P_{t+l}(\cdot|\ddobs_{t-k+1:t}),\ddobs_{t+l});
	\end{equation}
	in contrast to Eq.~\eqref{Eq:preq_SR} in the main text, we make explicit the dependence on $ k $ and $ l $ in the notation for $ S_T^{k,l} $ and introduce a scaling constant for simplicity, which however does not impact the following arguments.
	The notation in Eq.~\eqref{Eq:app_preq_SR_2} only makes sense if $ P $ is a $ (k+l-1) $-Markovian distribution, as otherwise $ \ddobs_{1:k+l-1} $ would also appear explicitly in the conditioning of $ P_{t+l} $ on the right hand-side. The notation therefore makes sense for $ P $ obtained from the generative network, as that is $ k $-Markovian with lag $ l $ which, as mentioned above, is a specific case of $ (k+l-1) $-Markovian.

	As mentioned in the main text, $ S_T^{k,l} $ is the prequential score and is a SR for distributions over $ \Ddobs_{k+l:T}| \ddobs_{1:k+l-1}$ which are $ (k+l-1) $-Markovian.

	From Eq.~\eqref{Eq:app_preq_SR_2}, we can define the expected SR as
	\begin{equation}\label{Eq:app_preq_SR_2_exp}
	\begin{aligned}
	S_T^{k,l} (P_{k+l:T}(\cdot|\ddobs_{1:k+l-1}), &P^\star_{k+l:T}(\cdot|\ddobs_{1:k+l-1})):=\\
	&\E_{\Ddobs_{k+l:T}\sim P^\star_{k+l:T}(\cdot|\ddobs_{1:k+l-1})} S_T^{k,l} (P_{k+l:T}(\cdot|\ddobs_{1:k+l-1}), \Ddobs_{k+l:T}).
	\end{aligned}
	\end{equation}

	For the scoring rule defined in Eq.~\eqref{Eq:app_preq_SR_2}, the following Theorem holds, which we state in more generality with respect to Theorem~\ref{Th:main_proper} in the main text:
	
	\begin{theorem}\label{th:proper_preq_l}
		If the scoring rule $ S $ is proper, then, for all choices of $ \ddobs_{1:k+l-1} $, the prequential score $ S_T^{k,l} $ in Eq.~\eqref{Eq:app_preq_SR_2} is proper for distributions on $ \Ddobs_{k+l:T}|\ddobs_{1:k+l-1} $ which are $ (k+l-1)- $Markovian; namely, the following inequality holds
		\begin{equation}\label{Eq:preq_SR_proper}
		S_T^{k,l}(P_{k+l:T}^\star(\cdot|\ddobs_{1:k+l-1}), P_{k+l:T}^\star(\cdot|\ddobs_{1:k+l-1}))\le 	S_T^{k,l}(P_{k+l:T}(\cdot|\ddobs_{1:k+l-1}), P_{k+l:T}^\star(\cdot|\ddobs_{1:k+l-1})),
		\end{equation}
		where $ P_{1:T} $ and $ P^\star_{1:T} $ are $ (k+l-1) $-Markovian.
		
		If additionally $ S $ is strictly proper, then, for all choices of $ \ddobs_{1:k+l-1} $, $ S_T^{k,l} $ is proper for distributions on $ \Ddobs_{k+l:T}|\ddobs_{1:k+l-1} $ which are $ k $-Markovian with lag $ l $,
		i.e. the equality in Eq.~\eqref{Eq:preq_SR_proper} only holds if $ P_{k+l:T}(\cdot|\ddobs_{1:k+l-1})=P_{k+l:T}^\star (\cdot|\ddobs_{1:k+l-1})$, where $ P_{1:T} $ and $ P^\star_{1:T} $ are $ k $-Markovian with lag $ l $.
	\end{theorem}
	
	The prequential score $ S_T^{k+l} $ is non-strictly proper for distributions that are $ (k+l-1)$-Markovian but not $ k $-Markovian with lag $ l $. In fact, it builds forecasts from $ P_{k+l:T}^\star(\cdot|\ddobs_{1:k+l-1}) $ with lead of $ l $ timesteps, meaning that the information included in observations $ \ddobs_{t+1:t+l-1} $ is not used in formulating the forecast for $ \Ddobs_{t+l} $. It is therefore unable to distinguish between different distributions for $ \Ddobs_{k+l:T}|\ddobs_{1:k+l-1} $ which have the same conditionals at lead $ l $, but for which the conditionals change if one takes into account $\ddobs_{t+1:t+l-1} $ in forecasting $ \Ddobs_{t+l} $. Therefore, you need to restrict the class of distributions to those in which the value $\ddobs_{t+1:t+l-1} $ does not impact the distribution for $ \Ddobs_{t+l} $ in order to get strict propriety. 
	
	We now prove the Theorem.
	
	\begin{proof}
		The proof steps follow those of Theorem~\ref{th:propriety_generic}.
		
		By definition of proper SR, we have that, for all $ t\ge k $
		\begin{equation}\label{Eq:proper_SR_app}
		\E_{\Ddobs_{t+l}\sim P_{t+l}^\star(\cdot|\ddobsuptminus)} S(P^\star_{t+l}(\cdot|\ddobsuptminus) ,\Ddobs_{t+l}) \le \E_{\Ddobs_{t+l}\sim P_{t+l}^\star(\cdot|\ddobsuptminus)} S(P_{t+l}(\cdot|\ddobsuptminus), \Ddobs_{t+l})
		\end{equation}
		for any conditional distribution $ P_{t+l}(\cdot|\ddobsuptminus) $ and for any values $ \ddobsuptminus $.
		
		For the expected prequential SR, it holds that
		\begin{equation}\label{key}
		\begin{aligned}
		S_T^{k,l}(P^\star_{k+l:T}&(\cdot|\ddobs_{1:k+l-1}), P_{k+l:T}^\star(\cdot|\ddobs_{1:k+l-1})) \\
		&= \E_{\Ddobs_{k+l:T}\sim P_{k+l:T}^\star(\cdot|\ddobs_{1:k+l-1})} S_T^{k,l}(P^\star_{k+l:T}(\cdot|\ddobs_{1:k+l-1}), \Ddobs_{k+l:T}) \\&= \E_{\Ddobs_{1:T}\sim P_{1:T}^\star(\cdot|\ddobs_{1:k+l-1})} S_T^{k,l}(P^\star_{k+l:T}(\cdot|\Ddobs_{1:k+l-1}), \Ddobs_{k+l:T}) \\
		&= \frac{1}{T-l-k+1} \sum_{t=k}^{T-l} \E_{\Ddobs_{1:T}\sim P^\star_{1:T}(\cdot|\ddobs_{1:k+l-1})} S(P^\star_{t+l}(\cdot|\Ddobs_{t-k+1:t}),\Ddobs_{t+l}) \\
		&= \frac{1}{T-l-k+1} \sum_{t=k}^{T-l} \E_{\Ddobs_{1:t+l}\sim P^\star_{1:t+l}(\cdot|\ddobs_{1:k+l-1})} S(P^\star_{t+l}(\cdot|\Ddobs_{t-k+1:t}),\Ddobs_{t+l}); \\
		\end{aligned}
		\end{equation}
		the second equality in the Equation above is trivial but we use it to simplify notation in the following. Now
		\begin{equation}\label{Eq:preq_SR_proof2_2}
		\begin{aligned}
		&\E_{\Ddobs_{1:t+l}\sim P^\star_{1:t+l}(\cdot|\ddobs_{1:k+l-1})} S(P^\star_{t+l}(\cdot|\Ddobs_{t-k+1:t}),\Ddobs_{t+l})\\
		&= \E_{\Ddobs_{t-k+1:t} \sim P^\star_{t-k+1:t}(\cdot|\ddobs_{1:k+l-1})}\left[\E_{\Ddobs_{t+l}\sim P^\star_{t+l}(\cdot|\Ddobs_{t-k+1:t},\ddobs_{1:k+l-1})}  S(P^\star_{t+l}(\cdot|\Ddobsuptminus), \Ddobs_{t+l}) \right] \\
		&= \E_{\Ddobs_{t-k+1:t} \sim P^\star_{t-k+1:t}(\cdot|\ddobs_{1:k+l-1})}\left[\E_{\Ddobs_{t+l}\sim P^\star_{t+l}(\cdot|\Ddobs_{t-k+1:t})}  S(P^\star_{t+l}(\cdot|\Ddobsuptminus), \Ddobs_{t+l}) \right] \\
		&\le \E_{\Ddobs_{t-k+1:t} \sim P^\star_{t-k+1:t}(\cdot|\ddobs_{1:k+l-1})}\left[\E_{\Ddobs_{t+l}\sim P^\star_{t+l}(\cdot|\Ddobs_{t-k+1:t})}  S(P_{t+l}(\cdot|\Ddobsuptminus), \Ddobs_{t+l}) \right]\\
		&= 	\E_{\Ddobs_{1:t+l}\sim P^\star_{1:t+l}(\cdot|\ddobs_{1:k+l-1})} S(P_{t+l}(\cdot|\Ddobs_{t-k+1:t}),\Ddobs_{t+l});
		\end{aligned}
		\end{equation}
		in the first equality above, we have marginalized over all components of $ \Ddobs_{1:t+l} $ which do not appear in the expected quantity and we have used the definition of conditional probability together with the tower property of expectations.	In the second equality, we have exploited the $ (k+l-1)-$Markov property\footnote{Technically, you can relax the $ (k+l-1)- $Markov assumption for the full sequence to assuming $ (k+l-1)-$Markovianity for $ \Ddobs_{1:2k + l -1} $ and independence of $ \Ddobs_{2k + l:T} $ on $ \Ddobs_{1:k+l-1} $; this is however quite artificial.} of $ P^\star $ which ensures that the distribution for $ \Ddobs_{t+l} $ does not depend on $ \Ddobs_{1:t-k} $.
		The inequality holds for any conditional distribution $ P_{t+l}(\cdot|\ddobsuptminus) $ and for any values $ \ddobsuptminus $ thanks to Eq.~\eqref{Eq:proper_SR_app}. Finally, the last equality is   obtained via the reverse of the argument used for the first one.
		
		Now, we can write
		\begin{equation}\label{Eq:preq_SR_proof_2}
		\begin{aligned}
		S_T^{k,l}(P^\star_{k+l:T}&(\cdot|\ddobs_{1:k+l-1}), P_{k+l:T}^\star(\cdot|\ddobs_{1:k+l-1})) \\
		&\le \frac{1}{T-l-k+1} \sum_{t=k}^{T-l} \E_{\Ddobs_{1:t+l}\sim P^\star_{1:t+l}(\cdot|\ddobs_{1:k+l-1})} S(P_{t+l}(\cdot|\Ddobs_{t-k+1:t}),\Ddobs_{t+l})  \\
		&=\frac{1}{T-l-k+1} \sum_{t=k}^{T-l} \E_{\DdobsupT\sim P^\star_{1:T}(\cdot|\ddobs_{1:k+l-1})} S(P_{t+l}(\cdot|\Ddobs_{t-k+1:t}),\Ddobs_{t+l})  \\
		&=\frac{1}{T-l-k+1} \sum_{t=k}^{T-l} \E_{\Ddobs_{k+l:T}\sim P^\star_{k+l:T}(\cdot|\ddobs_{1:k+l-1})} S(P_{t+l}(\cdot|\Ddobs_{t-k+1:t}),\Ddobs_{t+l})  \\
		&= 	S_T^{k,l}(P_{k+l:T}(\cdot|\ddobs_{1:k+l-1}), P_{k+l:T}^\star(\cdot|\ddobs_{1:k+l-1})),
		\end{aligned}
		\end{equation}
		which proves that $ S_T^{k,l} $ is proper for distributions over $ \Ddobs_{k+l:T}|\ddobs_{1:k+l-1} $ which are $ (k+l) -$Markov.
		
		Now, consider $ P_{1:T} $ and $ P_{1:T}^\star $ to be $ k $-Markovian with lag $ l $. The $ \le $ sign in Eq.~\eqref{Eq:preq_SR_proof_2} is an equality if and only if the $ \le  $ sign in Eq.~\eqref{Eq:preq_SR_proof2_2} is an equality for all $ k\le t \le T-l $. As $ S $ is proper, the latter requires $$ \E_{\Ddobs_{t+l}\sim P^\star_{t+l}(\cdot|\ddobs_{t-k+1:t})}  S(P^\star_{t+l}(\cdot|\ddobsuptminus), \Ddobs_{t+l}) = \E_{\Ddobs_{t+l}\sim P^\star_{t+l}(\cdot|\ddobs_{t-k+1:t})}  S(P_{t+l}(\cdot|\ddobsuptminus), \Ddobs_{t+l}) $$ for all values of $ \ddobsuptminus $, If $ S $ is strictly proper, however, the latter is satisfied if and only if $ P_{t+l}(\cdot|\ddobs_{t-k+1:t}) = P^\star_{t+l}(\cdot|\ddobsuptminus)\ \forall \ \ddobsuptminus  $ and for $  k\le t\le T-l $, which implies that $  P_{k+l:T}(\cdot|\ddobs_{1:k+l-1})=P_{k+l:T}^\star (\cdot|\ddobs_{1:k+l-1}) $ due to the $ k $-Markov with lag $ l $ property. This implies that $ S_T $ is strictly proper for distributions which are $ k $-Markov with lag $ l $.
	\end{proof}

	\subsection{Proof and precise statement of the consistency result (Theorem~\ref{Th:main_consistency})}\label{app:consistency}

	We follow here the notation introduced at the start of Appendix~\ref{app:proof_preqSR}. Specifically, $ P^\star $ denotes the data generating distribution for $ (\Ddobs_1, \Ddobs_2, \ldots, \Ddobs_t, \ldots)  = (\Ddobs_t)_t$. 
	
	We consider a model class parametrized by a set of parameters $ \phi $. For such models, we assume the conditional distributions $ P^\phi_{t+l}(\cdot|\ddobs_{1:t}) $ for $ \Ddobs_{t+l} $ only depends on the last $ k $ observations, i.e. $ P^\phi_{t+l}(\cdot|\ddobsupt) =P^\phi_{t+l}(\cdot|\ddobsuptminus) $. Additionally, we assume that the conditional distribution does not depend explicitly on $ t $, such that $ P^\phi_{t+l}(\cdot|\ddobsuptminus) = P^\phi_{(l)}(\cdot|\ddobsuptminus) $, where the bracketed subscript denotes that the forecast is for $ l $ steps ahead. This is the setting considered in the main manuscript.

	In this specific case, therefore, the scoring rule used to penalize the forecast $ P^\phi_{(l)}(\cdot|\ddobsuptminus)$ against the verification $\ddobs_{t+l}  $ (Eq.~\ref{Eq:app_preq_SR_2}) becomes
	\begin{equation}\label{}
	S(P^\phi_{(l)}(\cdot|\ddobsuptminus),\ddobs_{t+l}   ).
	\end{equation}
	Therefore, the prequential score defined in Eq.~\eqref{Eq:app_preq_SR_2} becomes
	\begin{equation}\label{Eq:preq_SR_def}
	S_T^{k,l}(P^\phi_{k+l:T}(\cdot|\ddobs_{1:k+l-1}), \ddobs_{k+l:T}) = \frac{1}{T-l-k+1} \sum_{t=k}^{T-l}	S(P_{(l)}^\phi(\cdot|\ddobsuptminus),\ddobs_{t+l});
	\end{equation}
	notice that we introduce here a scaling constant for simplicity; that however does not impact any of the following arguments.
	Recall also the definition of the expected prequential score
	\begin{equation}\label{Eq:preq_SR_def_exp}
	\begin{aligned}
	S_T^{k,l}(P^\phi_{k+l:T}(\cdot|\ddobs_{1:k+l-1}),&P^\star_{k+l:T}(\cdot|\ddobs_{1:k+l-1})) \\
	&:= \E_{\Ddobs_{k+l:T}\sim P^\star_{k+l:T}(\cdot|\ddobs_{1:k+l-1})} 	S_T^{k,l}(P^\phi_{k+l:T}(\cdot|\ddobs_{1:k+l-1}), \Ddobs_{k+l:T}),
	\end{aligned}
	\end{equation}
	for which we will use the following notation for brevity
	\begin{equation}\label{}
	\tilde S_T^{k,l}(P^\phi_{k+l:T}(\cdot|\ddobs_{1:k+l-1})) := S_T^{k,l}(P^\phi_{k+l:T}(\cdot|\ddobs_{1:k+l-1}),P^\star_{k+l:T}(\cdot|\ddobs_{1:k+l-1}))
	\end{equation}
	As discussed in Appendix~\ref{app:proper_preq_l} and shown in Theorem~\ref{th:proper_preq_l}, provided that $ S $ is strictly proper, $ S_T^{k,l} $ is a strictly proper SR for $ k $-Markovian with lag $ l $ distributions over $ \Ddobs_{k+l:T}|\ddobs_{1:k+l-1} $, for all values of $ \ddobs_{1:k+l-1} $. 
	
	We will also consider the minimizer of the expectation of the expected prequential SR in Eq.~\eqref{Eq:preq_SR_def_exp} with respect to the initial data $ \ddobs_{1:k+l-1} $, i.e.
	\begin{equation}\label{Eq:preq_SR_def_exp_exp}
	\begin{aligned}
	S_T^{k,l\star}(P^\phi_{k+l:T}):&=	\E_{\Ddobsupkplusl\sim P^\star_{1:k+l-1} } 	S_T^{k,l}(P^\phi_{k+l:T}(\cdot|\Ddobs_{1:k+l-1}),P^\star_{k+l:T}(\cdot|\Ddobs_{1:k+l-1})) \\
	&= \E_{\Ddobs_{1:T}\sim P^\star_{1:T}} S_T^{k,l}(P_{k+l:T}^\phi(\cdot|\Ddobsupkplusl), \Ddobs_{k+l:T}).
	\end{aligned}
	\end{equation}
	
	Theorem~\ref{Th:main_consistency} in the main text states that the value of $ \phi $ minimizing the empirical prequential SR (Eq.~\eqref{Eq:preq_SR_def}) converges to both the minimizer of the expected (with respect to $ \DdobsupTfromk|\ddobsupkplusl $ for fixed $ \ddobsupkplusl $) SR in Eq.~\eqref{Eq:preq_SR_def_exp} and to the minimizer of the expected (with respect to $ \DdobsupTfromk $) SR in Eq.~\eqref{Eq:preq_SR_def_exp_exp}. We will split the original result in two separate statements, which hold under similar Assumptions. 
	
	We now set notation and introduce the relevant quantities. From now onwards, we will drop $ k $ and $ l $ for brevity in the definition of $ S_T $; all following results hold for each fixed value of $ k $ and $ l $. We write therefore $ S_T(P_{k+l:T}^\phi(\cdot|\ddobs_{1:k+l-1}), \ddobs_{k+l:T}) = S_T^{k,l}(P_{k+l:T}^\phi(\cdot|\ddobs_{1:k+l-1}), \ddobs_{k+l:T}) $, $ 	\tilde S_T(P^\phi_{k+l:T}(\cdot|\ddobs_{1:k+l-1})) = 	\tilde S_T^{k,l}(P^\phi_{k+l:T}(\cdot|\ddobs_{1:k+l-1})) $ and $ S_T^{\star}(P^\phi_{k+l:T}) = S_T^{k,l\star}(P^\phi_{k+l:T})  $.
	Next, we define the minimizers of the empirical and expected prequential scores
	\begin{equation}\label{Eq:def_minimizers}
	\begin{aligned}
	\hat \phi_T(\ddobsupT) &:  S_T(P_{k+l:T}^{\hat\phi_T(\ddobsupT)}(\cdot|\ddobsupkplusl), \ddobsupTfromk)= \min_{\phi\in\Phi} S_T(P_{k+l:T}^\phi(\cdot|\ddobsupkplusl), \ddobsupTfromk)\\
	\tilde\phi_T(\ddobsupkplusl) &:  \tilde S_T(P^{\tilde\phi_T(\ddobsupkplusl)}_{k+l:T}(\cdot|\ddobs_{1:k+l-1})) = \min_{\phi\in\Phi} \tilde S_T(P^\phi_{k+l:T}(\cdot|\ddobs_{1:k+l-1})).\\	
	\phi^\star_T &:  S_T^\star(P_{k+l:T}^{\phi^\star_T})= \min_{\phi\in\Phi} S_T^\star(P^\phi_{k+l:T}).
	\end{aligned}
	\end{equation}
	
	\subsubsection{Convergence of $ \hat \phi_T $ to $ \phi^\star_T $}\label{app:consistency_statement_part_1}
	We first introduce Assumptions and give the statement linking $ 	\hat \phi_T(\ddobsupT) $ to $ \phi^\star_T $ (Theorem~\ref{Th:consistency_main}). We require the sequence $ (\Ddobs_t)_t $ to be stationary and to satisfy some mixing properties. Specifically, the following Assumptions are required. The precise definition of the mixing properties is postponed to later in Appendix~\ref{app:mixing}.	\begin{enumerate}[label=\textbf{A\arabic*}]
		\item \label{ass:B1} $ \Phi $ is compact.
		\item \label{ass:A1} $ \phi^\star_T $ is unique; additionally, there exist a metric $ d $ on $ \Phi $ such that, for all $ \epsilon>0 $
		\begin{equation}\label{}
		\liminf_{T\to+\infty} \left\{ \min_{\phi: d(\phi, {\phi^\star_T})\ge\epsilon} S_T^\star(P_{k+l:T}^\phi) - S_T^\star(P_{k+l:T}^{\phi^\star_T})    \right\} >0
		\end{equation}
		\item \label{ass:A3} (Asymptotic stationarity) Let $ G_t $ be the marginal distribution of $ \Ddobs_{t-k+1:t+l} $ for $ t\ge k $; then, $ (T-l-k+1)^{-1} \sum_{t=k}^{T-l} G_t $ converges weakly to some probability measure on $ \mathcal{Y}^{k+l} $ as $ T \to\infty$.
		\item \label{ass:A4} Both conditions below are satisfied:
		\begin{enumerate}
			\item (Mixing) Either one of the following holds: 
			\begin{enumerate}
				\item $ (\Ddobs_t)_t $ is $ \alpha $-mixing with mixing coefficient of size $ r/(2r-1) $, with $ r \ge 1 $, or
				\item $ (\Ddobs_t)_t $ is $ \varphi $-mixing with mixing coefficient of size $ r/(r-1) $ with $ r > 1 $.
			\end{enumerate}
			\item (Moment boundedness) Define $ H(\ddobs_{t-k+1:t+l}) = \sup_{\phi \in \Phi } |S(P^\phi(\cdot|\ddobs_{t-k+1:t}), \ddobs_{t+l})| $; then, $$ \sup_{t\ge k} \E\left[H(\Ddobs_{t-k+1:t+l})^{r+\delta}\right] <\infty$$ for some $ \delta>0 $, for the value of $ r $ corresponding to the condition above which is satisfied.
		\end{enumerate}
	\end{enumerate}
	$ S $ being strictly proper and $ P^\phi_{k+l:T}(\cdot|\ddobsupkplusl) $ being a well specified model for $ \DdobsupTfromk|\ddobsupkplusl $ is a sufficient (but not necessary) condition for the uniqueness of $ \phi_T^\star $ in Assumption~\ref{ass:A1} (see Lemma~\ref{lemma:unique} in Appendix~\ref{app:boundedness_SRs}), provided that the parameters $ \phi $ are identifiable. Notice that neural networks do not have identifiable parameters; we require however this assumption to prove the Theorem. In case the parameters are not identifiable, we believe it is possible to show asymptotic convergence of the distributions minimizing the empirical and expected prequential SR, instead of convergence of the parameters. Extending the proof to this setting is technically challenging, as the distance in Assumption~\ref{ass:B1} needs to be replaced by a divergence between probability distributions. We leave this extension for future work.
	
	The rest of Assumption~\ref{ass:A1} is a standard condition ensuring that the function which we are minimizing does not get flatter and flatter around the optimal value as $ T\to\infty $. The asymptotic stationarity condition in Assumption~\ref{ass:A3} is implied by the stronger condition of the marginals $ G_t $ being the same for each $ t $.
	Assumption~\ref{ass:A4}(a) is a mixing condition, ensuring that the dependence between two  different $ \Ddobs_t,\Ddobs_t'  $ decreases as $ t-t'\to\infty $ (defined precisely in Appendix~\ref{app:mixing}). Finally, Assumption~\ref{ass:A4}(b) is a boundedness condition; for the specific case of the Kernel and Energy SR, that can be verified by simpler conditions as discussed in Lemmas~\ref{lemma:bounded_kernel} and \ref{lemma:bounded_energy} in Appendix~\ref{app:boundedness_SRs}.
	
	We will now state our first result.	
	\begin{theorem}\label{Th:consistency_main}
		If $ (\ddobs_{t-k+1:t+l}, \phi) \to S(P^\phi(\cdot|\ddobs_{t-k+1:t}), \ddobs_{t+l}) $ is continuous on $ \mathcal{Y}^{k+l} \times \Phi $, and if Assumptions \ref{ass:B1}, \ref{ass:A1}, \ref{ass:A3} and \ref{ass:A4} hold, then $ d(\hat \phi_T(\DdobsupT), \phi^\star_T) \to 0 $ with probability 1  with respect to $(\Ddobs_t)_t\sim P^\star $.
	\end{theorem}
	
	The Theorem above relies on a generic consistency result (discussed in Appendix~\ref{app:generic_consistency}) for which a uniform law of large numbers is required. Such a uniform law of large numbers can be obtained under stationarity and mixing conditions; we report in Appendix~\ref{app:ULLN} a result ensuring this. We prove Theorem~\ref{Th:consistency_main} by combining the above two elements in Appendix~\ref{app:proof_consistency}.

	\subsubsection{Convergence of $ \hat \phi_T $ to $ \tilde \phi_T $}\label{app:consistency_statement_part_2}
	We now give the statement linking $ 	\hat \phi_T(\ddobsupT) $ to $ \tilde\phi_T(\ddobsupt) $ (Theorem~\ref{Th:consistency_fixed_corollary}). We will require similar Assumptions to what considered above, but holding for fixed values of $ \ddobs_{1:k+l-1} $:
	\begin{enumerate}[label=\textbf{B\arabic*}]
		\item \label{ass:C2} $ 	\tilde\phi_T(\ddobsupkplusl) $ is unique; additionally, there exist a metric $ d $ on $ \Phi $ such that, for all $ \epsilon>0 $
		\begin{equation}\label{}
		\liminf_{T\to+\infty} \left\{ \min_{\phi: d(\phi, {	\tilde\phi_T(\ddobsupkplusl)})\ge\epsilon} \tilde S_T(P^\phi_{k+l:T}(\cdot|\ddobs_{1:k+l-1})) - \tilde S_T(P^{\tilde\phi_T(\ddobsupkplusl)}_{k+l:T}(\cdot|\ddobs_{1:k+l-1}))    \right\} >0
		\end{equation}
		\item \label{ass:C3} (Asymptotic stationarity) Let $\tilde G_t $ be the marginal distribution of $ \Ddobs_{t-k+1:t+l}|\ddobsupkplusl $ for $ t\ge k $; then, $$ (T-l-k+1)^{-1} \sum_{t=k}^{T-l} \tilde G_t $$ converges weakly to some probability measure on $ \mathcal{Y}^{k+l} $ as $ T \to\infty$.
		\item \label{ass:C4} Both conditions below are satisfied:
		\begin{enumerate}
			\item (Mixing) Let $ (\Ddsim_t)_t \sim P^\star(\cdot|\ddobsupkplusl)$; then, either one of the following holds: 
			\begin{enumerate}
				\item $ (\Ddsim_t)_t $ is $ \alpha $-mixing with mixing coefficient of size $ r/(2r-1) $, with $ r \ge 1 $, or
				\item $ (\Ddsim_t)_t $ is $ \varphi $-mixing with mixing coefficient of size $ r/(r-1) $ with $ r > 1 $.
			\end{enumerate}
			\item (Moment boundedness) Define $ H(\ddobs_{t-k+1:t+l}) = \sup_{\phi \in \Phi } |S(P^\phi(\cdot|\ddobs_{t-k+1:t}), \ddobs_{t+l})| $; then, $$ \sup_{t\ge k} \E_{\Ddobs_{t-k+1:t+l}|\ddobsupkplusl}\left[H(\Ddobs_{t-k+1:t+l})^{r+\delta}\right] <\infty$$ for some $ \delta>0 $, for the value of $ r $ corresponding to the condition above which is satisfied.
		\end{enumerate}
	\end{enumerate}
	
	We can therefore state the following: 
	\begin{theorem}\label{Th:consistency_fixed}
		If $ (\ddobs_{t-k+1:t+l}, \phi) \to S(P^\phi(\cdot|\ddobs_{t-k+1:t}), \ddobs_{t+l}) $ is continuous on $ \mathcal{Y}^{k+l} \times \Phi $, and if Assumptions  \ref{ass:B1}, \ref{ass:C2}, \ref{ass:C3} and \ref{ass:C4} hold, then $ d(\hat \phi_T(\ddobsupkplusl, \DdobsupTfromk), \tilde\phi_T(\ddobsupkplusl)) \to 0 $ with probability 1  with respect to $(\Ddobs_t)_t\sim P^\star(\cdot|\ddobsupkplusl) $.
	\end{theorem}
	Notice how now in $ \hat \phi_T $ we split the dependence with respect to the fixed $ \ddobsupkplusl $ and the random $ \DdobsupTfromk $. 
	\begin{proof}
		Theorem~\ref{Th:consistency_fixed} is proven following the same steps as Theorem~\ref{Th:consistency_main} (given in Appendix~\ref{app:proof_consistency}). Specifically, Corollary~\ref{Th:ULLN_2} can be used to obtain a uniform Law of Large Numbers such as in Assumption~\ref{ass:A2}. Then, an equivalent to Theorem~\ref{Th:consistency} can be shown following the exact same steps. That implies the result of Theorem~\ref{Th:consistency_fixed}.
	\end{proof}

	The above result is saying that, for the sequence $ (\Ddobs_t)_t $ conditioned on $ \ddobsupkplusl $, if stationarity and mixing conditions hold for a fixed $ \ddobsupkplusl  $, then the empirical minimizer $ \hat \phi_T $ converges to the minimizer $ \tilde \phi $, both with fixed $ \ddobsupkplusl  $. 
	
	Clearly, if the above Assumptions hold for all values of $ \ddobsupkplusl  $, the statement also does. This is made precise by the following Corollary: 
	\begin{corollary}\label{Th:consistency_fixed_corollary}
		If Assumptions  \ref{ass:B1}, \ref{ass:C2}, \ref{ass:C3} and \ref{ass:C4} hold almost surely for $ \Ddobsupkplusl \sim P^\star_{1:k+l-1}$, and if $ (\ddobs_{t-k+1:t+l}, \phi) \to S(P^\phi(\cdot|\ddobs_{t-k+1:t}), \ddobs_{t+l}) $ is continuous on $ \mathcal{Y}^{k+l} \times \Phi $, then $$ d(\hat \phi_T(\Ddobsupkplusl, \DdobsupTfromk), \tilde\phi_T(\Ddobsupkplusl)) \to 0 $$ with probability 1  with respect to $(\Ddobs_t)_t\sim P^\star$.
	\end{corollary}
	\begin{proof}
		If Assumptions \ref{ass:B1}, \ref{ass:C2}, \ref{ass:C3} and \ref{ass:C4} hold almost surely for $ \Ddobsupkplusl \sim P^\star_{1:k+l-1}$, and under the continuity condition, the following statement holds with probability 1 with respect to $ \Ddobsupkplusl \sim P^\star_{1:k+l-1}$:
		``$ d(\hat \phi_T(\Ddobsupkplusl, \DdobsupTfromk), \tilde\phi_T(\Ddobsupkplusl)) \to 0 $ with probability 1  with respect to $(\Ddobs_t)_t\sim P^\star(\cdot|\Ddobsupkplusl) $,''
		from which the result follows by considering that a statement holding with probability 1 with respect to $(\Ddobs_t)_t\sim P^\star(\cdot|\Ddobsupkplusl) $, for each value $ \Ddobsupkplusl $ takes, and with probability 1 with respect to $ \Ddobsupkplusl \sim P^\star_{1:k+l-1} $ holds almost surely with respect to $(\Ddobs_t)_t\sim P^\star$.
	\end{proof}
	
	\subsubsection{Putting the two results together}
	Finally, we also have the following, which correspond to Theorem~\ref{Th:main_consistency} in the main text with the two sets of assumptions for the conditional and unconditional case kept separate:
	\begin{corollary}
		If Assumptions \ref{ass:B1}, \ref{ass:A1}, \ref{ass:A3} and \ref{ass:A4} hold, and if Assumptions \ref{ass:C2}, \ref{ass:C3} and \ref{ass:C4} hold almost surely for $ \Ddobsupkplusl \sim P^\star_{1:k+l-1}$, and if $ (\ddobs_{t-k+1:t+l}, \phi) \to S(P^\phi(\cdot|\ddobs_{t-k+1:t}), \ddobs_{t+l}) $ is continuous on $ \mathcal{Y}^{k+l} \times \Phi $, then
		\begin{enumerate}
			\item  $ d(\hat \phi_T(\DdobsupT), \phi^\star_T) \to 0 $ with probability 1 with respect to $ (\Ddobs_t)_t  \sim P^\star $;
			\item $ d(\hat \phi_T(\DdobsupT), \tilde\phi_T(\Ddobsupkplusl)) \to 0 $ with probability 1 with respect to $ (\Ddobs_t)_t  \sim P^\star $;
			\item $ d(\phi^\star_T, \tilde\phi_T(\Ddobsupkplusl)) \to 0 $ with probability 1 with respect to $ \Ddobsupkplusl\sim P^\star_{1:k+l-1} $.
		\end{enumerate} 
	\end{corollary}
	\begin{proof}
		Under the Assumptions, both Theorem~\ref{Th:consistency_main} and Corollary~\ref{Th:consistency_fixed_corollary} hold, from which the first two statements follow. For the last statement, applying the triangle inequality yields
		\begin{equation}\label{key}
		d(\phi^\star_T, \tilde\phi_T(\Ddobsupkplusl)) \le d(\hat \phi_T(\DdobsupT), \tilde\phi_T(\Ddobsupkplusl)) +  d(\hat \phi_T(\DdobsupT), \phi^\star_T)  \to 0.
		\end{equation}As the left-hand side above depends only on $ \Ddobsupkplusl $, the result holds almost surely with respect to $ \Ddobsupkplusl\sim P^\star_{1:k+l-1} $.
	\end{proof}
	
	In case in which all the Assumption hold, therefore, the minimizer of the expected prequential SR over $ \DdobsupTfromk|\Ddobsupkplusl $ converges to the minimizer of the expected prequential SR over $ \DdobsupT $, which is a deterministic quantity. Therefore, this result is saying that for large $ T $, $ \tilde \phi_T $ does not depend on the initial conditions, as it is intuitive under mixing and stationarity of $ (\Ddobs_t)_t $. Indeed, the same holds for the empirical minimizer $ \hat \phi_T $, in which no expectation at all is computed.

	In the next Subsections, we will discuss how to verify the Assumptions in some specific cases, and then move to introducing preliminary results for proving Theorem~\ref{Th:consistency_main}, which we do in Appendix~\ref{app:proof_consistency}. As mentioned above, the proof of Theorem~\ref{Th:consistency_fixed} follows the same steps as the one for Theorem~\ref{Th:consistency_main}, but with the corresponding set of Assumptions. For this reason, we do not give that in details.
	
	\subsubsection{Verifying the Assumptions in specific cases}\label{app:boundedness_SRs}
	
	Before delving into proving Theorem~\ref{Th:consistency_main}, we here show sufficient conditions under which $ \phi^\star_T $ and $ \tilde \phi_T(\ddobsupkplusl) $ are unique and under which Assumption~\ref{ass:A4}(b) holds. Specifically, for the former (Lemma~\ref{lemma:unique}), we consider the model $ P^\phi_{k+l:T}(\cdot|\ddobsupkplusl) $ to be a well specified model and the scoring rule $ S $ to be strictly proper; for the latter, we consider instead the Kernel and Energy SR and obtain more precise conditions, which are easily satisfied.

	First, consider uniqueness of $ \phi^\star_T $:
	
	\begin{lemma}\label{lemma:unique}
		If both
		\begin{itemize}
			\item $ S $ is strictly proper, and 
			\item for all values of $ T $, $ P^\phi_{k+l:T}(\cdot|\ddobsupkplusl) $ is a well specified model for $ \DdobsupTfromk|\ddobsupkplusl $ and the mapping $ \phi  \to P^\phi_{k+l:T}(\cdot|\ddobsupkplusl) $ is unique,
		\end{itemize} then $ \phi^\star_T $ and $ \tilde \phi_T(\ddobsupkplusl) $ are unique for all values of $ T $ and $ \ddobsupkplusl $.
	\end{lemma}
	\begin{proof}
		If $ P^\phi $ is well specified, there exists a $ \phi^\star $ such that $$ P^\star_{k+l:T}(\cdot|\ddobsupkplusl) = P^{\phi^\star}_{k+l:T}(\cdot|\ddobsupkplusl)\ \forall \ T, \ \forall \ \ddobsupkplusl. $$ Notice that this implies that $ P^\star $ is $ k $-Markovian with lag $ l $. If $ S $ is strictly proper, we have by Theorem~\ref{th:proper_preq_l} that
		\begin{equation}\label{}
		\phi^\star =\argmin_{\phi\in\Phi} S_T(P_{k+l:T}^\phi(\cdot|\ddobsupkplusl), P_{k+l:T}^\star(\cdot|\ddobsupkplusl)
		\end{equation}
		is unique, for all $ \ddobsupkplusl $. Therefore, $ \tilde \phi_T(\ddobsupkplusl) = \phi^\star $ for all values of $ \ddobsupkplusl $. Recalling now the definition of $ S_T^\star(P^\phi_{k+l:T}) $ in Eq.~\eqref{Eq:preq_SR_def_exp_exp}, notice that the quantity inside the expectation $ \E_{\Ddobsupkplusl\sim P^\star_{1:k+l-1}} $ is minimized uniquely by $ \phi= \phi^\star $, so that 
		$ S_T^\star(P^\phi_{k+l:T}) $ is also uniquely minimized by $ \phi_T^\star= \phi^\star $.
	\end{proof}
	
	The following two Lemmas show conditions under which Assumption~\ref{ass:A4}(b) holds.
	
	\begin{lemma}\label{lemma:bounded_kernel}
		When $ S = S_k $, Assumption~\ref{ass:A4}(b) is verified for a kernel $ k $ which satisfies either of the following:
		\begin{enumerate}
			\item with probability 1 with respect to $ (\Ddobs_t)_t\sim P^\star $,\footnote{Put simply, this condition means that the following has to be true for all observed sequences $ (\ddobs_t)_t $ which can be generated by the distribution $ P^\star $.} for all $ t\ge k $ and $ \phi $, \\ $ \E_{\Ddsim,\Ddsim'\sim P^\phi_{(l)}(\cdot|\Ddobs_{t-k+1:t})}|k(\Ddsim,\Ddsim')|<\infty  $ and $ \E_{\Ddsim\sim P^\phi_{(l)}(\cdot|\Ddobs_{t-k+1:t})}|k(\Ddsim,\Ddobs_{t+l})|<\infty $;
			\item $ k $ is bounded, i.e. $ |k(\ddobs,\ddsim)| <\kappa<+\infty \ \forall \ \ddobs,\ddsim \in \mathcal Y$ (this implies the above condition).
		\end{enumerate} 
	\end{lemma}
	
	\begin{proof}
		First, notice that $ \sup_{t\ge k} \E\left[H(\Ddobs_{t-k+1:t+l})^{r+\delta}\right] <\infty \iff \E\left[H(\Ddobs_{t-k+1:t+l})^{r+\delta}\right] <\infty\ \forall \ t\ge k $.

		Consider the kernel SR $ S=S_k $ 
		\begin{equation}\label{Eq:kernel_SR_bounded}
		\begin{aligned}
		|S_k(P^\phi_{(l)}(\cdot|\ddobs_{t-k+1:t}), \ddobs_{t+l})| &= |\E_{\Ddsim,\Ddsim'\sim P^\phi_{(l)}(\cdot|\ddobs_{t-k+1:t})}[k(\Ddsim,\Ddsim') - 2k(\Ddsim, \ddobs_{t+l})]|\\
		&\le \E_{\Ddsim,\Ddsim'\sim P^\phi_{(l)}(\cdot|\ddobs_{t-k+1:t})}|k(\Ddsim,\Ddsim') - 2k(\Ddsim, \ddobs_{t+l})|\\
		&\le \E_{\Ddsim,\Ddsim'\sim P^\phi_{(l)}(\cdot|\ddobs_{t-k+1:t})}[|k(\Ddsim,\Ddsim')| + 2|k(\Ddsim, \ddobs_{t+l})|].
		\end{aligned}
		\end{equation}
		We first show why condition 1 yields the result. If, with probability 1 with respect to $ (\Ddobs_t)_t\sim P^\star $, for all $ t\ge k $ and $ \phi $ \begin{equation}\label{Eq:kernel_condition_1}
		\E_{\Ddsim,\Ddsim'\sim P^\phi_{(l)}(\cdot|\Ddobs_{t-k+1:t})}|k(\Ddsim,\Ddsim')|\le \kappa_1<\infty  \text{   and    } \E_{\Ddsim\sim P^\phi_{(l)}(\cdot|\Ddobs_{t-k+1:t})}|k(\Ddsim,\Ddobs_{t+l})|\le \kappa_2<\infty, 
		\end{equation}we have that
		\begin{equation}\label{}
		\begin{aligned}
		|S_k(P^\phi_{(l)}(\cdot|\Ddobs_{t-k+1:t}), \Ddobs_{t+l})|\le \kappa_1 + 2\kappa_2<\infty,
		\end{aligned}
		\end{equation}
		from which
		\begin{equation}\label{key}
		\begin{aligned}
		\E\left[H(\Ddobs_{t-k+1:t+l})^{r+\delta}\right] &= \E\left[\left( \sup_{\phi \in \Phi } |S_k(P^\phi_{(l)}(\cdot|\Ddobs_{t-k+1:t}), \Ddobs_{t+l})| \right)^{r+\delta}\right]\\
		&\le \E\left[\left( \sup_{\phi \in \Phi } \kappa_1 + 2\kappa_2 \right)^{r+\delta}\right] = \left(\kappa_1 + 2\kappa_2\right)^{r+\delta}<\infty.
		\end{aligned}
		\end{equation}
		Now, condition 2 implies condition 1. Therefore, condition 2 yields the result.
	\end{proof}
	
	\begin{lemma}\label{lemma:bounded_energy}
		When $ S = S_E^{(\beta)} $, Assumption~\ref{ass:A4}(b) is verified  when either of the following holds: 
		\begin{enumerate}
			\item with probability 1 with respect to $ (\Ddobs_t)_t\sim P^\star $, for all $ t \ge k$ and $ \phi $,\\ $ \E_{\Ddsim,\Ddsim'\sim P^\phi_{(l)}(\cdot|\Ddobs_{t-k+1:t})}||\Ddsim-\Ddsim'||<\infty  $ and $ \E_{\Ddsim\sim P^\phi_{(l)}(\cdot|\Ddobs_{t-k+1:t})}||\Ddsim-\Ddobs_{t+l}||<\infty $;
			\item the space $ \mathcal{Y} $ is bounded, such that $ ||\ddobs|| \le B<\infty \ \forall \ \ddobs\in\mathcal{Y}$ (this implies the first condition);
			\item $\beta\ge1$, $ \E ||\Ddobs_{t+l}||^{\beta(r+\delta)} <\infty $ for all $ t $ and, with probability 1 with respect to $ (\Ddobs_t)_t\sim P^\star $, for all $ t $ and $ \phi $, $  \E_{\Ddsim\sim P^\phi_{(l)}(\cdot|\ddobs_{t-k+1:t})}||\Ddsim||^\beta \le B<\infty$.
		\end{enumerate}	
		
	\end{lemma}
	
	\begin{proof}
		First, notice that $ \sup_{t\ge k} \E\left[H(\Ddobs_{t-k+1:t+l})^{r+\delta}\right] <\infty \iff \E\left[H(\Ddobs_{t-k+1:t+l})^{r+\delta}\right] <\infty\ \forall \ t\ge k $.
		
		Notice how the kernel SR recovers the Energy SR when $ k(\ddobs,\ddsim) = -||\ddobs-\ddsim  ||^\beta $; condition 1 for the kernel SR corresponds therefore to condition 1 for the Energy SR; therefore, the result holds under condition 1.

		For condition 2 for the Energy SR, notice that \begin{equation}\label{key}
		|k(\ddobs,\ddsim)|= ||\ddobs-\ddsim  ||^\beta \le \left( ||\ddobs||+||\ddsim||  \right)^\beta \le (2B)^\beta,
		\end{equation}
		where the first inequality comes from applying the triangle inequality and the second comes from condition 2 for the Energy SR. Therefore, condition 2 for the Energy SR implies condition 2 for the corresponding Kernel SR, from which the result follows. 
		
		Finally, an alternative route leads to condition 3. Specifically, for the Energy SR, Equation~\eqref{Eq:kernel_SR_bounded} becomes 
		\begin{equation}\label{Eq:eng_SR_bounded}
		\begin{aligned}
		&|S_E^{(\beta)}(P^\phi_{(l)}(\cdot|\ddobs_{t-k+1:t}), \ddobs_{t+l})| \\
		&\qquad\le \E_{\Ddsim,\Ddsim'\sim P^\phi_{(l)}(\cdot|\ddobs_{t-k+1:t})}[||\Ddsim-\Ddsim'||^\beta + 2||\Ddsim- \ddobs_{t+l}||^\beta]\\
		&\qquad \le \E_{\Ddsim,\Ddsim'\sim P^\phi_{(l)}(\cdot|\ddobs_{t-k+1:t})}[(||\Ddsim||+||\Ddsim'||)^\beta + 2(||\Ddsim||+||\ddobs_{t+l}||)^\beta]
		\end{aligned}
		\end{equation}
		by triangle inequality. Now, for any $ \beta>1 $, $ a,b>0 $, $ (a+b)^\beta\le2^{\beta-1}(a^\beta+b^\beta) $;\footnote{This inequality is well-known and can be shown by convexity.} therefore, 
		\begin{equation}\label{}
		\begin{aligned}
		&|S_E^{(\beta)}(P^\phi_{(l)}(\cdot|\ddobs_{t-k+1:t}), \ddobs_{t+l})| \\
		&\qquad  \le \E_{\Ddsim,\Ddsim'\sim P^\phi_{(l)}(\cdot|\ddobs_{t-k+1:t})}[2^{\beta-1}(||\Ddsim||^\beta+||\Ddsim'||^\beta)+ 2^\beta(||\Ddsim||^\beta+||\ddobs_{t+l}||^\beta)].
		\end{aligned}
		\end{equation}
		From the above, we have that 
		\begin{equation}\label{key}
		\begin{aligned}
		&\E\left[H(\Ddobs_{t-k+1:t+l})^{r+\delta}\right] = \E\left[\left( \sup_{\phi \in \Phi } |S_E^{(\beta)}(P^\phi_{(l)}(\cdot|\Ddobs_{t-k+1:t}), \Ddobs_{t+l})| \right)^{r+\delta}\right]\\
		\le &\E\left[\left( \sup_{\phi \in \Phi } \E_{\Ddsim,\Ddsim'\sim P^\phi_{(l)}(\cdot|\Ddobs_{t-k+1:t})}[2^{\beta-1}(||\Ddsim||^\beta+||\Ddsim'||^\beta)+ 2^\beta(||\Ddsim||^\beta+||\Ddobs_{t+l}||^\beta)] \right)^{r+\delta}\right].
		\end{aligned}
		\end{equation}
		If, with probability 1 with respect to $ (\Ddobs_t)_t\sim P^\star $, for all $ t\ge k $ and $ \phi $, $ \E_{\Ddsim\sim P^\phi_{(l)}(\cdot|\Ddobs_{t-k+1:t})}||\Ddsim||^\beta \le B<\infty$, we have therefore
		\begin{equation}\label{key}
		\begin{aligned}
		\E\left[H(\Ddobs_{t-k+1:t+l})^{r+\delta}\right] &\le \E\left[\left(2^{\beta-1}(B + B)+ 2^\beta(B+||\Ddobs_{t+l}||^\beta) \right)^{r+\delta}\right] \\&=\E\left[\left( 2^{\beta+1}B + 2^\beta||\Ddobs_{t+l}||^\beta \right)^{r+\delta}\right].
		\end{aligned}
		\end{equation}
		Now, denote $ \delta' = r+\delta $; $\delta'>1$ by assumption. It holds therefore, as above, $ (a+b)^{\delta'}\le2^{\delta'-1}(a^{\delta'}+b^{\delta'}) $ for $ a,b>0 $; we have therefore that
		\begin{equation}\label{key}
		\begin{aligned}
		\E\left[H(\Ddobs_{t-k+1:t+l})^{\delta'}\right] &\le 2^{-1}\left( 2^{\beta+2}B\right)^{\delta'} +  2^{\delta'(\beta+1)-1}\E ||\Ddobs_{t+l}||^{\beta\delta'};
		\end{aligned}
		\end{equation}
		the above expression is therefore bounded whenever $ \E ||\Ddobs_{t+l}||^{\beta(r+\delta)} <\infty $.
	\end{proof}

	\subsubsection{Defining the mixing conditions}  \label{app:mixing}
	Here, we give the precise definitions for the mixing conditions stated in Assumption~\ref{ass:A4}(a). More background on the following definitions can be found, for instance, in \cite{bradley2005basic}.
	
	\begin{definition}[Measures of dependence]
		Consider a probability space $ (\Omega, \mathcal{F}, P )$; for any two sigma algebras $ \mathcal{A} \subseteq\mathcal{F}$ and $ \mathcal{B} \subseteq\mathcal{F}$, define
		\begin{equation}\label{}
		\begin{aligned}
		\alpha_P(\mathcal{A}, \mathcal{B}) &:= \sup_{A\in\mathcal{A},B\in\mathcal{B}} \left|P(A\cap B) - P(A)P(B)\right|,\\
		\varphi_P(\mathcal{A}, \mathcal{B}) &:= \sup_{A\in\mathcal{A},B\in\mathcal{B}: P(B)>0} \left|P(B|A) - P(B)\right|.
		\end{aligned}
		\end{equation}
	\end{definition}

	For $ 1\le r\le s \le \infty $, define the Borel $ \sigma $-algebra of events generated from \\$( \Ddobs_r, \Ddobs_{r+1}, \ldots, \Ddobs_{s-1}, \Ddobs_s )$ as $ \mathcal{G}_r^s $. Then, we define
	\begin{equation}\label{Eq:mixing_coeff_Z}
	\alpha^\Ddobs(m) = \sup_{r\ge 1} \alpha_{P^\star}(\mathcal{G}_{1}^r, \mathcal{G}_{r+m}^{+\infty}), \quad\varphi^\Ddobs(m) = \sup_{r\ge 1} \varphi_{P^\star}(\mathcal{G}_{1}^r, \mathcal{G}_{r+m}^{+\infty}).
	\end{equation}	
	
	\begin{definition}
		The random sequence $ (\Ddobs_t)_{t} $ is said $ \alpha $-mixing if $ \alpha^\Ddobs(m )\to 0$ as $ m\to\infty $ and $ \varphi $-mixing if $ \varphi^\Ddobs(m )\to 0$ as $ m\to\infty $. It can be seen that $ \varphi $-mixing implies $ \alpha $-mixing \citep{domowitz1982}.
	\end{definition}
	
	\begin{definition}
		We say that the mixing coefficients $ \varphi^\Ddobs(m) $ are of size $ s $ \citep{domowitz1982} if $ \varphi^\Ddobs(m) = \mathcal{O}(m^{-\lambda}) $ for $\lambda>s$; similar definition can be given for the coefficients $ \alpha^\Ddobs(m) $.
	\end{definition}
	
	In \cite{bradley2005basic}, the definitions for the quantities above consider a sequence $ (\Ddsim_t)_{t\in\mathbb Z} $, and defined
	$$ \alpha^\Ddsim(m) = \sup_{r\in\mathbb Z} \alpha_{P}(\mathcal{G}_{-\infty}^r, \mathcal{G}_{r+m}^{+\infty}), $$
	for some distribution $ P $, and similar for $ \phi^\Ddsim(m) $. Our definition can be cast in this way by defining $ \Ddsim_t = \Ddobs_t \ \forall \ t\ge 1 $ and $ \Ddsim_t = 0 \ \forall\ t \le 0 $.

	\subsubsection{Generic consistency result}\label{app:generic_consistency}
	
	We consider here the following Assumption:

	\begin{enumerate}[label=\textbf{A\arabic*}]
		\setcounter{enumi}{4}
		\item \label{ass:A2}(Uniform Law of Large Numbers.) The following holds with probability $ 1 $ with respect to $(\Ddobs_t)_t\sim P^\star $
		\begin{equation}\label{}
		\sup_{\phi \in \Phi} \left| S_T(P^\phi_{k+l:T}(\cdot|\Ddobsupkplusl), \DdobsupTfromk) - S^\star_T(P^\phi_{k+l:T})  \right| \to 0.
		\end{equation}
	\end{enumerate}
	
	We give here a consistency result more general than Theorem~\ref{Th:consistency_main}, as in fact Assumption~\ref{ass:A2} is more general than the stationarity and mixing conditions in Assumption~\ref{ass:A3} and \ref{ass:A4}.

	\begin{theorem}[Theorem 5.1 in \citealp{skouras1998optimal}]\label{Th:consistency}
		If Assumptions \ref{ass:A1} and \ref{ass:A2} hold, then $ d(\hat \phi_T(\DdobsupT), \phi^\star_T) \to 0 $ with probability 1  with respect to $(\Ddobs_t)_t\sim P^\star$.
	\end{theorem}
	We report here a proof for ease of reference.
	\begin{proof}
		By the definition of $ \liminf $, for a fixed $ \epsilon>0 $, Assumption \ref{ass:A1} implies that there exists $ T_1(\epsilon) $ such that
		\begin{equation}\label{Eq:proof_inflim}
		\delta(\epsilon) :=\left\{  \inf_{T>T_1(\epsilon)} \min_{\phi: d(\phi, {\phi^\star_T})\ge\epsilon} S_T^\star(P_{k+l:T}^\phi) - S_T^\star(P_{k+l:T}^{\phi^\star_T})    \right\} >0.
		\end{equation}

		Due to Assumption \ref{ass:A2}, with probability 1 with respect to $(\Ddobs_t)_t\sim P^\star$, there exists $ T_2((\Ddobs_t)_t, \delta(\epsilon)) $ such that, for all $ T>T_2((\Ddobs_t)_t, \delta(\epsilon))$
		\begin{equation}
		\left| S_T(P^{\phi^\star_T}_{k+l:T}(\cdot|\Ddobsupkplusl), \DdobsupTfromk) - S^\star_T(P^{\phi^\star_T}_{k+l:T})  \right| <\delta(\epsilon)/2,
		\end{equation}
		which implies
		\begin{equation}\label{Eq:proof_ineq1}
		\begin{aligned}
		S^\star_T(P^{\phi^\star_T}_{k:T}) &> S_T(P^{\phi^\star_T}_{k:T}(\cdot|\Ddobsupkplusl), \DdobsupTfromk) -\delta(\epsilon)/2 \\
		&\ge  S_T(P^{\hat \phi_T  (\DdobsupT)}_{k+l:T}(\cdot|\Ddobsupkplusl), \DdobsupTfromk) -\delta(\epsilon)/2,
		\end{aligned}
		\end{equation}
		where the second inequality is valid thanks to the definition of $ \hat \phi_T  (\DdobsupT) $.
		
		Similarly, by exploiting Assumption \ref{ass:A2} again, with probability 1 with respect to $(\Ddobs_t)_t\sim P^\star $, there exists $  T_3((\Ddobs_t)_t, \delta(\epsilon)) $ such that, for all $ T>T_3((\Ddobs_t)_t, \delta(\epsilon))$
		\begin{equation}\label{Eq:proof_ineq2}
		\left| S^\star_T(P^{\hat \phi_T (\DdobsupT)}_{k+l:T}) - S_T(P^{\hat \phi_T (\DdobsupT)}_{k+l:T}(\cdot|\Ddobsupkplusl), \DdobsupTfromk) \right| <\delta(\epsilon)/2.
		\end{equation}
		Then, with probability 1 with respect to $(\Ddobs_t)_t\sim P^\star$, for all\\$ T>\max\{T_2((\Ddobs_t)_t, \delta(\epsilon)), T_3((\Ddobs_t)_t, \delta(\epsilon))\} $
		\begin{equation}\label{Eq:proof_ineq3}
		\begin{aligned}
		S_T^\star(P^{\hat \phi_T (\DdobsupT)}_{k+l:T}) - S_T^\star(P^{\phi^\star_T}_{k+l:T}) &\le S_T^\star(P^{\hat \phi_T (\DdobsupT)}_{k+l:T}) - S_T(P^{\hat \phi_T (\DdobsupT)}_{k+l:T}(\cdot|\Ddobsupkplusl), \DdobsupTfromk) + \delta(\epsilon)/2 \\
		&<  \delta(\epsilon)/2+ \delta(\epsilon)/2 = \delta(\epsilon),
		\end{aligned}
		\end{equation}
		where the first inequality is thanks to Eq.~\eqref{Eq:proof_ineq1} and the second is thanks to Eq~\eqref{Eq:proof_ineq2}. 
		
		Now, Eq.~\eqref{Eq:proof_inflim} and Eq.~\eqref{Eq:proof_ineq3} both hold with probability 1 with respect to $(\Ddobs_t)_t\sim P^\star $ for all $ T>\max\{T_1(\delta(\epsilon)), T_2((\Ddobs_t)_t, \delta(\epsilon)), T_3((\Ddobs_t)_t, \delta(\epsilon))\} $. Notice that Eq.~\eqref{Eq:proof_ineq3} ensures that the difference considered in Eq.~\eqref{Eq:proof_inflim} is smaller than $ \delta(\epsilon )$ for $ \phi =  \hat \phi_T(\DdobsupT)$; However, Eq.~\eqref{Eq:proof_inflim} states that the same difference is larger or equal than $ \delta(\epsilon) $ for all $ \phi:d(\phi, {\phi^\star_T})\ge\epsilon $, from which it follows that 
		$  d({\hat \phi_T(\DdobsupT)}, {\phi^\star_T})<\epsilon $ with probability 1 with respect to $(\Ddobs_t)_t\sim P^\star$. As $ \epsilon $ is however arbitrary, it follows that, with probability 1 with respect to $(\Ddobs_t)_t\sim P^\star$
		\begin{equation}\label{}
		d({\hat \phi_T(\DdobsupT)}, {\phi^\star_T}) \to 0.
		\end{equation}
	\end{proof}

	\subsubsection{Uniform law of large numbers}\label{app:ULLN}
	
	We will here show how the Uniform Law of Large Numbers in Assumption~\ref{ass:A2} can be obtained from the stationarity and mixing conditions in~\ref{ass:A3} and~\ref{ass:A4}. To this aim, we exploit a result in \cite{potscher1989uniform}. 
	
	We consider now a generic sequence of random variables $ \Ddz_t \in \mathcal{Z}$, and a function $ q:\mathcal{Z}\times \Phi\to \R $. 
	Let us denote now by $ \mathcal{F} $ the Borel $ \sigma $-algebra generated by the sequence $ (\Ddz_t)_t $, $ \Omega_\Ddz $ the space of realizations of $ (\Ddz_t)_t $ and $ Q^\star $ the probability distribution for it.

	Consider the following Assumptions:

	\begin{enumerate}[label=\textbf{C\arabic*}]
		\item \label{ass:B2} (Dominance condition) For $ D(\ddz) = \sup_{\phi \in \Phi } |q(\ddz,\phi) |  $, there is some $ \delta>0 $ such that $$ \sup_t \frac{1}{N}\sum_{t=1}^N \E\left[ D(\Ddz_t)^{1+\delta} \right] <\infty.$$
		\item \label{ass:B3} (Asymptotic stationarity) Let $ Q_t^\star $ be the marginal distribution of $ \Ddz_t $; then, $ N^{-1} \sum_{t=1}^N Q_t^\star $ converges weakly to some probability measure $ F $ on $ \mathcal{Z} $.
		\item \label{ass:B4} (Pointwise law of large numbers) For some metric $ \rho $ on $ \Phi $, let
		\begin{equation}\label{}
		\bar q(\ddz, \phi, \tau) := \sup_{\phi': \rho(\phi, \phi')<\tau} q(\ddz, \phi'), \quad \underline{q} (\ddz, \phi, \tau) := \inf_{\phi': \rho(\phi, \phi')<\tau} q(\ddz, \phi').
		\end{equation}
		For all $ \phi \in \Phi $, there exists a sequence of positive numbers $ \tau_i(\phi) $ such that $ \tau_i(\phi)\to0 $ as $ i\to\infty $, and such that for each $ \tau_i $ the random variables $ \bar q(\Ddz_t, \phi, \tau_i) $ and $ \underline q(\Ddz_t, \phi, \tau_i) $ satisfy a strong law of large numbers, i.e., as $ N\to\infty $:
		\begin{equation}\label{}
		\begin{aligned}
		&\frac{1}{N} \sum_{t=1}^N \left\{\bar q (\Ddz_t, \phi, \tau_i) - \E\left[\bar q (\Ddz_t, \phi, \tau_i)\right] \right\}\to0\\
		&\frac{1}{N} \sum_{t=1}^N \left\{\underline q (\Ddz_t, \phi, \tau_i) - \E\left[\underline q (\Ddz_t, \phi, \tau_i)\right] \right\}\to0,
		\end{aligned}
		\end{equation}
		where the two above equations hold with probability 1 with respect to $ (\Ddz_t)_t \sim Q^\star $.
	\end{enumerate}

	\begin{theorem}[Theorem 2 in \citealp{potscher1989uniform}] \label{Th:ULLN}
		If Assumptions \ref{ass:B1}, \ref{ass:B2}, \ref{ass:B3} and \ref{ass:B4} hold and if $ q(\ddz, \phi) $ is continuous on $ \mathcal{Z}\times \Phi $, then:
		\begin{enumerate}[label=(\roman*)]
			\item with probability 1 with respect to $ (\Ddz_t)_t \sim Q^\star $,
			\begin{equation}\label{}
			\lim_{t\to\infty} \sup_{\phi \in \Phi} \left|\frac{1}{N} \sum_{t=1}^N \left\{ q(\Ddz_t,\phi) - \E\left[ q(\Ddz_t,\phi) \right] \right\} \right| = 0;
			\end{equation}
			\item $ \int q(\ddz,\phi) dF(\ddz) $ exists and is finite, continuous on $ \Phi $ and, with probability 1 with respect to $ (\Ddz_t)_t \sim Q^\star $,
			\begin{equation}\label{}
			\lim_{t\to\infty} \sup_{\phi \in \Phi} \left|\frac{1}{N} \sum_{t=1}^N q(\Ddz_t,\phi) - \int q(\ddz,\phi) dF(\ddz)  \right| = 0;
			\end{equation}
		\end{enumerate}
	\end{theorem}

	We now give sufficient conditions for Assumption \ref{ass:B4} to hold. In fact, sequences for which the dependence of $ \Ddz_t $ on a past observation $ \Ddz_{t-m} $ decreases to 0 quickly enough as $ m\to\infty $ satisfy Assumption \ref{ass:B4}. This can be made more rigorous considering the definitions of $ \alpha $- and $ \varphi $-mixing sequences given in~Appendix~\ref{app:mixing}. 
	
	Given the sequence $ (\Ddz_t)_t$, for $ 1\le r\le s \le \infty $, define the Borel $ \sigma $-algebra of events generated from $( \Ddz_r, \Ddz_{r+1}, \ldots, \Ddz_{s-1}, \Ddz_s )$ as $ \mathcal{F}_r^s $. Then, we define
	the mixing coefficients for $ (\Ddz_t)_t $ as
	\begin{equation}\label{}
	\alpha^\Ddz(m) = \sup_{r\ge 1} \alpha_{Q^\star}(\mathcal{F}_{1}^r, \mathcal{F}_{r+m}^{+\infty}), \quad\varphi^\Ddz(m) = \sup_{r\ge 1} \varphi_{Q^\star}(\mathcal{F}_{1}^r, \mathcal{F}_{r+m}^{+\infty}).
	\end{equation}
	
	Similarly to before, the random sequence $ (\Ddz_t)_{t\in\mathbb{Z}} $ is said $ \alpha $-mixing if $ \alpha^\Ddz(m )\to 0$ as $ m\to\infty $ and $ \varphi $-mixing if $ \varphi^\Ddz(m )\to 0$ as $ m\to\infty $. Additionally, we say that the mixing coefficients $ \varphi^\Ddz(m) $ are of size $ s $ \citep{domowitz1982} if $ \varphi^\Ddz(m) = \mathcal{O}(m^{-\lambda}) $ for $\lambda>s$; similar definition can be given for the coefficients $ \alpha^\Ddz(m) $.

	Let us define now the following additional assumption: 
	\begin{enumerate}[label=\textbf{C\arabic*}]
		\setcounter{enumi}{3}
		\item \label{ass:B5} 
		Both conditions below hold:
		\begin{enumerate}
			\item (Mixing) Either one of the following holds: 
			\begin{enumerate}
				\item $ (\Ddz_t)_t $ is $ \alpha $-mixing with mixing coefficient of size $ r/(2r-1) $, with $ r \ge 1 $, or
				\item $ (\Ddz_t)_t $ is $ \varphi $-mixing with mixing coefficient of size $ r/(r-1) $ with $ r > 1 $.
			\end{enumerate}
			\item  (Moment boundedness) $ \sup_t \E\left[D(\Ddz_t)^{r+\delta}\right] <\infty$ for some $ \delta>0 $, for the value of $ r $ corresponding to the condition above which is satisfied.
		\end{enumerate}
	\end{enumerate}

	We give the following Lemma, which is contained in Corollary 1 in \cite{potscher1989uniform}.
	
	\begin{lemma}[Corollary 1 in \citealp{potscher1989uniform}]
		Assumption \ref{ass:B5} implies Assumptions \ref{ass:B2} and \ref{ass:B4}.	
	\end{lemma}
	
	We can therefore state the following.
	\begin{corollary}\label{Th:ULLN_2}
		If Assumptions \ref{ass:B1}, \ref{ass:B3} and \ref{ass:B5} hold and if $ q(\ddz, \phi) $ is continuous on $ \mathcal{Z}\times \Phi $, then the conclusions of Theorem~\ref{Th:ULLN} are satisfied.
	\end{corollary}
	
	\subsubsection{Proving Theorem~\ref{Th:consistency_main}}\label{app:proof_consistency}
	Here, we finally prove Theorem~\ref{Th:consistency_main} by combining the generic consistency result in Appendix~\ref{app:generic_consistency} with the uniform law of large number result reported in Appendix~\ref{app:ULLN}.
	
	Notice that, in stating Theorem~\ref{Th:ULLN} and Corollary~\ref{Th:ULLN_2}, we have considered a generic sequence $  (\Ddz_t)_t  $. In the setting of our interest, however, we want to study the prequential scoring rule defined in Eq.~\eqref{Eq:preq_SR_def}, and use Corollary~\ref{Th:ULLN_2} to state conditions under which Assumption~\ref{ass:A2}, and therefore Theorem~\ref{Th:consistency}, hold.
	
	To this aim, we identify now $N=T-k-l+1$, $\Ddz_t = \Ddobs_{t:t+k+l-1} $ and $ q(\Ddz_t, \phi) = S(P^\phi_{(l)}(\cdot|\Ddobs_{t:t+k-1}), \Ddobs_{t+k+l-1}) $; which  leads to
	\begin{equation}\label{}
	\begin{aligned}
	\frac{1}{N} \sum_{t=1}^N  q(\Ddz_t,\phi) &= \frac{1}{T-k-l+1} \sum_{t=1}^{T-k-l+1}  S(P^\phi_{(l)}(\cdot|\Ddobs_{t:t+k-1}),\Ddobs_{t+k+l-1}) \\
	&= \frac{1}{T-k-l+1} \sum_{t=k}^{T-l}  S(P^\phi_{(l)}(\cdot|\Ddobs_{t-k+1:t}),\Ddobs_{t+l})\\
	&=  S_T (P^\phi_{k+l:T}(\cdot|\Ddobsupkplusl), \DdobsupTfromk).
	\end{aligned}
	\end{equation}
	The distribution $ Q^\star $ on $ (\Ddz_t)_t  $ considered in the previous section is induced therefore by $ P^\star$ over $ (\Ddobs_t)_t $.

	We want now to relate $ \alpha^\Ddobs(m) $ and $ \varphi^\Ddobs(m) $ to $ \alpha^\Ddz(m) $ and $ \varphi^\Ddz(m) $; in order to do so, notice that, as $\Ddz_t = \Ddobs_{t:t+k+l-1} $, $ \mathcal{F}_r^s= \mathcal{G}_{r}^{s+k+l-1} $. Therefore, 
	\begin{equation}\label{}
	\begin{aligned}
	\alpha^\Ddz(m) &= \sup_{r\ge 1} \alpha^\Ddz(\mathcal{F}_{1}^r,\mathcal{F}_{r+m}^{+\infty}) = \sup_{r\ge 1} \alpha^\Ddobs(\mathcal{G}_{1}^{r+k+l-1}, \mathcal{G}_{r+m}^{+\infty}) \\
	&= \sup_{r\ge k+l} \alpha^\Ddobs(\mathcal{G}_{1}^{r}, \mathcal{G}_{r+m-k-l+1}^{+\infty}) \le \sup_{r\ge 1} \alpha^\Ddobs(\mathcal{G}_{1}^{r}, \mathcal{G}_{r+m-k-l+1}^{+\infty}) = \alpha^\Ddobs(m-k-l+1),
	\end{aligned}
	\end{equation}
	and, similarly, $ \varphi^\Ddz(m)\le \varphi^\Ddobs(m-k-l+1)$. As $ k $ is fixed, $ \varphi^\Ddobs(m) \to 0 \implies  \varphi^\Ddz(m) \to 0$ as $ m\to\infty $, which is to say, $ (\Ddobs_t)_t$ being $ \varphi $-mixing implies $ (\Ddz_t)_t $ is $ \varphi $-mixing as well, and similar for $ \alpha $-mixing. Additionally, if the mixing coefficients for $ (\Ddz_t)_t $ have a given size $ s $, then the mixing coefficients for $ (\Ddobs_t)_t $ will have the same size, and viceversa. In fact, $ \varphi^\Ddz(m) \le \varphi^\Ddobs(m-k-l+1) =\mathcal{O}(m^{-\lambda}) $ implies either $ \varphi^\Ddobs(m) =\mathcal{O}(m^{-\lambda}) $ or $ \varphi^\Ddobs(m)=o(m^{-\lambda}) $, and similar for $ \alpha $-mixing.

	We are now ready to prove Theorem~\ref{Th:consistency_main}.
	
	\begin{proof}[Proof of Theorem~\ref{Th:consistency_main}.]
		
		Notice that, by identifying $  \Ddz_t = \Ddobs_{t:t+k+l-1} $ and $ q(\Ddz_t, \phi) = S(P^\phi_{(l)}(\cdot|\ddobs_{t:t+k-1}), \ddobs_{t+k+l-1}) $, Assumption \ref{ass:A3} corresponds to Assumption~\ref{ass:B3}, and Assumption~\ref{ass:A4} implies Assumption~\ref{ass:B5}, due to the conservation of size of the mixing coefficients discussed above.	
		
		Together with Assumption~\ref{ass:B1} and the continuity condition, therefore, Corollary~\ref{Th:ULLN_2} holds, from which you have that, 
		with probability 1 with respect to $ (\Ddobs_t)_t\sim P^\star$,
		\begin{equation}\label{}
		\lim_{T\to\infty} \sup_{\phi \in \Phi} \left|\frac{1}{T-k-l+1} \sum_{t=k}^{T-l} \left\{ S(P^\phi_{(l)}(\cdot|\Ddobs_{t-k+1:t}), \Ddobs_{t+l}) - \E\left[ S(P^\phi_{(l)}(\cdot|\Ddobs_{t-k+1:t}), \Ddobs_{t+l}) \right] \right\} \right| = 0;
		\end{equation}
		which, recalling the definition of $ S_T(P_{k+l:T}^\phi(\cdot|\Ddobsupkplusl), \DdobsupTfromk) $ and $ S_T^\star(P_{k+l:T}^\phi) $ in Eqs.~\eqref{Eq:preq_SR_def} and \eqref{Eq:preq_SR_def_exp}, is the same as Assumption~\ref{ass:A2}. Thanks to this and Assumption~\ref{ass:A1}, therefore, Theorem~\ref{Th:consistency} holds, from which the result follows.
	\end{proof}

	\section{More details on the different methods}\label{app:more_details}
	
	\subsection{Training generative networks via divergence minimization}\label{app:GAN}

	\subsubsection{$ f $-GAN}\label{app:f-GAN}
	The $ f $-GAN approach is defined by considering an $ f $-divergence in place of $ D $ in Eq.~\eqref{Eq:div_min} in the main text
	\begin{equation}\label{key}
	D_f(P^\star||P^\phi) = \int_{\mathcal{Y}} p^\phi(\ddobs) f\left(\frac{p^\star(\ddobs)}{p^\phi(\ddobs)}\right)d\mu(\ddobs),
	\end{equation}
	where $ f:\mathbb{R}_+\to \R $  is a convex, lower-semicontinuous function for which $ f(1)=0 $, and where $ p^\phi $ and $ p^\star $ are densities of $ P^\phi $ and $ P^\star $ with respect to a base measure $ \mu $. Let now $ \operatorname{dom}_f $ denote the domain of $ f $. By exploiting the Fenchel conjugate $ f^*(t) = \sup_{u \in \operatorname{dom}_f } \left\{ut - f(u)\right\}$, \cite{nowozin2016f} obtain the following variational lower bound
	\begin{equation}\label{}
	D_f(P^\star||P^\phi)	\ge \sup_{c \in \mathcal{C}} \left(\mathbb{E}_{\Ddobs \sim P^\star}c(\Ddobs)-\mathbb{E}_{\Ddsim \sim P^\phi}f^{*}(c(\Ddsim))\right),
	\end{equation}
	which holds for any set of functions $ \mathcal{C} $ from $\Y $ to $ \operatorname{dom}_{f^*}$.
	By considering a parametric set of functions $ \mathcal{C} =\{ c_\psi:\mathcal{Y} \to  \operatorname{dom}_{f^*}, \psi \in \Psi\} $, a surrogate to the problem in Eq.~\eqref{Eq:div_min} in the main text becomes:
	\begin{eqnarray}\label{}
	\min_\phi \max_\psi  \left(\mathbb{E}_{\Ddobs \sim P^\star}c_\psi(\Ddobs)-\mathbb{E}_{\Ddsim \sim P^\phi}f^{*}(c_\psi(\Ddsim))\right).
	\end{eqnarray}
	
	In the conditional setting discussed in Section~\ref{sec:GAN_via_div} in the main text, the above generalizes to
	\begin{equation}\label{Eq:f-gan-cond2}
	\min_\phi \max_\psi  \E_{ \ddtheta \sim \Pi}\big(\mathbb{E}_{\Ddobs \sim P^\star(\cdot|\ddtheta)}c_\psi(\Ddobs; \ddtheta)\\
	-\mathbb{E}_{\Ddobs \sim P^\phi(\cdot|\ddtheta)}f^{*}(c_\psi(\Ddobs;\ddtheta))\big),
	\end{equation}
	By denoting as $ P_{\ddtheta, \Ddobs}^\star$ and $P_{\ddtheta, \Ddobs}^\phi $ the joint distributions over $ \Theta\times\Y $, Eq.~\eqref{Eq:f-gan-cond2} corresponds to the relaxation of $ D_f(P_{\ddtheta, \Ddobs}^\star||P_{\ddtheta, \Ddobs}^\phi) $ under the constraint that the marginal of $ P_{\ddtheta, \Ddobs}^\phi $ for $ \ddtheta $ is equal to $ \Pi $.
	
	In order to solve the problem in Eq.~\eqref{Eq:f-gan-cond2}, alternating optimization over $ \phi $ and $ \psi $ can be performed; in Algorithm \ref{alg:cGAN}, we show a single epoch (i.e., a loop on the full training data set) of conditional \textit{f}-GAN training; for simplicity, we consider here using a single pair $ (\ddtheta_i, \ddobs_i) $ to estimate the expectations in Eq.~\eqref{Eq:f-gan-cond2} (i.e., the batch size is 1), but using a larger number of samples is indeed possible. Notice how in Algorithm \ref{alg:cGAN} we update the critic once every generator update; however, multiple critic updates can be done.

	\begin{algorithm}
		\caption{Single epoch conditional \textit{f}-GAN training.}
		\label{alg:cGAN}
		\begin{algorithmic}
			\REQUIRE Parametric map $ h_\phi $, critic network $ c_\psi $, learning rates $ \epsilon $, $ \gamma $.
			\FOR{each training pair $( \ddtheta_i, \ddobs_i)$}
			\STATE Sample $ \mathbf{z} \sim Q$ 
			\STATE Obtain $ \hat \ddsim^{\phi}_i = h_\phi (\mathbf{z}, \ddtheta_i) $
			\STATE Set
			$ \psi \leftarrow \psi + \gamma \cdot \nabla_\psi \Big[ c_\psi(\ddobs_i, \ddtheta_i) -f^*(c_\psi(\hat \ddsim^{\phi}_i, \ddtheta_i)) \Big] $
			\STATE Set $\phi \leftarrow\phi - \epsilon \cdot \nabla_\phi \Big[ -f^*(c_\psi(\hat \ddsim^{\phi}_i, \ddtheta_i)) \Big] $
			\ENDFOR
		\end{algorithmic}
	\end{algorithm}

	\subsubsection{Wasserstein-GAN (WGAN)}\label{app:WGAN}
	\cite{arjovsky2017wasserstein} exploited the following expression for the 1-Wasserstein distance
	\begin{equation}\label{Eq:WGAN}
	W\left({P}^{\star}, {P}^{\phi}\right)=\sup _{c: \|c\|_{L} \leq 1} \mathbb{E}_{\Ddobs \sim \mathbb{P}^{\star}}[c(\Ddobs)]-\mathbb{E}_{\Ddsim \sim \mathbb{P}^{\phi}}[c(\Ddsim)],
	\end{equation}
	where $ ||c||_L  $ denotes the Lipschitz constant of the function $ c $. The different notation here highlights how $ W $ is a symmetric function. Plugging Eq.~\eqref{Eq:WGAN} into Eq.~\eqref{Eq:div_min}  in the main text leads again to an adversarial setting; here, the Lipschitz constraint can be enforced by clipping the weights of the neural network to a given range \citep{arjovsky2017wasserstein}. Alternatively, this hard constraint can be relaxed to a soft one via gradient penalization \citep{gulrajani2017improved}.

	\subsubsection{MMD-GAN}		\label{app:MMDGAN}

	A specific case of the MMD (Eq.~\ref{Eq:MMD}  in the main text) is the Energy Distance
	\begin{equation}\label{Eq:energy_d}
	\begin{aligned}
	\mathcal{E} \left({P}^{\star}, {P}^{\phi}\right) &= \E\left[ 2 ||\Ddsim- \Ddobs||_2^\beta - ||\Ddsim-  \Ddsim^{\prime}||_2^\beta-||\Ddobs - \Ddobs^{\prime}||_2^\beta\right],
	\end{aligned}
	\end{equation}
	where $ \beta \in (0,2) $ and $ ||\cdot||_2 $ denotes the $ \ell_2 $ norm. In \cite{bellemare2017cramer}, the above is used to define an algorithm to train generative networks, termed Cramer-GAN.
	
	In \cite{li2017mmd}, the authors proposed to compute the kernel $ k $ in Eq.~\eqref{Eq:MMD}  in the main text on a learnable transformation $ c_\psi $, whose weights are trained to maximize the discrepancy. Specifically, that leads to a new discrepancy measure
	\begin{equation}\label{}
	\max_\psi \E\big[k\left(c_\psi(\Ddsim), c_\psi(\Ddsim^{\prime})\right)-2 k(c_\psi(\Ddsim), c_\psi(\Ddobs))
	+k\left(c_\psi(\Ddobs), c_\psi(\Ddobs^{\prime})\right)\big],
	\end{equation} 
	which is a meaningful divergence between probability distributions \citep{li2017mmd}. 
	In this setting, again people resort to alternating maximization steps over $ \psi $ with minimization over $ \phi $. This, as mentioned in the main text, leads to biased estimates of gradients. However, for MMD-GANs, training is made easier by applying the gradient regularization techniques described in \cite{gulrajani2017improved}, as shown in \cite{binkowski2018demystifying}.

	Notice that, in minimizing Equations~\eqref{Eq:MMD} in the main text with respect to $ \phi $, one could ignore the term involving $ \Ddobs, \Ddobs' $; however, when introducing $ c_\psi $, this cannot be done as that term depends on $ \psi  $ as well.
	
	In the conditional setting, a natural approach for MMD-GAN is minimizing \\$\E_{ \ddtheta \sim \Pi} [\operatorname{MMD}^2 \left({P}^{\star}(\cdot|\ddtheta ), {P}^{\phi}(\cdot|\ddtheta ) \right) ] $, as $ \operatorname{MMD}^2(P_{\ddtheta, \Ddobs}^\star, P_{\ddtheta, \Ddobs}^\phi) $ would require computing kernel over $ \Theta\times \Y $. 
	
	Notice however how, in estimating $ \operatorname{MMD}^2 \left({P}^{\star}(\cdot|\ddtheta ), {P}^{\phi}(\cdot|\ddtheta )\right)$, multiple samples $ \Ddobs,\Ddobs'\sim {P}^{\star}(\cdot|\ddtheta) $ are used (see Eq.~\ref{Eq:MMD} in the main text), but those are unavailable (empirical samples are of the form in Eq.~\ref{Eq:joint_data} in the main text); as discussed before, however, $ k(\Ddobs,\Ddobs') $ does not depend on $ \phi $, so that it can be discarded in the minimization process.
	However, if the data is transformed via $ c_\psi $, $ k(c_\psi(\Ddobs), c_\psi(\Ddobs')) $ cannot be dropped anymore, which makes the problem intractable. In \cite{bellemare2017cramer}, this problem is solved by replacing $ k(c_\psi(\Ddobs), c_\psi(\Ddobs'))  $ with some other tractable terms; however, that approach leads to an ill-defined statistical divergence, as it can be minimized by two distributions which are not the same \citep{binkowski2018demystifying}.

	\subsection{Scoring Rules}\label{app:scores}
	We now introduce some common SRs; let $ \Ddsim,  \Ddsim' \sim P^\phi $ be independent samples for the forecast distribution $ P^\phi $.
	
	\subsubsection{Energy Score}
	For $ \beta \in (0,2) $, the energy score is
	\begin{equation}\label{Eq:eng_score}
	\SE^{(\beta)}(P^\phi, \ddobs) = 2 \cdot \E \| \Ddsim - \ddobs\|_2^\beta - \E\|\Ddsim- \Ddsim'\|_2^\beta.
	\end{equation}
	
	The probabilistic forecasting literature \citep{gneiting2007strictly} use a different convention of the energy score and the subsequent kernel score, which amounts to multiplying our definitions by $ 1/2 $. We follow here the convention used in the statistical inference literature \citep{rizzo2016energy, cherief2020mmd, nguyen2020approximate}
	
	The Energy Score is strictly proper for the class of probability measures $ P^\phi $ such that $ \E_{\Ddsim\sim P^\phi}\|\Ddsim\|^\beta < \infty $ \citep{gneiting2007strictly}. The Energy Score is related to the Energy distance (Eq.~\eqref{Eq:energy_d}), which is a metric between probability distributions \citep{rizzo2016energy}. We will fix $ \beta=1$ in the rest of this work. Additionally, for a univariate distribution and $ \beta=1 $, the Energy Score recovers the Continuous Ranked Probability Score (CRPS), widely used in meteorology (e.g, see \citealp{hersbach2000decomposition}).

	\subsubsection{Kernel Score}
	For a positive definite kernel $ k(\cdot, \cdot) $, the kernel Scoring Rule can be defined as \citep{gneiting2007strictly} \begin{equation}\label{Eq:kernel_score2}
	S_k(P^\phi, \ddobs) = \E[k(\Ddsim,\Ddsim')] - 2\cdot\E [k(\Ddsim, \ddobs)].
	\end{equation}
	The Kernel Score is connected to the squared Maximum Mean Discrepancy (MMD, \citealp{gretton2012kernel}) relative to the kernel $ k $, see Eq.~\eqref{Eq:MMD} in the main text.
	$ S_k $ is proper for the class of probability distributions for which $ \E[k(\Ddsim,\Ddsim')] $ is finite (by Theorem 4 in \citealp{gneiting2007strictly}). Additionally, it is strictly proper under conditions on $ k $ ensuring that the MMD is a metric for probability distributions on $ \Y $ \cite{gretton2012kernel}. These conditions are satisfied, among others, by the Gaussian kernel (which we will use in this work)
	\begin{equation}\label{Eq:gau_k}
	k(\ddsim, \ddobs)=\exp \left(-\frac{\|\ddsim-\ddobs\|_{2}^{2}}{2 \gamma^{2}}\right),
	\end{equation}
	in which $ \gamma $ is a scalar bandwidth. 
	
	\FloatBarrier
	\subsubsection{Patched Score}
	\FloatBarrier
	
	For the Patched Score, we consider different overlapping patches of the input data; denote as $ \mathcal{P} $ the set of patches and as $ p \in \mathcal{P} $ an individual patch; the patches are of a given size and spaced by a given spacing. 
	
\begin{figure}
		\centering
		\includegraphics[width=0.4\columnwidth]{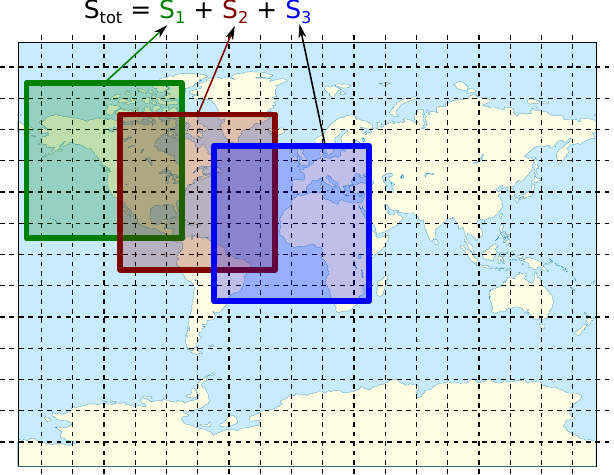}
		\caption{Patched SR: a SR for multivariate data is computed on localized patches, and the resulting values are summed.}
		\label{fig:patched}
	\end{figure}

	Then, we compute a SR $ S $ for multivariate distributions on each patch separately, and then add the results
	\begin{equation}\label{Eq:patched_SR}
	S_{p}(P^\phi, \ddobs) = \sum_{p \in \mathcal{P}} S(P^\phi|_p, \ddobs|_p),
	\end{equation}
	where $ \ddobs|_p$ denotes the components of $ \ddobs $ in the patch $ p $ and $ P^\phi|_p $ denotes the marginal distribution induced by $ P^\phi $ for components in the patch $ p $. See Figure~\ref{fig:patched} for a representation. As mentioned in the main body (Sec.~\ref{sec:SR_spatial} in the main text), the resulting SR is not strictly proper, as far away correlations are discarded. Notice how the topology of data for our global weather data set is periodic along the longitudinal direction (i.e., horizontally in Figure~\ref{fig:patched}). The patches we define follow this.

	\FloatBarrier
	\section{Stochastic Gradient Descent for generative-SR networks}\label{app:unbiased_estimates}
	
	We discuss here how we can get unbiased gradient estimates for the prequential SR in Eq.~\eqref{Eq:preq_SR} in the main text with respect to the parameters of the generative network $ \phi $. 
	
	In order to do that, we first discuss how to obtain unbiased estimates of the SRs we use across this work. Then, we show how those allow to obtain unbiased gradient estimates.

	\subsection{Unbiased scoring rule estimates}\label{app:unbiased_SR}
	
	Consider we have draws $\mathbf x_j\sim P, j=1, \ldots, m $.
	
	\subsubsection{Energy Score}
	An unbiased estimate of the energy score can be obtained by unbiasedly estimating the expectations in $\SE^{(\beta)}(P, \ddobs) $ in Eq.~\eqref{Eq:eng_score}
	\begin{equation}
	\hat S_{\text{E}}^{(\beta)}(\{\mathbf x_j\}_{j=1}^m, \ddobs) =\frac{2}{m} \sum_{j=1}^m \left\| \ddsim_j - \ddobs\right\|_2^\beta - \frac{1}{m(m-1)}\sumjk \left\|\ddsim_j-\ddsim_k\right\|_2^\beta.
	\end{equation}
	
	\subsubsection{Kernel Score} Similarly to the energy score, we obtain an unbiased estimate of $ S_k(P,\dobs) $ by
	\begin{equation}
	\hat S_k(\{\mathbf x_j\}_{j=1}^m, \ddobs) = \frac{1}{m(m-1)}\sumjk  k(\ddsim_j,\ddsim_k )-\frac{2}{m} \sum_{j=1}^m k(\ddsim_j,\ddobs).
	\end{equation}
	
	\subsubsection{Variogram Score} It is immediate to obtain an unbiased estimate of $\Sv^{(p)}(P, \ddobs) $ in Eq.~\eqref{Eq:var_score} in the main text by
	\begin{equation}\label{key}
	\hat \Sv^{(p)}(\{\mathbf x_j\}_{j=1}^m, \ddobs)=\sum_{i, j=1}^{d} w_{i j}\left(\left|y_{i}-y_{j}\right|^{p}-\frac{1}{m}\sum_{k=1}^m\left|x_{k,i}-x_{k,j}\right|^{p}\right)^{2}.
	\end{equation}
	
	\subsubsection{Patched SR}
	Assume the patched SR in Eq.~\eqref{Eq:patched_SR} is built from a SR $ S $ which admits an unbiased empirical estimate $ \hat S(\{\mathbf x_j\}_{j=1}^m, \ddobs) $. Therefore, an unbiased estimate of the patched SR can be obtained as
	\begin{equation}\label{}
	\hat S_{p}(\{\mathbf x_j\}_{j=1}^m, \ddobs) = \sum_{p \in \mathcal{P}} S(\{\mathbf x_j|_p\}_{j=1}^m, \ddobs|_p),
	\end{equation}
	as in fact the components of samples $ \mathbf x_j $ in the patch $ p $ are samples from the marginal distribution over the patch $ P|_p $.

	\subsubsection{Sum of SRs}	
	When adding multiple SRs, an unbiased estimate of the sum can be obtained by adding unbiased estimates of the two addends.

	\subsection{Unbiased estimate for gradient of $ S_T $}\label{app:unbiased_estimate_obj}

	Recall now we want to solve:	
	\begin{gather}\label{Eq:argmin_problem}
	\hat \phi_T(\ddobsupT) :=\argmin_\phi S_T (P^\phi_{k+l:T}(\cdot|\ddobs_{1:k+l-1}), \ddobs_{k+l:T}),
	\end{gather}
	where, for simplicity, we re-define $ S_T $ in Eq.~\eqref{Eq:preq_SR} in the main text with an additional scaling constant: 	
	\begin{equation}\label{Eq:preq}
	S_T (P^\phi_{k+l:T}(\cdot|\ddobs_{1:k+l-1}), \ddobs_{k+l:T}):=	\frac{1}{T-l-k+1}	\sum\limits_{t=k}^{T-l}S(P^\phi_{t+l}(\cdot|\ddobs_{t-k+1:t}),\ddobs_{t+l}).
	\end{equation}

	In order to do this, we exploit Stochastic Gradient Descent (SGD), which requires unbiased estimates of $  S_T (P^\phi_{k+l:T}(\cdot|\ddobs_{1:k+l-1}), \ddobs_{k+l:T}) $ (notice we are not talking here of unbiased estimates with respect to the observed sequence $\ddobsupT $).

	Notice how, for all the Scoring Rules used across this work, as well as any weighted sum of those, we can write:
	$ S(P,\ddobs) = \E_{\Ddobs, \Ddobs'\sim P}\left[g(\Ddobs, \Ddobs',\ddobs)\right] $ for some function $ g $; namely, the SR is defined through an expectation over (possibly multiple) samples from $ P $. That is the form exploited in Appendix~\ref{app:unbiased_SR} to obtain unbiased SR estimates.
	
	Now, we will use this fact to obtain unbiased estimates for the objective in Eq.~\eqref{Eq:preq}. For brevity, let us now denote $ J(\phi ) = 	S_T (P^\phi_{k+l:T}(\cdot|\ddobs_{1:k+l-1}), \ddobs_{k+l:T}) $, which we can rewrite as (letting	$ N =T-l-k+1 $ for brevity)
	\begin{equation}\label{Eq:J2}
	\begin{aligned}
	J(\phi)&=\frac{1}{N}\sum\limits_{t=k}^{T-l} \E_{\Ddobs, \Ddobs'\sim P^\phi(\cdot|\ddobs_{t-k+1:t})}\left[g(\Ddobs, \Ddobs',\ddobs_{t+l})\right] \\
	&= \frac{1}{N}\sum\limits_{t=k}^{T-l} \E_{\mathbf{Z},\mathbf{Z}'\sim Q}\left[g(h_\phi(\mathbf{Z};\ddobs_{t-k+1:t}),h_\phi(\mathbf{Z}';\ddobs_{t-k+1:t}),\ddobs_{t+l})\right],
	\end{aligned}
	\end{equation}
	where we used the fact that $ P^\phi$ is the distribution induced by a generative network with transformation $ h_\phi$; this is called the reparametrization trick \cite{kingma2013auto}. Now
	\begin{equation}\label{Eq:J_grad}
	\begin{aligned}
	\nabla_\phi J(\phi)&= \nabla_\phi \frac{1}{N}\sum\limits_{t=k}^{T-l} \E_{\mathbf{Z},\mathbf{Z}'\sim Q}\left[g(h_\phi(\mathbf{Z};\ddobs_{t-k+1:t}),h_\phi(\mathbf{Z}';\ddobs_{t-k+1:t}),\ddobs_{t+l})\right] \\
	&= \frac{1}{N}\sum\limits_{t=k}^{T-l} \E_{\mathbf{Z},\mathbf{Z}'\sim Q}\left[\nabla_\phi  g\left(h_\phi(\mathbf{Z};\ddobs_{t-k+1:t}),h_\phi(\mathbf{Z}';\ddobs_{t-k+1:t}),\ddobs_{t+l}\right)\right].
	\end{aligned}
	\end{equation}
	In the latter equality, the exchange between expectation and gradient is not a trivial step, due to the non-differentiability of functions (such as ReLU) used in $ h_\phi $. Luckily, Theorem 5 in \cite{binkowski2018demystifying} proved the above step to be valid almost surely with respect to a measure on $ \Phi $, under mild conditions on the NN architecture.
	
	We can now easily obtain an unbiased estimate of the above.	
	Additionally, Stochastic Gradient Descent 
	usually consider a small batch of training samples, obtained by considering a random subset $ \mathcal{T} \subseteq \{ k, k+1 \ldots, n-l-1, n-l \}$. Therefore, the following unbiased estimator of $ \nabla_\phi J(\phi) $ can be obtained, with samples $ \mathbf z_{t,j}\sim Q, j=1,\ldots,m $
	\begin{equation}\label{}
	\widehat {\nabla_\phi J(\phi)}= \frac{1}{|\mathcal{T}|}\sum\limits_{t\in \mathcal{T}}  \frac{1}{m(m-1)} \sumij  \nabla_\phi g(h_\phi(\mathbf z_{t,i};\ddobs_{t-k+1:t}),h_\phi(\mathbf z_{t,j};\ddobs_{t-k+1:t}),\ddobs_{t+l}).
	\end{equation}
	In practice, we then use autodifferentiation libraries (see for instance \citealp{pytorch}) to compute the gradients in the above quantity.

	In Algorithm~\ref{alg:gen-SR-conditional}, we train a generative network for a single epoch using a scoring rule $ S $ for which unbiased estimators can be obtained by using more than one sample from $ P^\phi $.
	As in Algorithm~\ref{alg:cGAN}, we use a single pair $ (\ddtheta_i, \ddobs_i) $ to estimate the gradient.

	\begin{algorithm}
		\caption{Single epoch generative-SR training.}
		\label{alg:gen-SR-conditional}
		\begin{algorithmic}
			\REQUIRE Parametric map $ h_\phi $, SR $ S $, learning rate $ \epsilon $.
			\FOR{each training pair $( \ddtheta_i, \ddobs_i)$}
			\STATE Sample {\textbf{multiple}} $ \mathbf z_1, \ldots,\mathbf z_m $ 
			\STATE Obtain $ \hat \ddsim_{i,j}^\phi = h_\phi (\mathbf z_j, \ddtheta_i)$
			\STATE Obtain unbiased estimate $ \hat S (P^\phi(\cdot| \ddtheta_i), \ddobs_i)$ from $ \hat \ddsim_{i,j}^\phi$
			\STATE Set $\phi \leftarrow \phi - \epsilon \cdot \nabla_\phi \hat  S (P^\phi(\cdot| \ddtheta_i), \ddobs_i) $
			\ENDFOR
		\end{algorithmic}		
	\end{algorithm}	
	
	\section{Performance measures for probabilistic forecast} \label{app:metric}
	
	\subsection{Deterministic performance measures}
	
	We discuss two measures of performance of a deterministic forecast $ \hat y_{t+l} $ for a realization $ y_{t+l} $; across our work, we take $ \hat y_{t+l}  $ to be the mean of the probability distribution $ \probfor $.

	\subsubsection{Normalized RMSE}
	We first introduce the Root Mean-Square Error (RMSE) as
	$$\operatorname{RMSE} =\sqrt{\frac{1}{N}\sum\limits_{t=1}^{N}\left(\hat y_{t+l} - y_{t+l} \right)^{2}},$$
	where we consider here for simplicity $ t=1, \ldots, N $. From the above, we obtain the Normalized RMSE (NRMSE) as
	$$\operatorname{NRMSE} = \frac{RMSE}{\max_t\{ y_{t+l} \} - \min_t\{ y_{t+l} \}}.$$
	$ \operatorname{NRMSE} =0 $ means that $ \hat y_{t+l} = y_{t+l} $ for all $ t $'s.
	
	\subsubsection{Coefficient of determination}
	The coefficient of determination $\operatorname{R}^2 $ measures how much of the variance in $ \{y_{t+l}\}_{t=1}^N $ is explained by $ \{\hat y_{t+l}\}_{t=1}^N $. Specifically, it is given by
	$$
	\operatorname{R}^{2}=1-\frac{\sum_{t=1}^{N} \left(y_{t+l}-\hat y_{t+l}\right)^{2}}{\sum_{t=1}^{N} \left(y_{t+l}-\bar{y} \right)^{2}},
	$$
	where $\bar{y} = \frac{1}{N} \sum\limits_{t=1}^{N} y_{t+l}$. $ R^2 \le 1$ and, when $ \operatorname{R}^{2}=1 $, $ \hat y_{t+l} = y_{t+l} $ for all $ t $'s. Notice how $ R^2 $ is unbounded from below, and can thus be negative.

	\subsection{Calibration error}
	
	We review here a measure of calibration of a probabilistic forecast; this measure considers the univariate marginals of the probabilistic forecast distribution $ \probfor $; for component $ i $, let us denote that by $ P_{\phi,i}(\cdot|\ddobs_{t-k+1:t}) $.

	The calibration error \citep{radev2020bayesflow} quantifies how well the credible intervals of the probabilistic forecast $ P_{\phi,i}(\cdot|\ddobs_{t-k+1:t}) $ match the distribution of the verification $ \Dobs_{t+l,i} $. Specifically, let $ \alpha^\star(i) $ be the proportion of times the verification $ \dobs_{t+l,i} $ falls into an $ \alpha $-credible interval of $ P_{\phi,i}(\cdot|\ddobs_{t-k+1:t}) $, computed over all values of $ t $. If the marginal forecast distribution is perfectly calibrated for component $ i $, $ \alpha^\star(i) =\alpha$ for all values of $ \alpha \in (0,1) $. 
	
	We define therefore the calibration error as the median of $ |\alpha^\star(i) -\alpha |$ over 100 equally spaced values of $ \alpha\in(0,1) $. Therefore, the calibration error is a value between $ 0 $ and 1, where $ 0 $ denotes perfect calibration. 
	
	In practice, the credible intervals of the predictive are estimated using a set of samples from $ \probfor $.

	\section{Additional experimental details}\label{app:exp_details}

	\subsection{Tuning $ \gamma $ in the Gaussian kernel}\label{app:gamma}
	
	Similar to what was suggested for instance in \cite{park2016k2}, we set $\gamma$ in the Gaussian kernel in Eq.~\eqref{Eq:gau_k} to be the median of the pairwise distances $||\ddobs_i- \ddobs_j||$ over all pairs of observations $\ddobs_i, \ddobs_j, i\neq j$ in the validation window.
	
	\subsection{Lorenz63 model}\label{app:Lorenz63}
	
	\subsubsection{Model definition}\label{app:Lorenz63_model}
	
	The Lorenz63 model \citep{lorenz1963deterministic} is defined by the following differential equations
	\begin{equation}\label{key}
	\begin{aligned}
	\frac{\mathrm{d}x}{\mathrm{d}t} &= \sigma (y - x), \\[6pt]
	\frac{\mathrm{d}y}{\mathrm{d}t} &= x (\rho - {z}) - y, \\[6pt]
	\frac{\mathrm{d}{z}}{\mathrm{d}t} &= x y - \beta z.
	\end{aligned} 
	\end{equation}
	
	To generate our data set, we consider $ \sigma=10 $, $ \rho=28 $, $ \beta=2.667 $ and integrate the model using Euler scheme with $ dt = 0.01 $ starting from $ x=0, y=1, {z}=1.05 $. We discard the first 10 time units and integrate the model for additional 9000 time units, during which we record the value of $ y $ every $ \Delta t = 0.3 $ and discard the values of $ x $ and $ z $.

	\subsubsection{Neural Networks architecture}\label{app:Lorenz63_nets}
	
	We experiment with Recurrent Neural Networks (RNNs), which capture the temporal structure in the data.
	
	For the generative network, the observation window is passed through a Gated Recurrent Units (GRU, \citealp{cho2014properties}) layer with depth 1 and hidden size 8 or 16 (that is a tuning hyperparameter, the choice of which we discuss below). The output of the GRU layer is then concatenated to the latent variable $ \mathbf{Z} $ with size $ 1 $ and passed through 3 fully connected layers, which output a forecast for the next timestep. For the deterministic setting trained with the regression loss, the architecture is analogous, the only difference being that no latent variable $ \mathbf Z $ is concatenated to the output of the GRU layer.
	
	In the adversarial settings, the critic has a GRU layer with depth 1 that, analogously to the generative net, processes the information in the past observation window. As above, we try hidden sizes 8 and 16. Then, the output of the GRU layer and the observation/forecast are concatenated and transformed by 3 fully connected layers. In the GAN case, the critic outputs a value between 0 and 1 indicating how confident the critic believes that is a fake sample. In the WGAN-GP case, the critic output is a real number.

	\subsubsection{Training hyperparameters}\label{app:Lorenz63_hyperparameters}
	
	For the experiments on Lorenz63, we considered the batch size to be 1000. For the SR and deterministic approaches, we used Adam optimizer and tested the following learning rate values: $ 10^{-i} $ for $ i=1,\ldots, 6 $ for the SR methods and $ 10^{-i-1} $ and $ 3\cdot 10^{-i - 1} $ for $ i=1,\ldots, 3 $ for regression.
	We fix the GRU hidden size to 8. 
	We report then the performance achieved with the learning rate yielding lower loss on the validation set, which is indicated in Table~\ref{tab:lr_lorenz63_SR}.
	%
	\begin{table}
		\begin{center}
			\begin{tabular}{cccc}
				\toprule
				Energy & Kernel & Energy-Kernel & Regression \\
				\midrule
				0.01  & 0.001 & 0.01 & 0.001 \\
				\bottomrule
			\end{tabular}
		\end{center}
		\caption{Optimal learning rate values for SR and regression (deterministic) approaches for Lorenz63.}
		\label{tab:lr_lorenz63_SR}
	\end{table}

	For the GAN and WGAN-GP approach, we used Adam optimizer and we tested the following learning rate values for both critic and generative network: $ 10^{-i} $ and $ 3\cdot 10^{-i} $ for $ i=1,\ldots, 7$. In total, those are 14 learning rate values. 
	We tested GRU hidden size to 8 and 16; further, we experiment with 4 number of critic training steps for WGAN-GP (1, 3, 5, 10), in order to have the best possible results to compare with our SR methods, while we left the number of critic training steps to 1 for GAN. Overall, therefore, we had $ 2\cdot 14^2  = 392$ experiments for GAN and  $ 2\cdot 4 \cdot 14^2  = 1568$ for WGAN-GP; notice the extremely larger number number of experiments for the adversarial approaches with respect to SR ones, which highlights an advantage of our approach. We stress that such a number of trials could be possible only for the low-dimensional setting of the Lorenz63 and Lorenz96 models, in which training is cheap, but not in real-life applications.
	
	Additionally, the adversarial approaches do not allow to select hyperparameters according to loss on a validation set, as the generator loss depends on the current state of the discriminator (i.e., there is no absolute loss scale). Therefore, we report results for 3 different configurations for GAN and WGAN-GP, maximizing either deterministic performance (1) or calibration (2), or striking the best balance between these two (3). The resulting learning rates are in Table~\ref{tab:lr_lorenz63_GAN}.
	
	%

	\begin{table}
		\begin{center}
			\begin{adjustbox}{max width=\textwidth}
				\begin{tabular}{lcccccc}
					\toprule
					& GAN (1) & GAN (2) & GAN (3) & WGAN-GP (1) & WGAN-GP (2) & WGAN-GP (3) \\
					\midrule
					Generator  l.r. & 0.0003 & 0.001 & 0.0001 & 0.003& 0.0003 & 0.0003 \\
					Critic l.r. & 0.03 & 0.01 & 0.001 & 0.001 &0.1&0.03 \\
					GRU hidden size & 16 & 8 & 8& 8 & 8 & 8 \\
					Critic training steps & 1 & 1& 1 & 5 & 5& 5 \\
					\bottomrule
				\end{tabular}
			\end{adjustbox}
		\end{center}
		\caption{Optimal hyperparameter values for adversarial approaches for Lorenz63 model.}
		\label{tab:lr_lorenz63_GAN}
	\end{table}

	\subsection{Lorenz96 model}\label{app:Lorenz}

	\subsubsection{Model definition}\label{app:Lorenz_model}

	The Lorenz96 model \citep{lorenz1996predictability} is a toy representation of atmospheric behavior containing slow ($ \ddsim $) and fast ($ \ddobs $) evolving variables. 
	
	Specifically, the evolution of the variables is determined by the following differential equations
	\begin{equation*}
	\begin{aligned}
	\frac{\mathrm{d}x_k}{\mathrm{d}t} &= - x_{k-1}(x_{k-2}-x_{k+1}) -x_k + F -\frac{hc}{b} \sum_{j=J(k-1)+1}^{kJ} y_j;\\
	\frac{\mathrm{d}y_j}{\mathrm{d}t} &= - cb y_{j+1}(y_{j+2}-y_{j-1}) -c y_j + \frac{hc}{b} X_{\text{int}[(j-1)/J]+1},
	\end{aligned}
	\end{equation*}	
	where $ k=1, \ldots, K, $ and $ j=1, \ldots, JK$, and cyclic boundary conditions are assumed, so that index $ k=K+1 $ corresponds to $ k=1 $ and similarly for $ j $. The above equations connect the fast and slow variables in a cyclic way. Additionally, $ x_k $ reciprocally depends on $ J $ fast variables.

	Following \cite{gagne2020machine}, we take $ K = 8 $, $ J=32 $, $ h = 1 $, $ b = 10 $, $ c = 10 $ and $ F = 20 $. We then integrate the above equations with RK4 scheme with $ dt = 0.001 $, starting from $ x_k = y_j=0 $ for $ k=2, \ldots, K $ and $ j=2, \ldots JK $ and $ x_1=y_1=1 $. We discard the first 2 time units and record the values of $ \mathbf x $ every $ \Delta t = 0.2 $ (which corresponding to roughly one atmospheric day with respect to predictability, \citealp{gagne2020machine}). We do this for additional 4000 time units, and split the resulting data set in training, validation and test according to the proportions $ 60\% $, $ 20\% $ and $ 20\% $.

	\subsubsection{Neural Networks architecture}\label{app:Lorenz_nets}

	We experiment with Recurrent Neural Networks (RNNs), which capture the temporal structure in the data.
	

	For the generative network, the observation window is passed through a Gated Recurrent Units (GRU, \citealp{cho2014properties}) layer with depth 1 and hidden size 32 or 64 (that is a tuning hyperparameter, the choice of which we discuss below). The output of the GRU layer is then concatenated to the latent variable $ \mathbf{Z} $ with size $ 1 $ and passed through 3 fully connected layers, which output a forecast for the next timestep. For the deterministic setting trained with the regression loss, the architecture is analogous, the only difference being that no latent variable $ \mathbf Z $ is concatenated to the output of the GRU layer.
	
	In the adversarial settings, the critic has a GRU layer with depth 1 that, analogously to the generative net, processes the information in the past observation window. As above, we try hidden sizes 8 and 16. Then, the output of the GRU layer and the observation/forecast are concatenated and transformed by 3 fully connected layers. In the GAN case, the critic outputs a value between 0 and 1 indicating how confident the critic believes that is a fake sample. In the WGAN-GP case, the critic output is a real number.

	\subsubsection{Training hyperparameters}\label{app:Lorenz_hyperparameters}
	
	For the experiments on Lorenz96, we considered the batch size to be 1000. For the SR and deterministic approaches, we used Adam optimizer and tested the following learning rate values: $ 10^{-i} $ for $ i=1,\ldots, 6 $ for the SR methods and $ 10^{-i-1} $ and $ 3\cdot 10^{-i - 1} $ for $ i=1,\ldots, 3 $ for regression. We fix the GRU hidden size to 32. We report then the performance achieved with the learning rate yielding lower loss on the validation set, which is indicated in Table~\ref{tab:lr_lorenz96_SR}.
	
	\begin{table}
		\begin{center}
			\begin{tabular}{cccc}
				\toprule
				Energy & Kernel & Energy-Kernel & Regression \\
				\midrule
				0.01  & 0.001 & 0.001 & 0.003 \\
				\bottomrule
			\end{tabular}
		\end{center}
		\caption{Optimal learning rate values for SR and regression (deterministic) approaches for Lorenz96.}
		\label{tab:lr_lorenz96_SR}
	\end{table}
	
	%
	%

	For the GAN and WGAN-GP approach, we used Adam optimizer and we tested the following learning rate values for both critic and generative network: $ 10^{-i} $ and $ 3\cdot 10^{-i} $ for $ i=1,\ldots, 7$. In total, those are 14 learning rate values. 
	We tested hidden size 32 and 64; further, we experiment with 4 number of critic training steps for WGAN-GP (1, 3, 5, 10), in order to have the best possible results to compare with our SR methods, while we left the number of critic training steps to 1 for GAN. Overall, therefore, we had $ 2\cdot 14^2  = 392$ experiments for GAN and  $ 2\cdot 4 \cdot 14^2  = 1568$ for WGAN-GP; notice the extremely larger number number of experiments for the adversarial approaches with respect to SR ones, which highlights an advantage of our approach. We stress that such a number of trials could be possible only for the low-dimensional setting of the Lorenz63 and Lorenz96 models, in which training is cheap, but not in real-life applications.
	
	Additionally, the adversarial approaches do not allow to select hyperparameters according to loss on a validation set, as the generator loss depends on the current state of the discriminator (i.e., there is no absolute loss scale). Therefore, we report results for 3 different configurations for GAN and WGAN-GP, maximizing either deterministic performance (1) or calibration (2), or striking the best balance between these two (3). The resulting learning rates are in Table~\ref{tab:lr_lorenz96_GAN}. Notice that, for GAN, there was no configuration leading to intermediate performance between (1) and (2), so that the column for (3) is left empty.
	
	\begin{table}
		\begin{center}
			\begin{adjustbox}{max width=\textwidth}
				\begin{tabular}{lcccccc}
					\toprule
					& GAN (1) & GAN (2) & GAN (3) & WGAN-GP (1) & WGAN-GP (2) & WGAN-GP (3) \\
					\midrule
					Generator  l.r. & 0.01 & 0.0001 & 0.0001 & 0.001& 0.00003 & 0.0001 \\
					Critic l.r. & 0.001 & 0.003 & 0.001 & 0.001 &0.1 &0.01 \\
					GRU hidden size & 64 & 32 & 64& 64 & 64 & 64 \\
					Critic training steps & 1 & 1& 1 & 10 & 1 & 5 \\
					\bottomrule
				\end{tabular}
			\end{adjustbox}
		\end{center}
		\caption{Optimal hyperparameter values for adversarial approaches for Lorenz96 model.}
		\label{tab:lr_lorenz96_GAN}
	\end{table}

	\subsection{WeatherBench data set}
	
	\subsubsection{Variogram Score}\label{app:weatherbench_Variogram}
	
	For the Variogram Score, we use a weight matrix which is inversely proportional to the Haversine distance, which measures the angular distance between two points on the surface of a sphere.
	Specifically, by denoting the longitude and latitude (in radians) of component $ i $ of $ \ddobs $ as $ \operatorname{lon}_i, \operatorname{lat}_i $, the Haversine distance is defined as: 	
	\begin{equation}\label{key}
	d_{ij} = 2\arcsin\left[\sqrt{\sin^2((\operatorname{lat}_i - \operatorname{lat}_j) / 2)
		+ \cos(\operatorname{lat}_i)\cos(\operatorname{lat}_j)\sin^2((\operatorname{lon}_i - \operatorname{lon}_j) / 2)}\right]
	\end{equation}
	The physical distance along the sphere can be computed by multiplying the above by Earth's radius (approximately $ 6371 $ km). However, that is just a scaling constant, therefore we ignore it in defining the variogram, which we take to be $ w_{ij} = 1/d_{ij}$.

	\subsubsection{Choice of weights for summed scores}\label{app:weatherbench_weights}
	
	In the summed Scores (Energy-Variogram, Kernel-Variogram, Energy-Kernel and Patched Energy Score), we need to select the weights for the two addends. Notice that, in the Patched Energy Score, we consider the Energy Score computed on the full data to be the first addend, and the sum of the Energy Scores computed on each patch to be the second addend.
	
	We fix the weights such that the two addends have roughly the same magnitude. This results, for the Energy-Variogram, Kernel-Variogram, Energy-Kernel, in the choices reported the Table~\ref{Tab:weights}.

	\begin{table}
		\begin{center}
			\begin{tabular}{lccc}
				\toprule
				&Energy-Kernel & Energy-Variogram & Kernel-Variogram \\
				\midrule
				$ \alpha_1 $ & $ 1/70 $ & $ 1 $& $ 1 $  \\
				$ \alpha_2 $ & $ 1 $ &  $ 6.94 \cdot10^{-7} $&  $ 1.3 \cdot 10^{-8} $ \\
				\bottomrule
			\end{tabular}
		\end{center}
		\caption{Weights for summed Scores.}
		\label{Tab:weights}
	\end{table}

	For the Patched Energy Score, we use the following two setups in our experiments:
	\begin{itemize}
		\item Patches of size 16 separated by 8 grid points: this leads to 32 patches. As the Energy Score scales as the data dimensionality, each of the $ 16\times16=256 $ patches has relative magnitude with respect to Energy Score computed on the full WeatherBench grid $ 256/2048=0.125 $, where $ 32\times 64=2048 $ is the size of the WeatherBench grid. However, we sum the Score for each of the 32 patches, which leads to a quantity with magnitude 4 times the one of the overall Energy Score.
		\item Patches of size 8 separated by 4 grid points: this leads to 128 patches. Following the argument above, each $ 8\times 8=64  $ patch gives a Score with relative magnitude $ 64/2048=0.03125 $. As there are 128 patches, again the cumulative patched score has magnitude 4 times the overall one.
	\end{itemize}
	
	In both cases, we leave therefore $ \alpha_1=\alpha_2=1 $, as the patched and overall components are already of similar magnitude (they just differ by a factor 4).

	\subsubsection{Neural Networks architecture}\label{app:weatherbench_nets}
	
\begin{figure*}
		\centering
		\begin{subfigure}{0.2\textwidth}
			\begin{center}
				\includegraphics[width=\columnwidth]{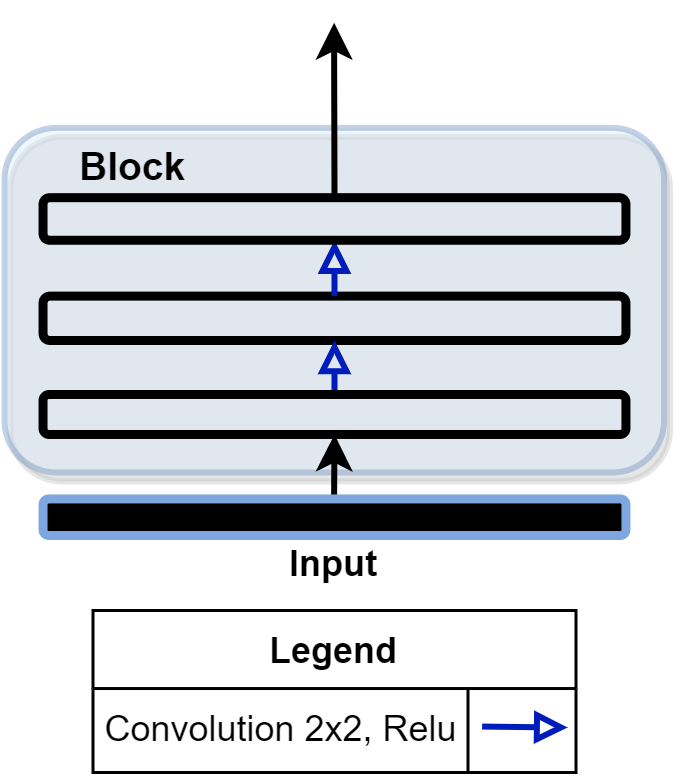}
				\caption{Structure of each block.}
			\end{center}
		\end{subfigure}~
		\begin{subfigure}{0.7\textwidth}
			\begin{center}
				\includegraphics[width=\columnwidth]{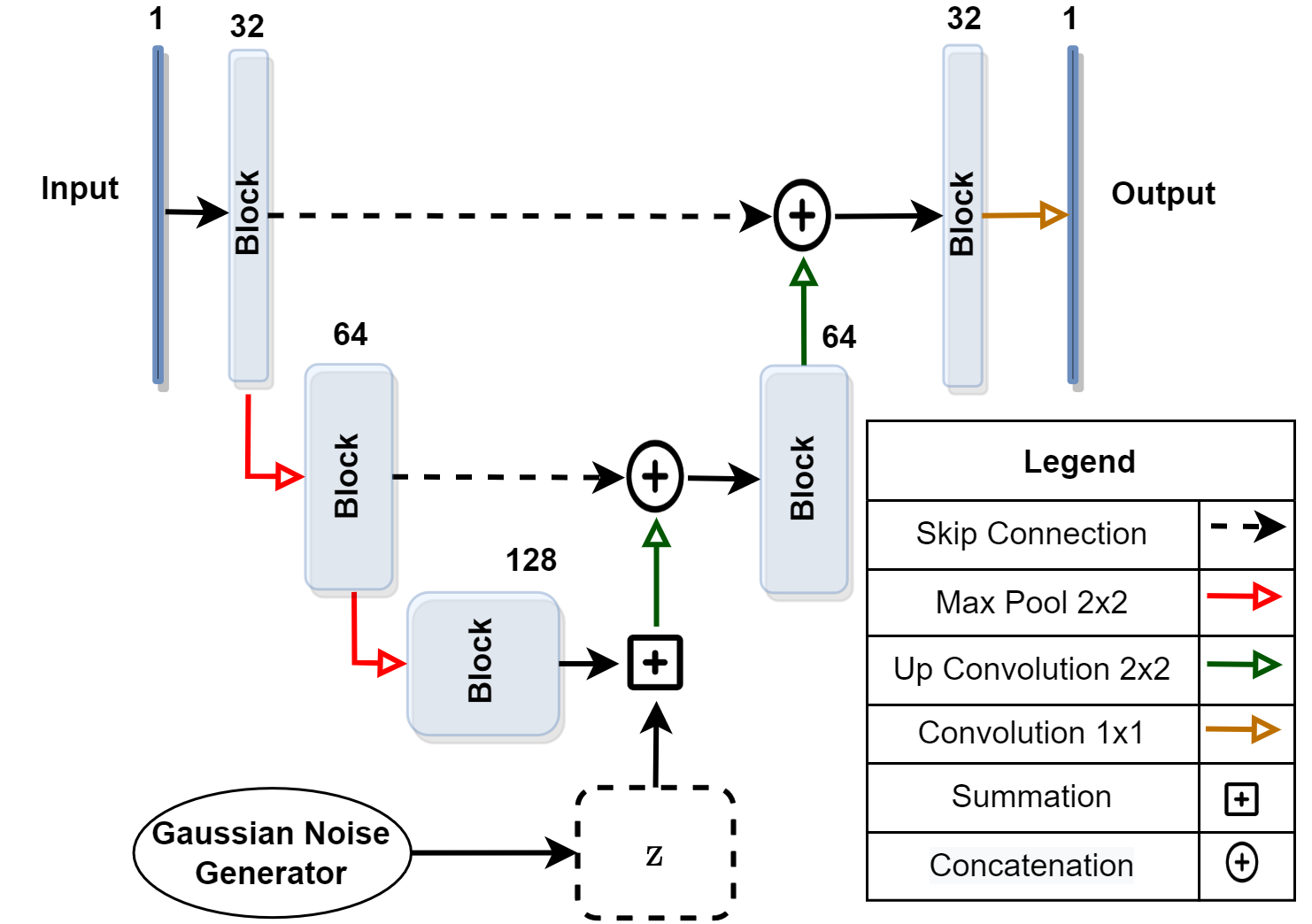}
				\caption{Full U-NET architecture.}
			\end{center}
		\end{subfigure}
		\caption{U-NET architecture.}	\label{fig:UNET}
	\end{figure*}

	For the generative network, we use a U-NET architecture \citep{ronneberger2015unet}, which is an encoder-decoder structure, where each subsequent layer of the encoder outputs a downscaled latent representation of the input variables. The final output of the encoder is passed to a bottleneck layer, which performs no up/down scaling. The output of this bottleneck layer is then passed to the decoder. Conversely to the encoder, each subsequent layer of the decoder outputs an upscaled latent representation of the bottleneck layer output. Additionally, skip connections allow information to pass directly between layers of the encoder and decoder at the same scale; in this way, both large scale structures and high-frequency information contributes to the output.
	The latent variable $ \mathbf{Z} $ is summed to the latent representation in the bottleneck layer. Figure~\ref{fig:UNET} gives a graphical representation of the UNet. For the deterministic setting trained with the regression loss, the architecture is analogous, the only difference being that no latent variable $ \mathbf{Z} $ is summed to the latent representation.

	In the adversarial setups, we use the PatchGAN critic suggested in \cite{isola2017image}. Specifically, this is a convolutional network which considers separate patches of the input image and outputs a numerical value for each patch, corresponding, in the original GAN setting of \cite{goodfellow2014generative}, to the confidence with which the critic believes that patch is real, in contrast to generated from the generative network. The GAN or WGAN loss is then computed for each of the output values and averaged. 	
	
	The PatchGAN critic employs some Batch Normalization layers; however, these cannot be used when the gradient penalization strategy of WGAN-GP is used \citep{gulrajani2017improved}. Therefore, as suggested in \cite{gulrajani2017improved}, we replace the Batch Normalization layers with Layer Normalization.
	
	As before, in the GAN case, the critic outputs a value between 0 and 1 indicating how confident the critic believes that is a fake sample. In the WGAN-GP case, the critic output is a real number.

	\subsubsection{Training hyperparameters}\label{app:weatherbench_hyperparmeters}
	
	For the SR approaches for the WeatherBench data set, we considered the batch size to be 128 for all experiments, except for those on the Energy-Variogram and Kernel-Variogram score, which resulted in GPU memory overflow with that batch size (in fact, computing the Variogram Score is an operation requiring quadratic memory with respect to data size); for these two, we fixed therefore the batch size to be 48. We used Adam optimizer and tested the following learning rate values $ 10^{-i} $ for $ i=1,\ldots, 6 $. We report then the performance achieved with the learning rate yielding lower loss on the validation set in Table~\ref{tab:weatherbench_lr_SRs}.
	
	For the deterministic network trained via regression, we test learning rule values $ 10^{-i-1} $ for $ i=1,\ldots, 4 $; additionally, we use an exponential learning rate scheduler which reduces the learning rate by multiplying it by a factor $ \gamma $ every $ 10 $ training epochs. We also use a $ \ell2 $ weight regularization with weight $ \lambda  $. We try different values of these parameters in conjunction with the learning rate values; the ones with which best validation loss is obtained are $ \gamma=0.8 $ and $ \lambda=0.001 $. The best learning rate value is reported in Table~\ref{tab:weatherbench_lr_SRs}. Notice that the same learning rate value was optimal for the full (non-patched) regression loss and for the patched loss in both configurations.

	\begin{table}
		\begin{center}
			\begin{tabular}{lccccc}
				\toprule
				& Regression& Energy & kernel & Energy-Kernel & Energy-Variogram  \\
				\midrule
				Learning rate & 0.01 &  0.0001 & 0.0001& 0.0001& $ 10^{-5} $ \\
				\bottomrule
			\end{tabular}
			\begin{tabular}{lccc}
				\toprule
				& Kernel-Variogram & Patched Energy (8) & Patched Energy (16) \\
				\midrule
				Learning rate &  $ 10^{-5} $& $ 10^{-5} $& $ 10^{-5} $ \\
				\bottomrule
			\end{tabular}
		\end{center}
		\caption{Optimal learning rate values for the SR and regression (deterministic) approaches for WeatherBench.}
		\label{tab:weatherbench_lr_SRs}
	\end{table}

	For the GAN and WGAN-GP approach, we used Adam optimizer and we tested the following learning rate values for both critic and generative network: $ 10^{-i} $, $ i=1,\ldots, 7$. In total, those are 7 learning rate values, which result in $ 7^2 = 49 $ experiments. Notice additionally that the adversarial approaches does not allow to select hyperparameters according to loss on a validation set, as the generator loss depends on the current state of the discriminator (i.e., there is no absolute loss scale). Additionally, the adversarial approaches do not allow to select hyperparameters according to loss on a validation set, as the generator loss depends on the current state of the discriminator (i.e., there is no absolute loss scale). Therefore, we report results for 3 different configurations for GAN, maximizing either deterministic performance (1) or calibration (2), or striking the best balance between these two (3). For WGAN-GP, a single configuration maximized both calibration and deterministic performance, so that we report that one. The resulting learning rates are in Table~\ref{tab:lr_Weatherbench_GAN}. 
	
	\begin{table}
		\begin{center}
			\begin{adjustbox}{max width=\textwidth}
				\begin{tabular}{lcccc}
					\toprule
					& GAN (1)	& GAN (2) &GAN (3) & WGAN-GP \\
					\midrule
					Generator learning rate & 0.001 &$ 10^{-6} $& $ 10^{-5} $& $ 10^{-5} $\\
					Critic learning rate &0.0001 & 0.0001& $ 10^{-5} $ & 0.01 \\
					\bottomrule
				\end{tabular}
			\end{adjustbox}
		\end{center}
		\caption{Optimal hyperparameter values for adversarial approaches for WeatherBench.}
		\label{tab:lr_Weatherbench_GAN}
	\end{table}

	\section{Additional experimental results}\label{app:exp_res}
	
	\subsection{Additional results for Lorenz63 model}\label{app:res_lorenz63}

\begin{figure*}[tb]
		\centering
		\begin{subfigure}{0.33\textwidth}
			\begin{center}
				\includegraphics[width=\columnwidth]{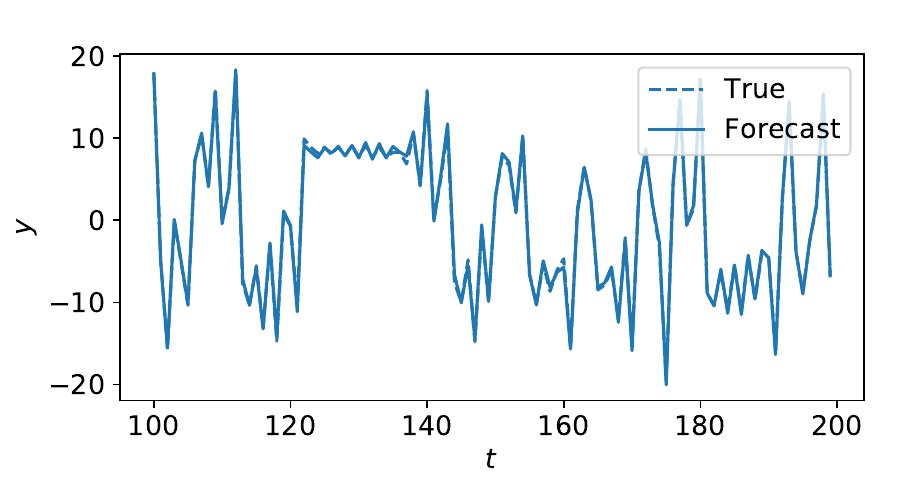}
			\end{center}
			\caption{Regression}
		\end{subfigure}~
		\begin{subfigure}{0.33\textwidth}
			\begin{center}
				\includegraphics[width=\columnwidth]{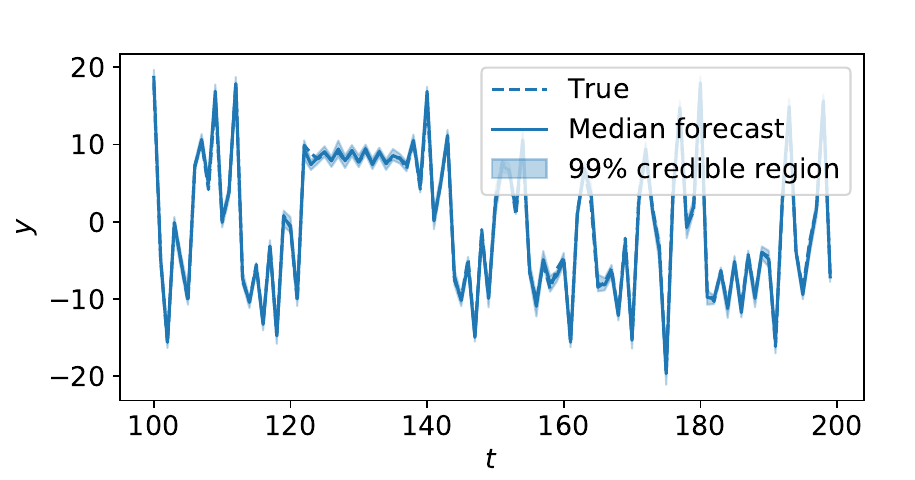}
			\end{center}
			\caption{Energy Score}
		\end{subfigure}\\
		\begin{subfigure}{0.33\textwidth}
			\begin{center}
				\includegraphics[width=\columnwidth]{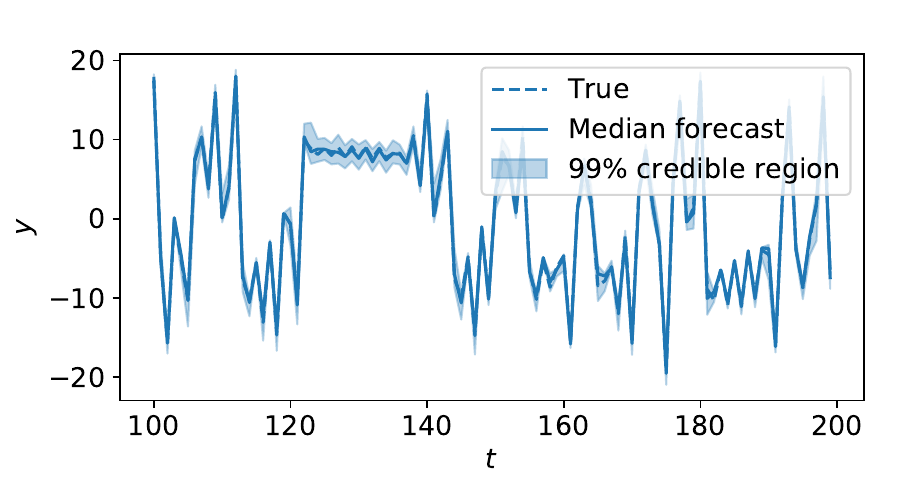}
				\caption{Kernel Score}
			\end{center}
		\end{subfigure}~
		\begin{subfigure}{0.33\textwidth}
			\begin{center}
				\includegraphics[width=\columnwidth]{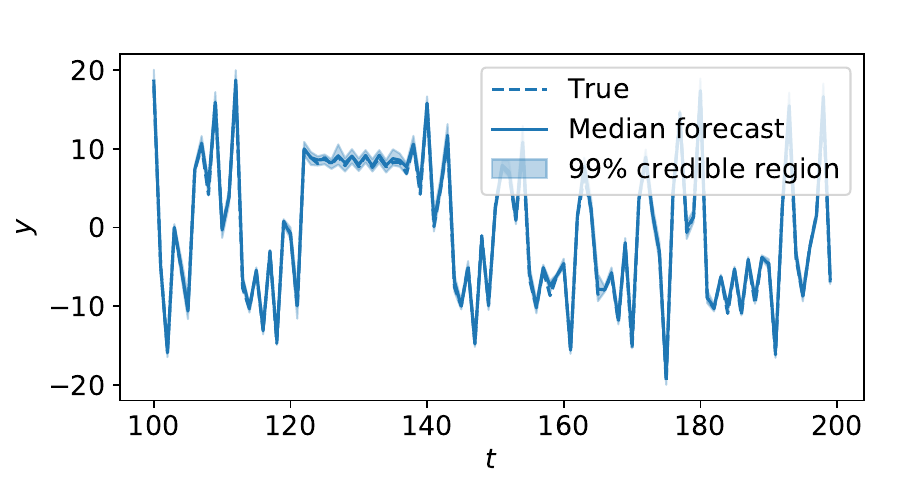}
				\caption{Energy-Kernel Score}
			\end{center}
		\end{subfigure}\\
		\begin{subfigure}{0.33\textwidth}
			\centering
			\includegraphics[width=\textwidth]{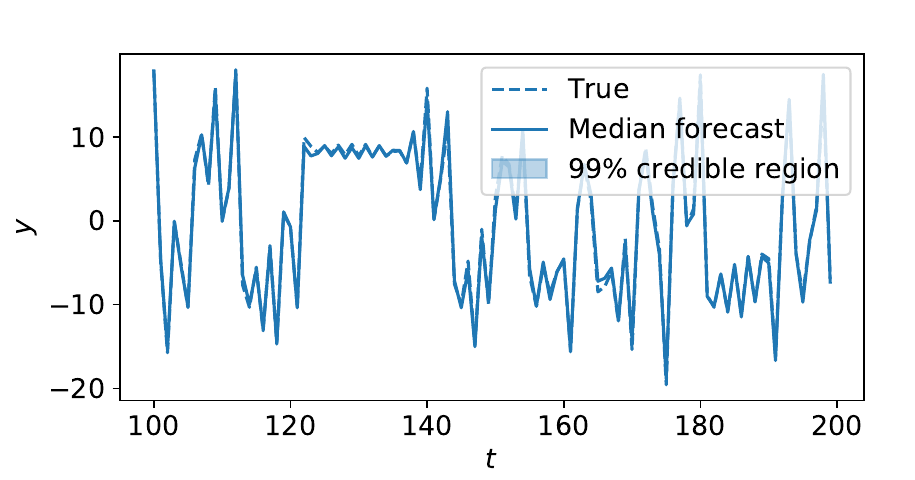}
			\caption{GAN (1) }
		\end{subfigure}~
		\begin{subfigure}{0.33\textwidth}
			\centering
			\includegraphics[width=\textwidth]{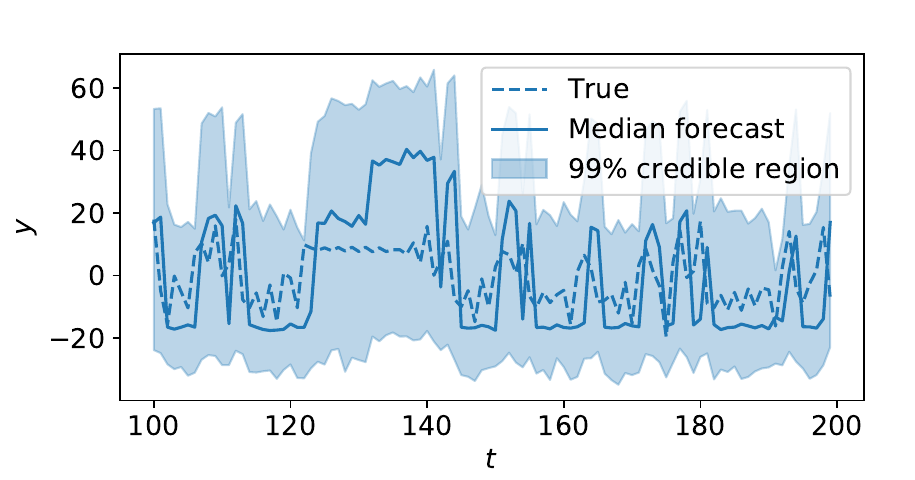}
			\caption{GAN (2)}
		\end{subfigure}~
		\begin{subfigure}{0.33\textwidth}
			\centering
			\includegraphics[width=\textwidth]{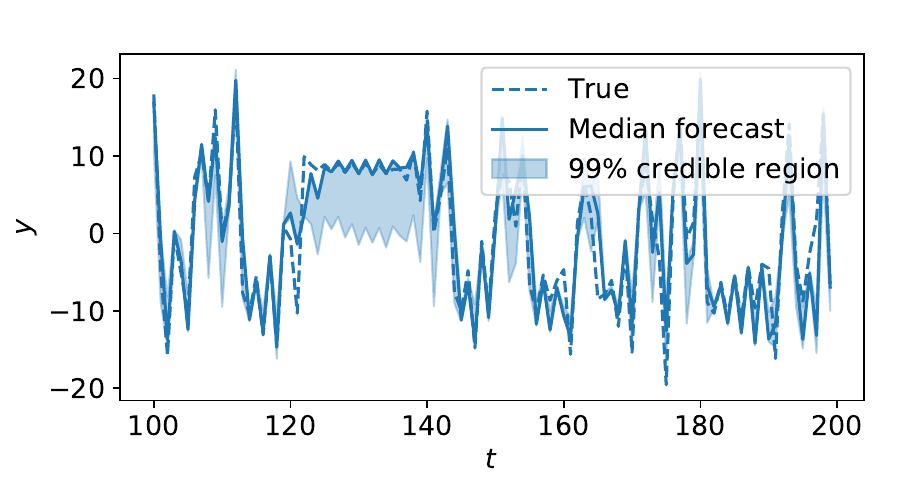}
			\caption{GAN (3)}
		\end{subfigure}\\
		\begin{subfigure}{0.33\textwidth}
			\begin{center}
				\includegraphics[width=\columnwidth]{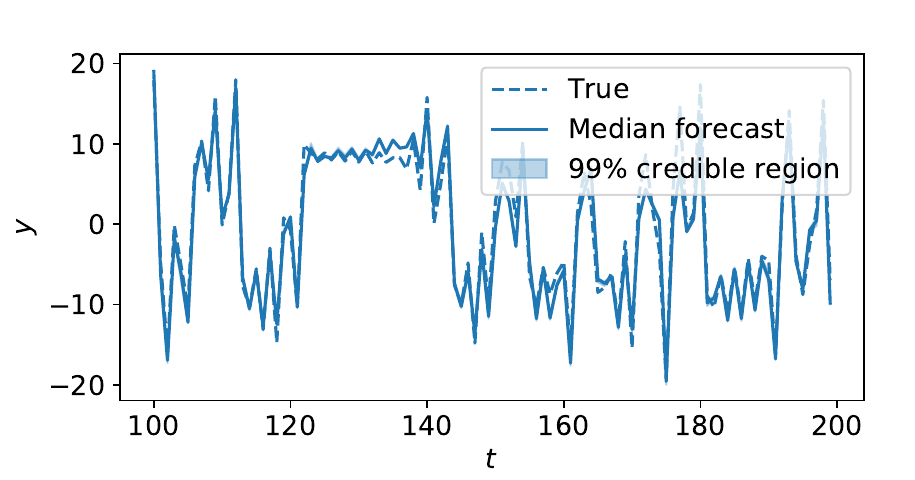}
				\caption{WGAN-GP (1) }
			\end{center}
		\end{subfigure}~
		\begin{subfigure}{0.33\textwidth}
			\begin{center}
				\includegraphics[width=\columnwidth]{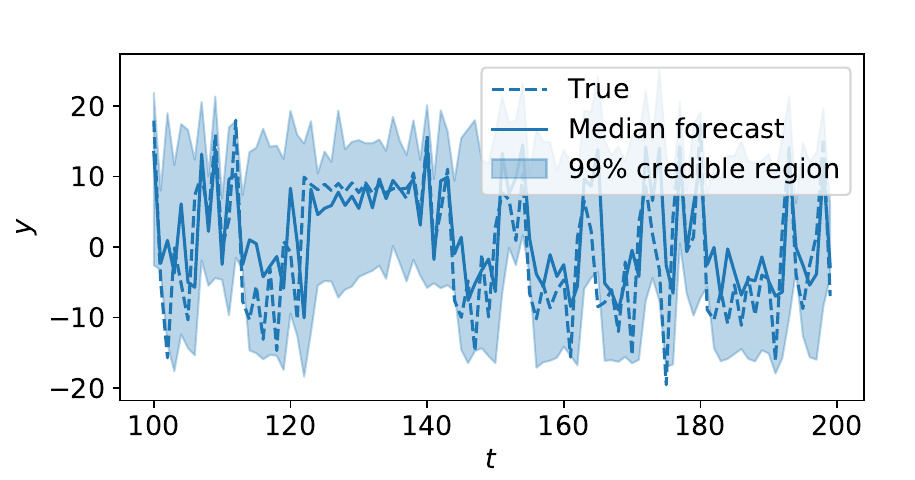}
				\caption{WGAN-GP (2) }
			\end{center}
		\end{subfigure}~
		\begin{subfigure}{0.33\textwidth}
			\begin{center}
				\includegraphics[width=\columnwidth]{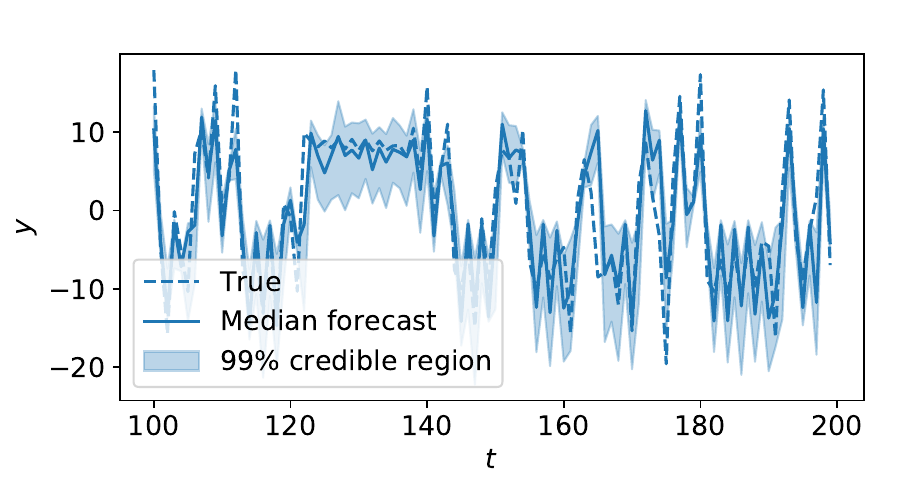}
				\caption{WGAN-GP (3) }
			\end{center}
		\end{subfigure}~
		\caption{Results for the Lorenz63 model with all considered methods. The figures show observations, median forecast and 99\% credible interval for a portion of the test set. For each time-step, forecasts are obtained using the previous observation window.}
		\label{fig:lorenz63_rnn}
	\end{figure*}

	
	We report here additional results. Figure~\ref{fig:lorenz63_rnn} contains separate plots for all methods showing forecasts and realization for a portion of the test set (the same used in Section~\ref{sec:lorenz63} in the main text).

	\subsection{Additional results for Lorenz96 model}\label{app:res_lorenz96}

	We report here additional results. 
	Table~\ref{tab:res_lorez96_rnn} reports the average and standard deviation of the different performance measures computed across the different data components. It contains the same results as Table~\ref{tab:res_lorez63_lorenz96} in the main text, where however the standard deviation was not reported.

\begin{figure*}[htb]
		\centering
		\begin{subfigure}{0.33\textwidth}
			\begin{center}
				\includegraphics[width=\columnwidth]{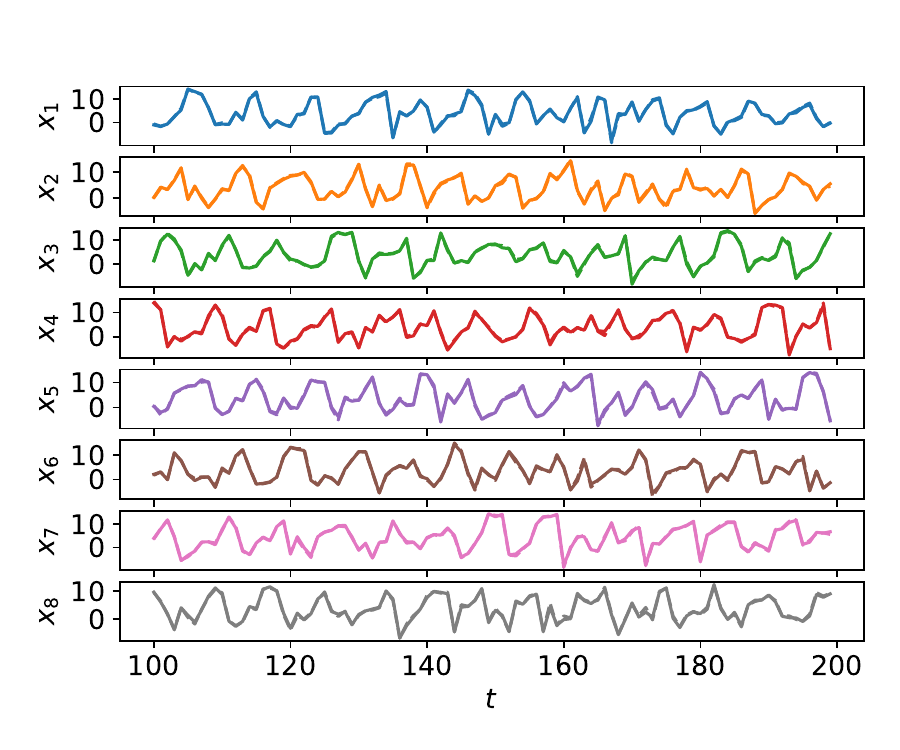}
			\end{center}
			\caption{Regression}
		\end{subfigure}~
		\begin{subfigure}{0.33\textwidth}
			\begin{center}
				\includegraphics[width=\columnwidth]{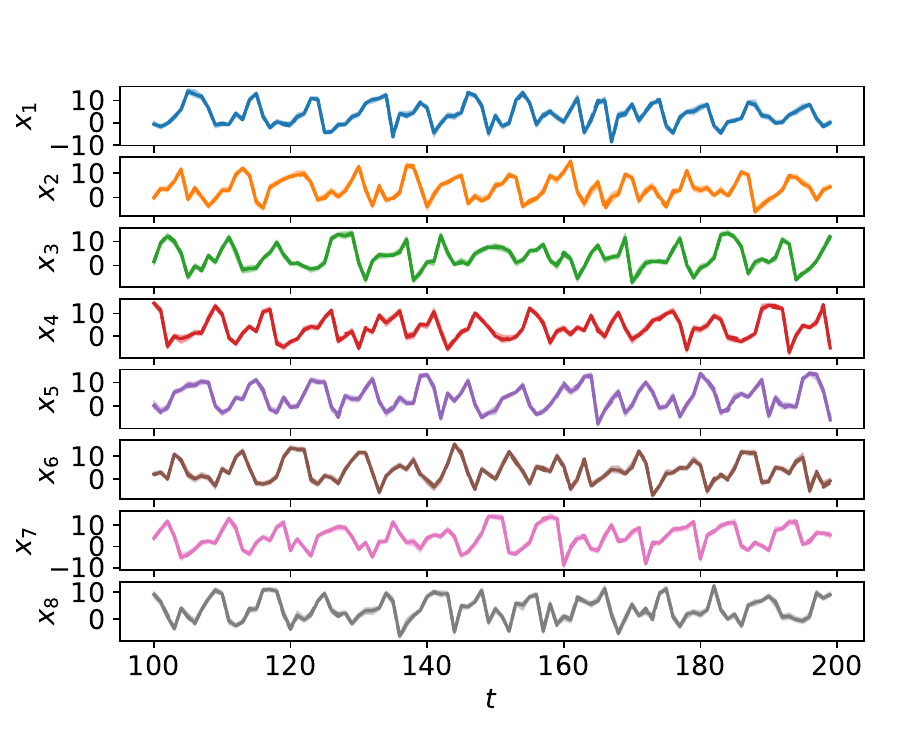}
			\end{center}
			\caption{Energy Score.}
		\end{subfigure}\\
		\begin{subfigure}{0.33\textwidth}
			\begin{center}
				\includegraphics[width=\columnwidth]{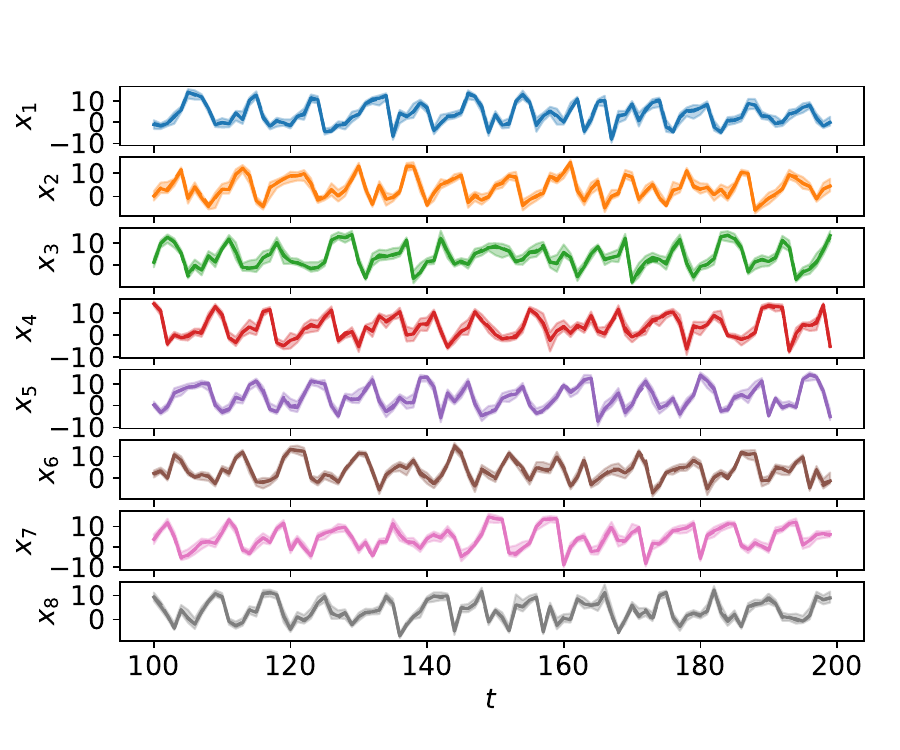}
			\end{center}
			\caption{Kernel Score.}
		\end{subfigure}~
		\begin{subfigure}{0.33\textwidth}
			\begin{center}
				\includegraphics[width=\columnwidth]{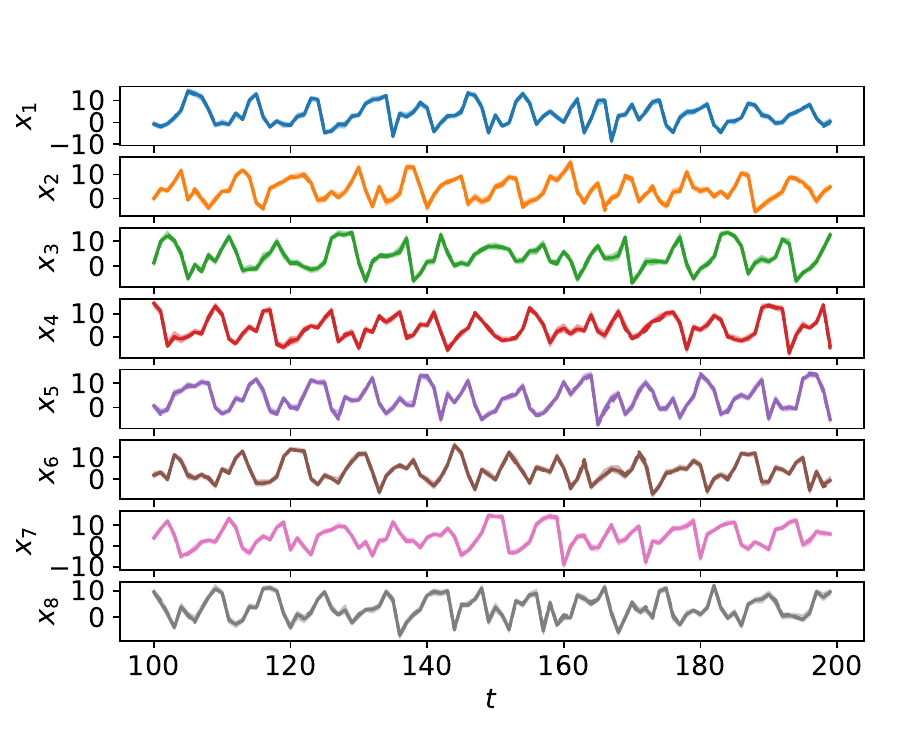}
			\end{center}
			\caption{Energy-Kernel Score}
		\end{subfigure}\\
		\begin{subfigure}{0.33\textwidth}
			\centering
			\includegraphics[width=\columnwidth]{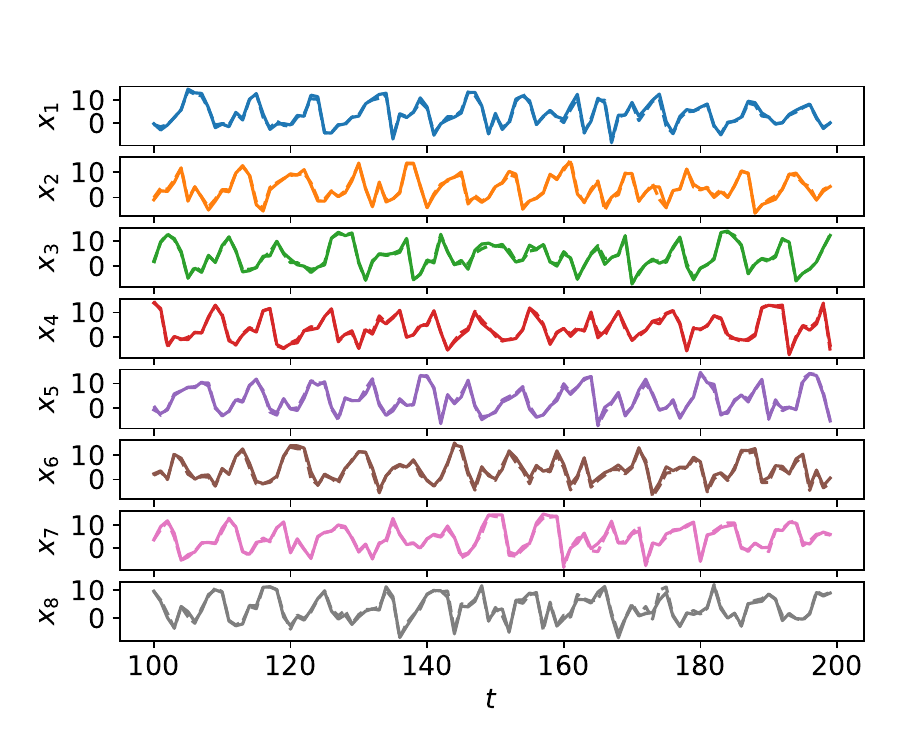}
			\caption{GAN (1)}
		\end{subfigure}~
		\begin{subfigure}{0.33\textwidth}
			\centering
			\includegraphics[width=\columnwidth]{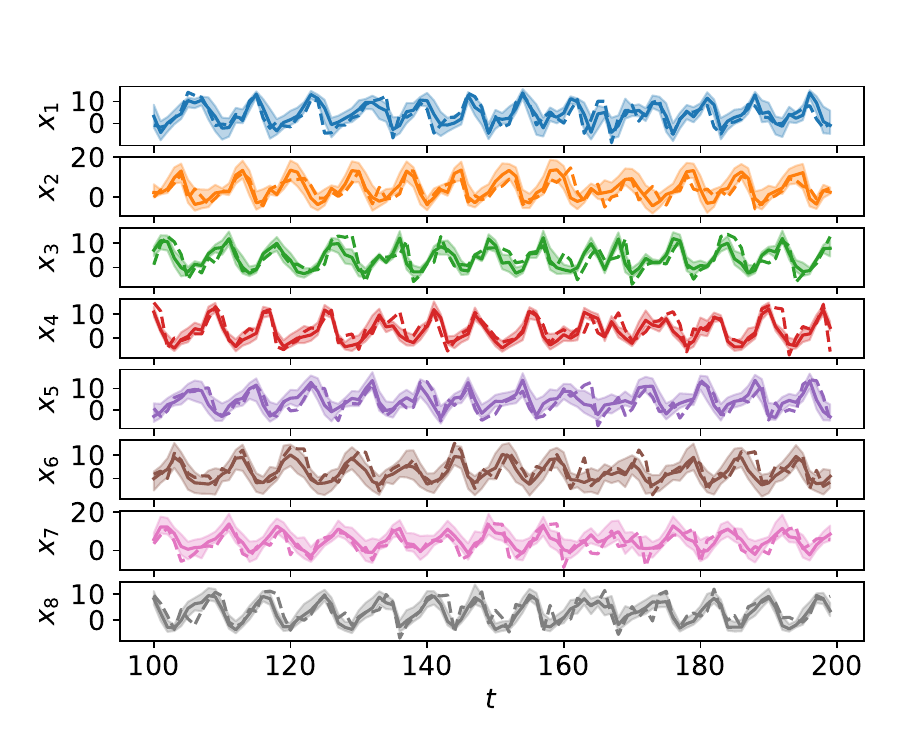}
			\caption{GAN (2)}
		\end{subfigure}~
		\begin{subfigure}{0.33\textwidth}
			\centering
			\includegraphics[width=\columnwidth]{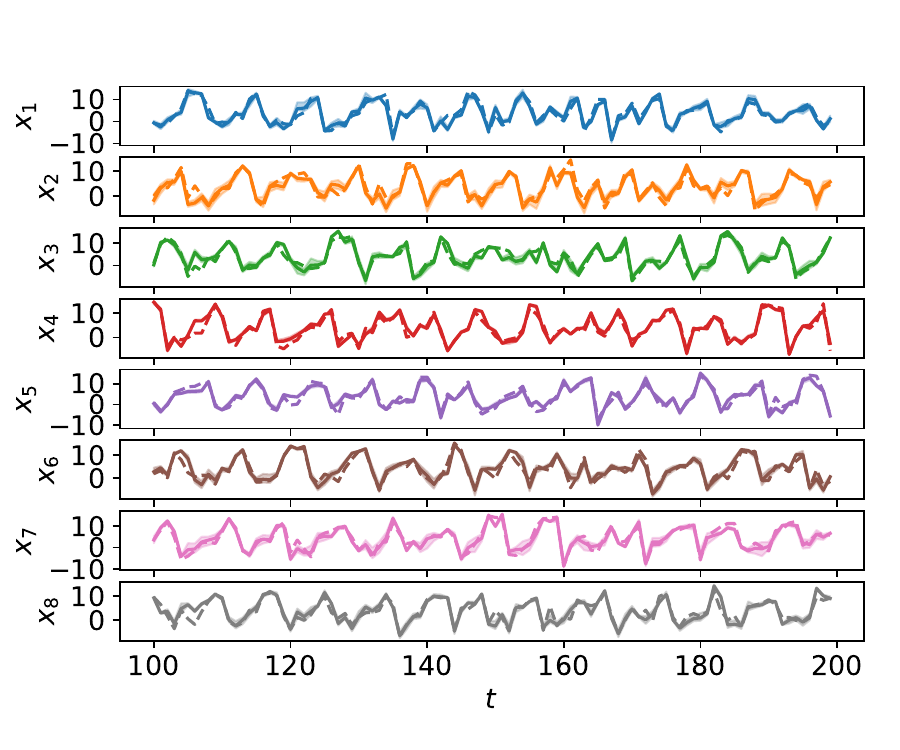}
			\caption{GAN (3)}
		\end{subfigure}\\
		\begin{subfigure}{0.33\textwidth}
			\centering
			\includegraphics[width=\columnwidth]{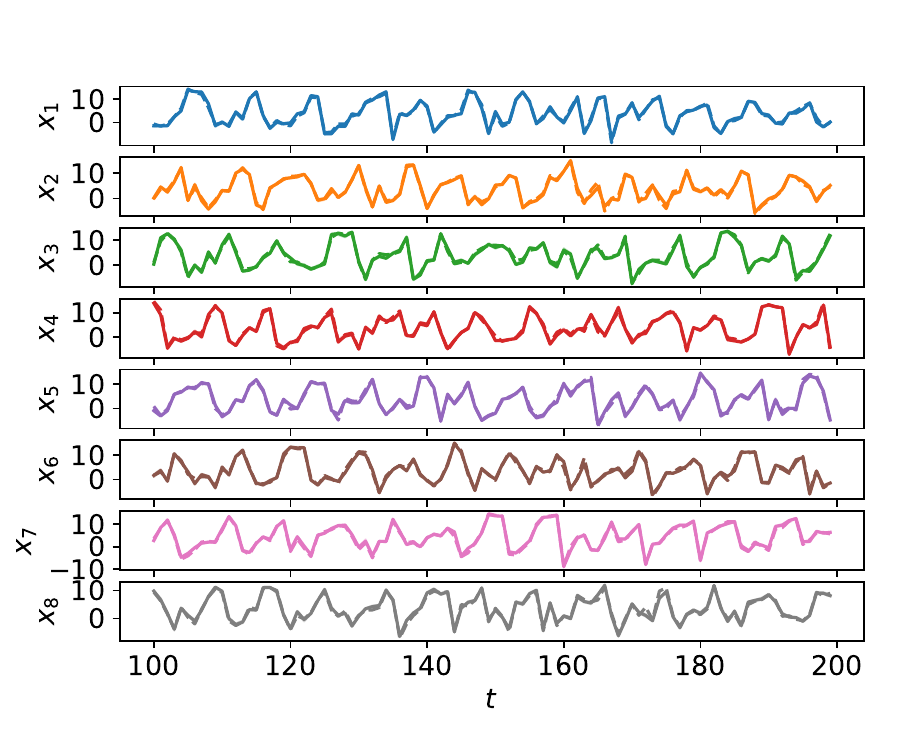}
			\caption{WGAN-GP (1)}
		\end{subfigure}~
		\begin{subfigure}{0.33\textwidth}
			\centering
			\includegraphics[width=\columnwidth]{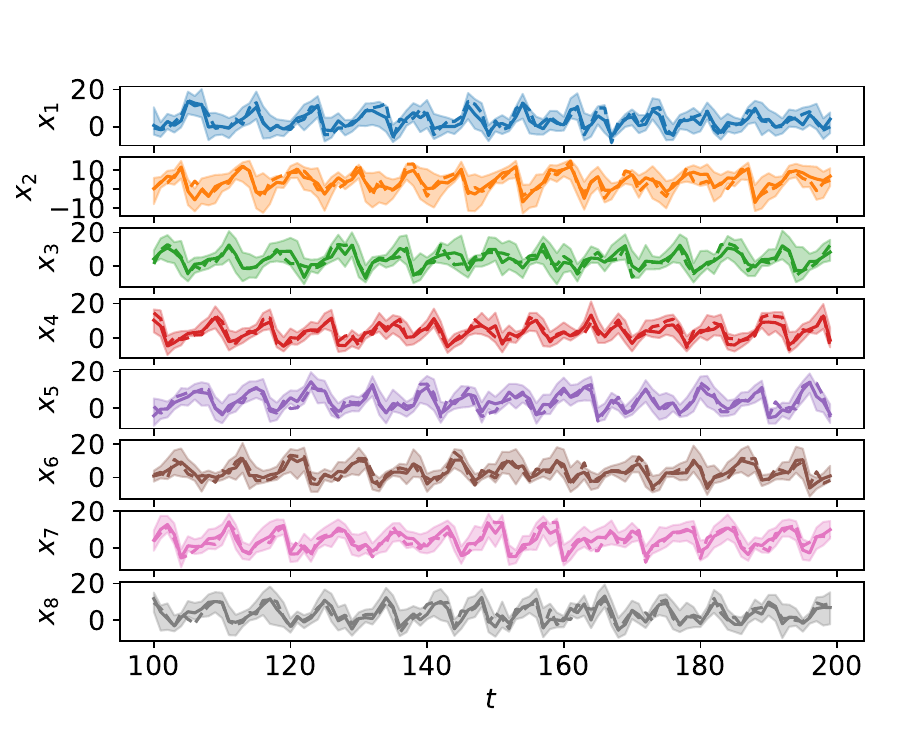}
			\caption{WGAN-GP (2)}
		\end{subfigure}~
		\begin{subfigure}{0.33\textwidth}
			\centering
			\includegraphics[width=\columnwidth]{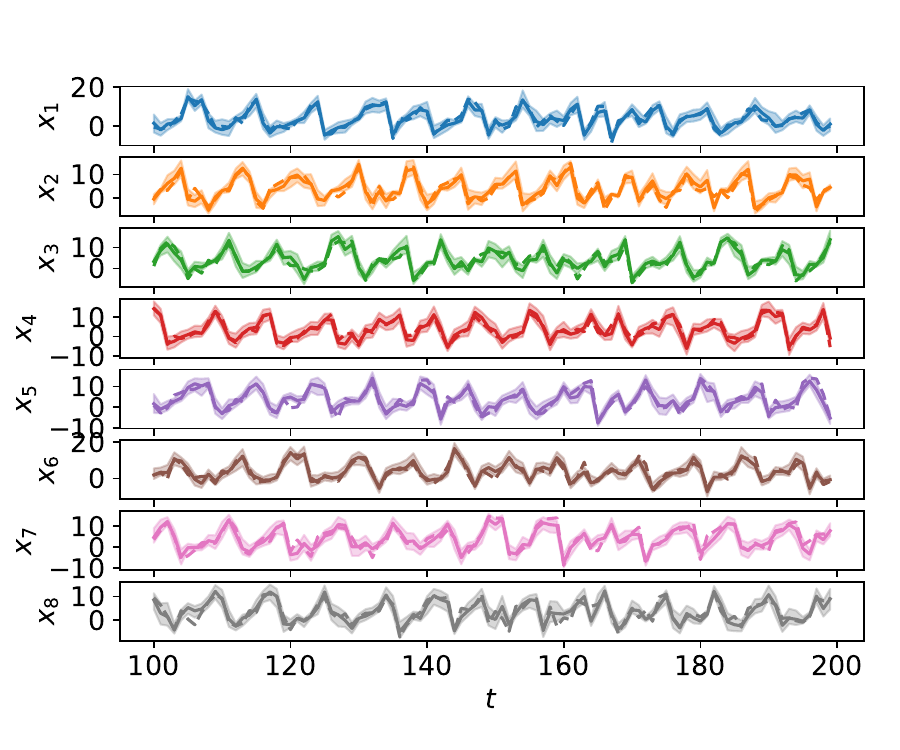}
			\caption{WGAN-GP (3)}
		\end{subfigure}\\
		\caption{Results for the Lorenz96 model with all considered methods. Panels show observations (dashed line), median forecast (solid line) and 99\% credible interval (shaded region) for a portion of the test set. That is done for all $ 8 $ components of $ \mathbf x $. For each time-step, forecasts are obtained using the previous observation window.}
		\label{fig:lorenz96_rnn}
	\end{figure*}

	Figure~\ref{fig:lorenz96_rnn} contains separate plots for all methods showing forecasts and realization for a portion of the test set (the same used in Section~\ref{sec:lorenz63} in the main text).

	\begin{table}[tb]
		\begin{center}
			\begin{tabular}{lccc}
				\toprule
				& Cal. error $ \downarrow $ & NRMSE $ \downarrow $ & R$ ^2 $ $ \uparrow $ \\
				\midrule
				Regression & - & 0.0198  $ \pm$ 0.0006 &  0.9905 $ \pm$ 0.0006  \\
				Energy & 0.0205 $ \pm$ 0.0176 & 0.0166  $ \pm$ 0.0014 &  0.9933 $ \pm$ 0.0012   \\
				Kernel & 0.2196 $ \pm$ 0.0123 & 0.0164  $ \pm$ 0.0003 &  0.9935 $ \pm$ 0.0003  \\
				Energy-Kernel & 0.0104 $ \pm$ 0.0060 & 0.0173  $ \pm$ 0.0004 &  0.9928 $ \pm$ 0.0004  \\
				GAN (1) & 0.4644 $ \pm$ 0.0062 & 0.0354  $ \pm$ 0.0026 &  0.9696 $ \pm$ 0.0044  \\
				GAN (2) & 0.2671 $ \pm$ 0.0559 & 0.1500  $ \pm$ 0.0090 &  0.4537 $ \pm$ 0.0619   \\
				GAN (3) & 0.3700 $ \pm$ 0.0369 & 0.0763  $ \pm$ 0.0030 &  0.8590 $ \pm$ 0.0099 \\
				WGAN-GP (1) & 0.4134 $ \pm$ 0.0051 & 0.0330  $ \pm$ 0.0007 &  0.9736 $ \pm$ 0.0009 \\
				WGAN-GP (2) & 0.0565 $ \pm$ 0.0339 & 0.1081  $ \pm$ 0.0037 &  0.7165 $ \pm$ 0.0200\\ 
				WGAN-GP (3) & 0.1648 $ \pm$ 0.0444 & 0.0786  $ \pm$ 0.0041 &  0.8502 $ \pm$ 0.0149 \\ 
				\bottomrule
			\end{tabular}
		\end{center}
		\caption{Average and standard deviation of performance measures for forecasts obtained with the different methods, on the test set for the Lorenz96 data set. Metrics are computed on each data component individually; then, the average and standard deviation is computed.}	\label{tab:res_lorez96_rnn}
	\end{table}

	\subsection{WeatherBench data set}\label{app:weatherbench_res}


	\subsubsection{Standard deviation of performance measures}
	In Table~\ref{tab:res_weatherbench_}, the average and standard deviation of the different performance measures are computed across the different data components.

	\begin{table}[tb]
		\begin{center}
			\begin{tabular}{lccc}
				\toprule
				& Cal. error $ \downarrow $ & NRMSE $ \downarrow $ & R$ ^2 $ $ \uparrow $ \\
				\midrule
				Regression & - & 0.1162  $ \pm$ 0.0256 &  0.5300 $ \pm$ 0.2559 \\
				Patched Regression, 8 & - & 0.1147  $ \pm$ 0.0238 &  0.5459 $ \pm$ 0.2297  \\
				Patched Regression, 16 & - & 0.1144  $ \pm$ 0.0227 &  0.5509 $ \pm$ 0.2188 \\
				Energy & 0.0863 $ \pm$ 0.0407 & 0.1208  $ \pm$ 0.0256 &  0.4968 $ \pm$ 0.2596 \\
				Kernel & 0.0797 $ \pm$ 0.0455 & 0.1200  $ \pm$ 0.0226 &  0.5097 $ \pm$ 0.2226 \\
				Energy-Kernel &0.0794 $ \pm$ 0.0433 & 0.1194  $ \pm$ 0.0226 &  0.5150 $ \pm$ 0.2225  \\
				Energy-Variogram & 0.0899 $ \pm$ 0.0541 & 0.1192  $ \pm$ 0.0220 &  0.5177 $ \pm$ 0.2180  \\
				Kernel-Variogram &0.1704 $ \pm$ 0.0607 & 0.1203  $ \pm$ 0.0238 &  0.5050 $ \pm$ 0.2399  \\
				Patched Energy, 8 & 0.0550 $ \pm$ 0.0348 & 0.1189  $ \pm$ 0.0209 &  0.5217 $ \pm$ 0.2064 \\
				Patched Energy, 16 & 0.0690 $ \pm$ 0.0478 & 0.1186  $ \pm$ 0.0208 &  0.5248 $ \pm$ 0.2034  \\
				GAN (1) & 	0.4845 $ \pm$ 0.0089 & 0.1573  $ \pm$ 0.0391 &  0.1418 $ \pm$ 0.5267  \\
				GAN (2) & 	0.3130 $ \pm$ 0.1143 & 0.2487  $ \pm$ 0.2248 &  -2.7970 $ \pm$ 17.1346  \\
				GAN (3) & 0.3625 $ \pm$ 0.0545 & 0.1693  $ \pm$ 0.0494 &  -0.0117 $ \pm$ 0.8348  \\
				WGAN-GP & 0.1009 $ \pm$ 0.0679 & 0.1302  $ \pm$ 0.0214 &  0.4340 $ \pm$ 0.2271  \\
				\bottomrule
			\end{tabular}
		\end{center}
		\caption{Average and standard deviation of performance measures for forecasts obtained with the different methods, on the test section of the WeatherBench data set. Metrics are computed on each data component individually; then, the average and standard deviation is computed.}	\label{tab:res_weatherbench_}
	\end{table}
	
	\subsubsection{Number of generator simulations for the SR methods}\label{app:n_outputs}
	
	We study here the effect of using different numbers of simulations from the generative network for each input (i.e., how many forecasts the generative network provides) during training. Recall in fact how the Energy and Kernel Score need multiple samples to be estimated (Appendix~\ref{app:scores}).

	Specifically, we consider the WeatherBench data set and the Energy Score, with learning rate $ 0.0001 $, which was found to be the optimal value when using 10 generator simulations (Appendix~\ref{app:weatherbench_hyperparmeters}). We report the measures used in the main text in Table~\ref{tab:weatherbench_ens_n}. Notice how good performance is achieved when using as little as 2 or 3 simulations.
	
	\begin{table}[tb]
		\begin{center}
			\begin{tabular}{lccc}
				\toprule
				& Cal. error $ \downarrow $ & NRMSE $ \downarrow $ & R$ ^2 $ $ \uparrow $ \\
				\midrule
				2 & 	0.0625 $ \pm$ 0.0340 & 0.1211  $ \pm$ 0.0258 &  0.4935 $ \pm$ 0.2656  \\
				3 & 0.0701 $ \pm$ 0.0342 & 0.1176  $ \pm$ 0.0208 &  0.5338 $ \pm$ 0.1961 \\
				5 & 		 0.0727 $ \pm$ 0.0348 & 0.1164  $ \pm$ 0.0198 &  0.5446 $ \pm$ 0.1842 \\
				10 & 0.0863 $ \pm$ 0.0407 & 0.1208  $ \pm$ 0.0256 &  0.4968 $ \pm$ 0.2596 \\			
				20 &		 0.0738 $ \pm$ 0.0336 & 0.1179  $ \pm$ 0.0206 &  0.5329 $ \pm$ 0.1925  \\
				30 	&	 0.0738 $ \pm$ 0.0350 & 0.1169  $ \pm$ 0.0202 &  0.5407 $ \pm$ 0.1864 \\
				50 &  0.0749 $ \pm$ 0.0356 & 0.1172  $ \pm$ 0.0203 &  0.5379 $ \pm$ 0.1889  \\
				\bottomrule
			\end{tabular}
		\end{center}
		\caption{Performance on test set of probabilistic forecasts obtained by training with the Energy Score, with different numbers of generator simulations, for the WeatherBench data set.}	\label{tab:weatherbench_ens_n}
	\end{table}

	\subsubsection{Computational cost and early stopping}\label{app:weatherbench_comp_cost}
	
	In Table~\ref{tab:comp_cost}, we report the computational cost and the early stopping achieved by the methods presented in the main text. All experiments are run on a Tesla v100 GPU, and methods are run for a maximum of 1000 epochs. We use early stopping for the SR methods, but not for GAN and WGAN-GP, for which early stopping is not possible.
	Recall that the methods with the Variogram Score used training batch size 48, while all others used 128; this fact contributes to the larger computational time for both the Energy-Variogram and Kernel-Variogram Scores.

	\begin{table}[tb]
		\begin{center}
			\begin{adjustbox}{max width=\textwidth}
				\begin{tabular}{lccc}
					\toprule
					& Per-epoch Computational cost & Early stopping at epoch & Total computational cost \\
					\midrule
					Regression & 8.45 & 250  &  2112    \\
					Patched Regression, 8  & 8.65 & 200  &  1729    \\
					Patched Regression, 16 & 8.5 & 250  &  2122    \\
					Energy & 54.2 & 100  &  5417  \\
					Kernel & 53.3  &  100  &  5329  \\
					Energy-Kernel & 55.4 &  100  &   5542  \\
					Energy-Variogram & 97.38  &  250  & 24346    \\
					Kernel-Variogram & 95.52 &  250  &  24393   \\
					Patched Energy, 8 & 56.71  &   400 &  22682  \\
					Patched Energy, 16 & 54.93  &  450  &  24717   \\
					GAN (1) & 	8.36  &  -  & 8357   \\
					GAN (2) & 	8.37  &  -  & 8373   \\
					GAN (3) & 8.33  &  -  &    8326  \\
					WGAN-GP & 7.00  &  -  &  7000   \\
					\bottomrule
				\end{tabular}
			\end{adjustbox}
		\end{center}
		\caption{Per-epoch and total computational cost, in seconds, for the different methods reported in the main text. We also report epoch at which early stopping occurred.}
		\label{tab:comp_cost}
	\end{table}
	
	Additionally, recall that, in order to achieve the performance reported in the main text, we tried 49 learning rate values for GAN and WGAN-GP, but only 6 for the SR methods. Therefore, the total computing time for GAN and WGAN-GP is the one below multiplied by 49, with respect to 6 for the SR methods. Under that perspective, even the total computing time for Energy-Variogram and Kernel-Variogram Scores is smaller than the one for the adversarial methods. For instance, if we consider Energy-Variogram, do not use early stopping and run for 1000 epochs 6 times, we get a total of $ 97.38\times 6000 = 584280 $ seconds. For WGAN-GP, we obtain instead $ 7.00\times49\times1000 = 343000 $ seconds, which is only slightly smaller than the grand total for Energy-Variogram. For the latter, this number does not take into account early stopping which, as can be seen from Table~\ref{tab:comp_cost}, reduces largely the total number of epochs required for training.
	
	Additionally, we highlight how, in the results used for Table~\ref{tab:comp_cost}, the SR methods were trained using 10 simulations from the generator for each observation window (i.e., 10 forecasts). In Appendix~\ref{app:n_outputs}, we studied the effect of the number of simulations used on training, highlighted how the performance is good with as little as 2 or 3 simulations. This greatly reduces the computational cost; we report that in Table~\ref{tab:comp_cost_2}; for this study, the Energy Score was used.

	\begin{table}[tb]
		\begin{center}
			\begin{tabular}{lccc}
				\toprule
				& Per-epoch Computational cost & Early stopping at epoch & Total computational cost \\
				\midrule
				2 & 13.7    &  100   &   1371   \\
				3 &  19.1  &   100  &   1913 \\
				5 & 	29.6	    &  100   &  2967   \\
				10 & 	 54.2 & 100  &  5417 	\\
				20 &	107.0	    &   100  &  10700    \\
				30 	&	 159.2   &  100   &   15916  \\
				50 &   258.7  &   100  &  25865    \\
				\bottomrule
			\end{tabular}
		\end{center}
		\caption{Per-epoch and total computational cost, in seconds, for the Energy Score for different numbers of generator simulations. We also report epoch at which early stopping occurred.}\label{tab:comp_cost_2}
	\end{table}

	\subsubsection{Maps for a chosen date}
	
	We provide figures similar to Fig.~\ref{fig:weatherbench} in the main text in \href{https://github.com/LoryPack/GenerativeNetworksScoringRulesProbabilisticForecasting/blob/main/additional_results.pdf}{this online PDF file}, due to space constraints in the present document. There, we also show deviation of draws from the forecast distribution and the realization from the forecast mean (obtained empirically from 100 draws from the forecast distribution).

	\subsubsection{Time-series plots for selected variables on the grid}\label{app:weatherbench_res_2}

\begin{figure*}[htb]
		\centering
		\begin{subfigure}{0.32\textwidth}
			\centering
			\includegraphics[width=\columnwidth]{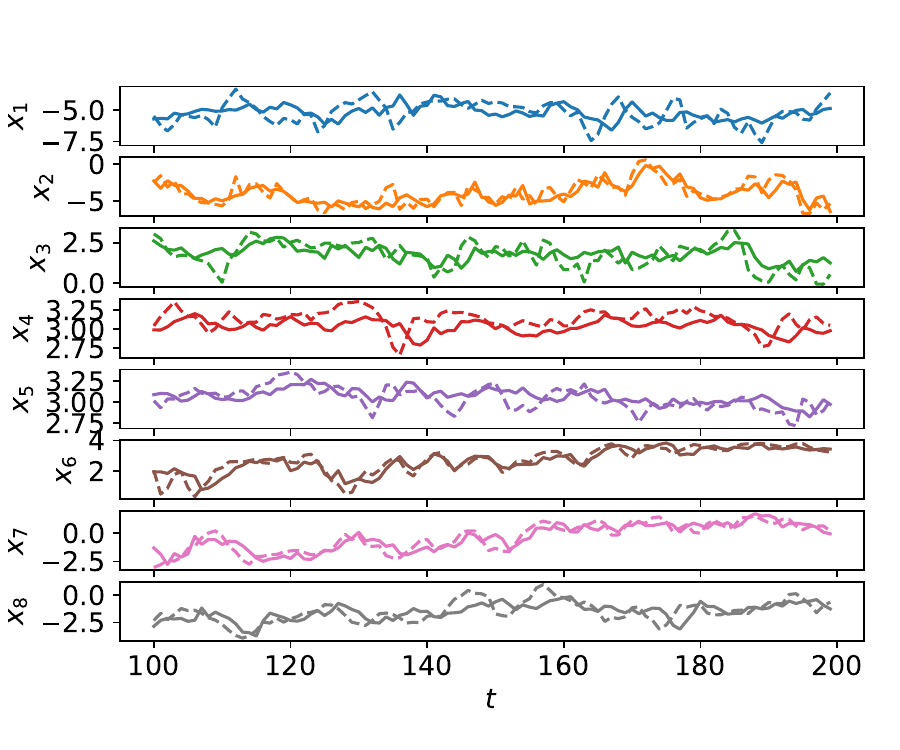}
			\caption{Regression}
		\end{subfigure}~
		\begin{subfigure}{0.32\textwidth}
			\begin{center}
				\includegraphics[width=\textwidth]{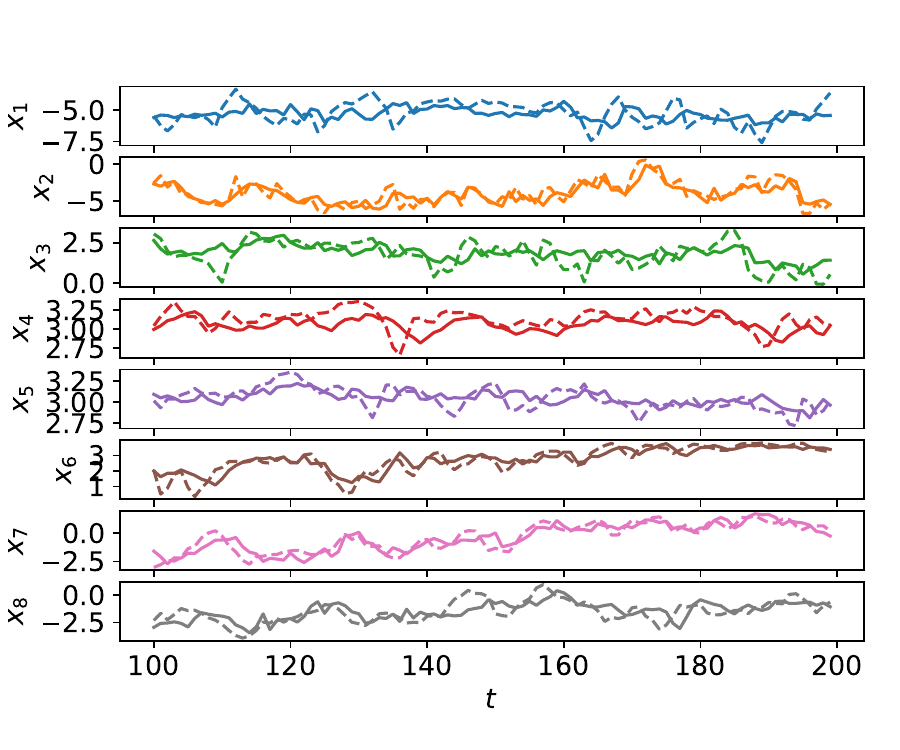}
			\end{center}
			\caption{Patched Regression, 8}
		\end{subfigure}~
		\begin{subfigure}{0.32\textwidth}
			\centering
			\includegraphics[width=\textwidth]{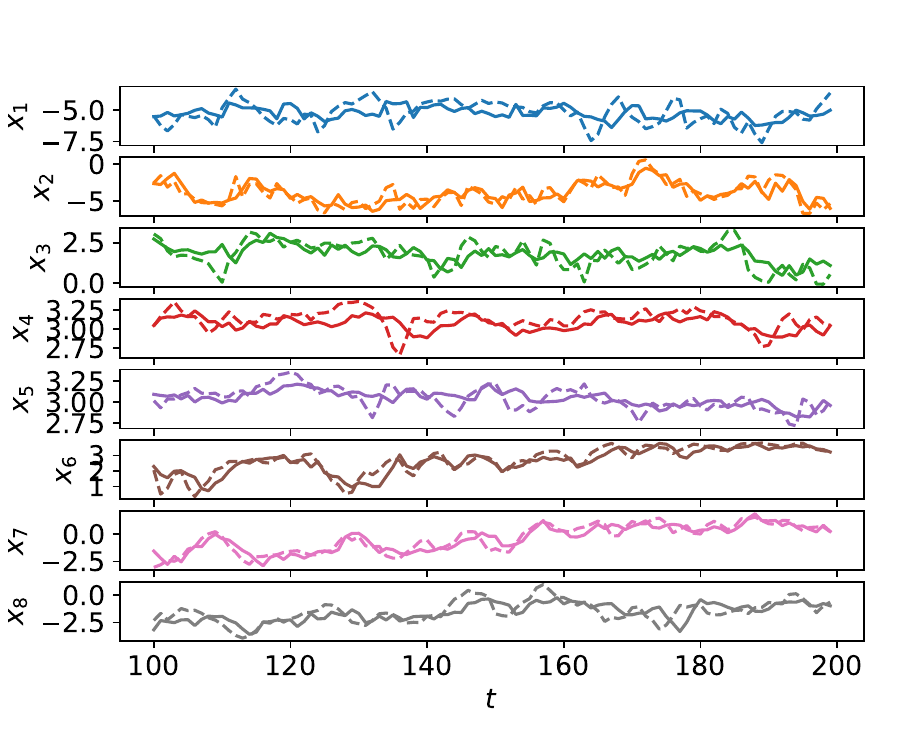}
			\caption{Patched Regression, 16}
		\end{subfigure}
		
		\caption{Results with the Regression and patched regression losses for 8 locations on the WeatherBench grid. The panels show observations (dashed line) and median forecast (solid line)}
		\label{fig:weatherbench_timeseries_det}
	\end{figure*}

\begin{figure*}[htb]
		\centering
		\begin{subfigure}{0.32\textwidth}
			\centering
			\includegraphics[width=\columnwidth]{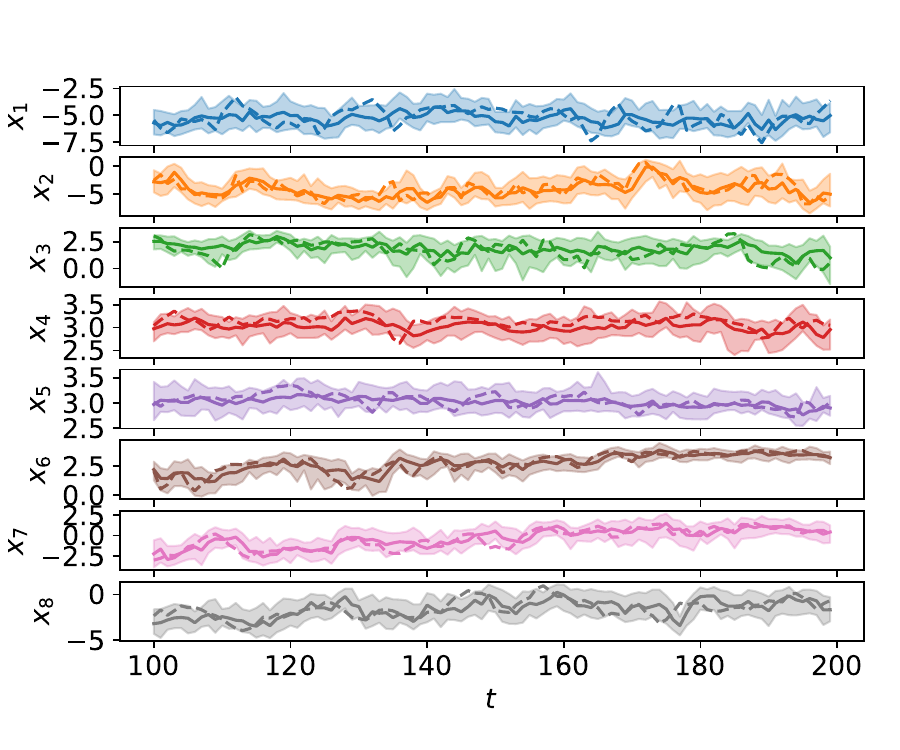}
			\caption{Energy Score.}
		\end{subfigure}~
		\begin{subfigure}{0.32\textwidth}
			\begin{center}
				\includegraphics[width=\textwidth]{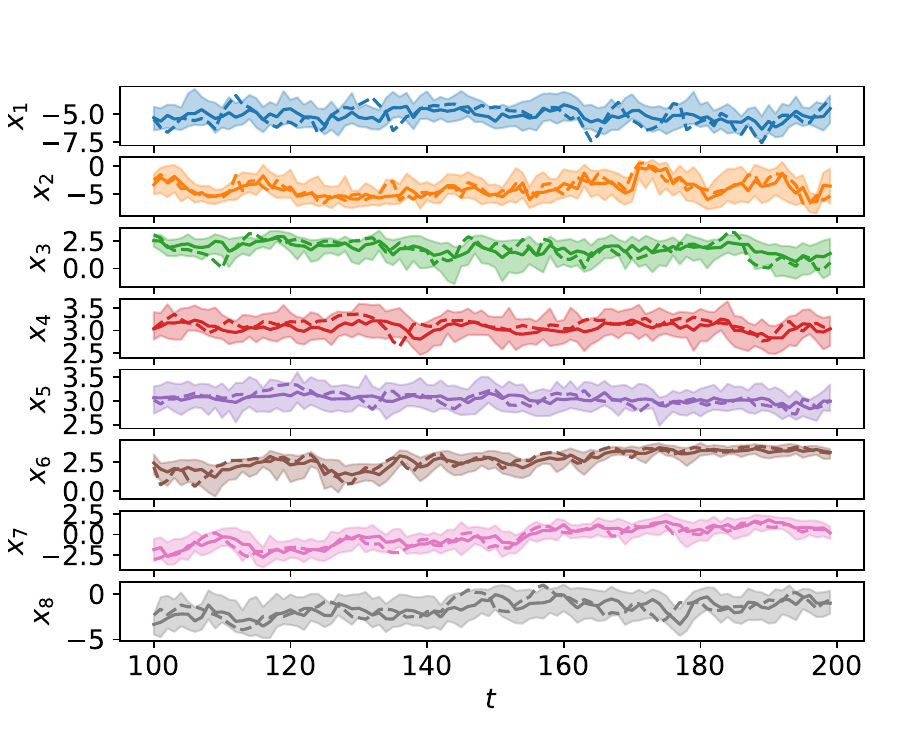}
			\end{center}
			\caption{Kernel Score.}
		\end{subfigure}~
		\begin{subfigure}{0.32\textwidth}
			\centering
			\includegraphics[width=\textwidth]{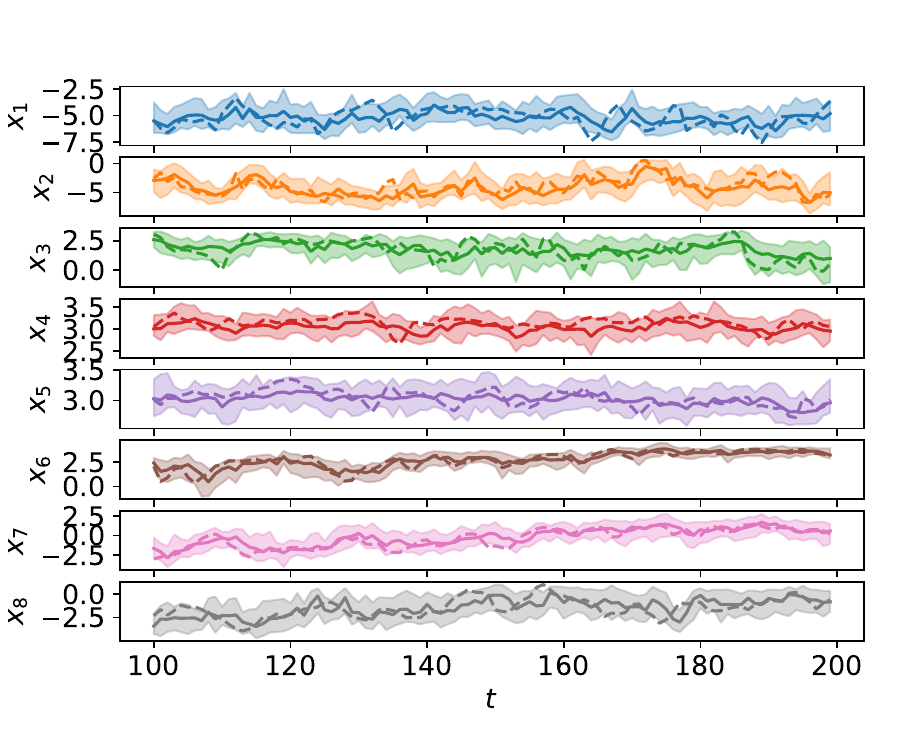}
			\caption{Energy-Kernel.}
		\end{subfigure}
		\caption{Results with the the Energy, Kernel and Energy-Kernel Scores for 8 locations on the WeatherBench grid. The panels show observations (dashed line), median forecast (solid line) and 99\% credible interval (shaded region) for a portion of the test set.}
		\label{fig:weatherbench_timeseries}
	\end{figure*}

\begin{figure*}[htb]
		\centering
		\begin{subfigure}{0.32\textwidth}
			\centering
			\includegraphics[width=\columnwidth]{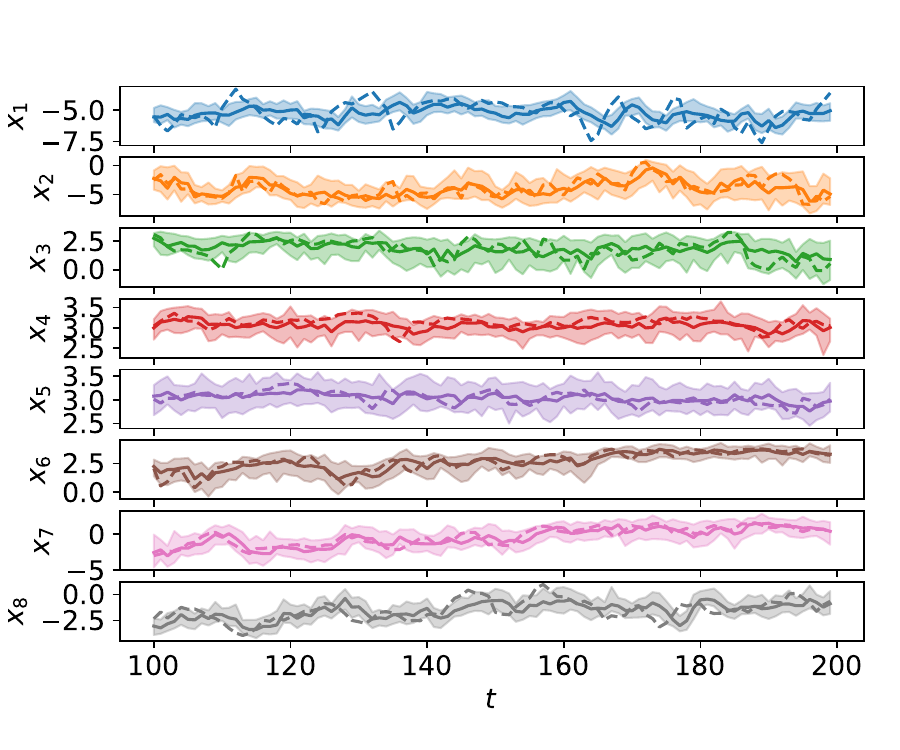}
			\caption{Energy-Variogram Score}
		\end{subfigure}~
		\begin{subfigure}{0.32\textwidth}
			\begin{center}
				\includegraphics[width=\textwidth]{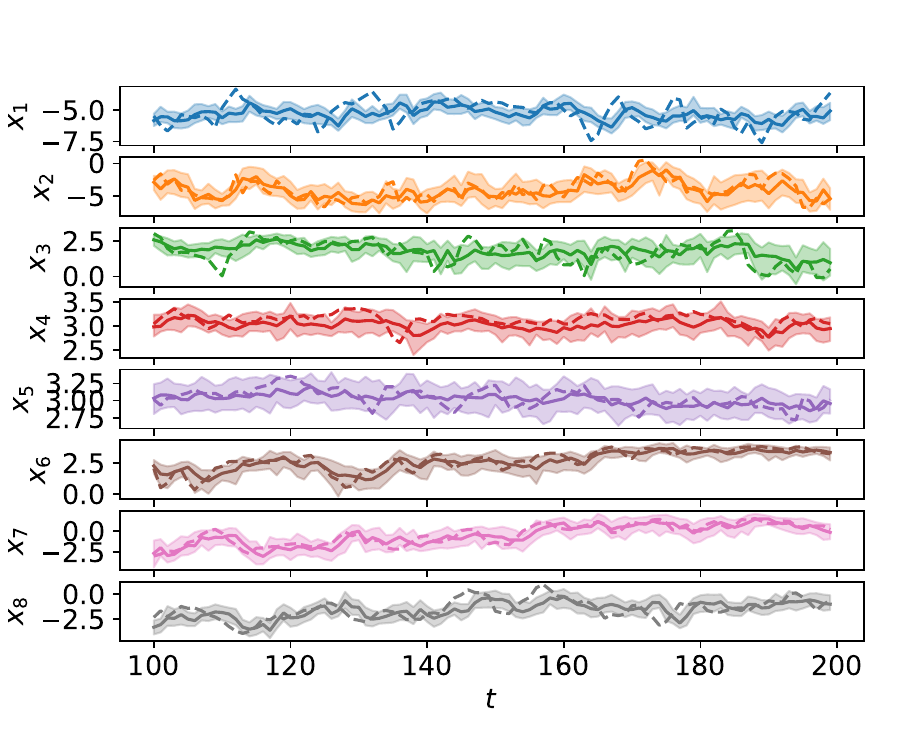}
			\end{center}
			\caption{Kernel-Variogram Score}
		\end{subfigure}\\
		\begin{subfigure}{0.32\textwidth}
			\centering
			\includegraphics[width=\textwidth]{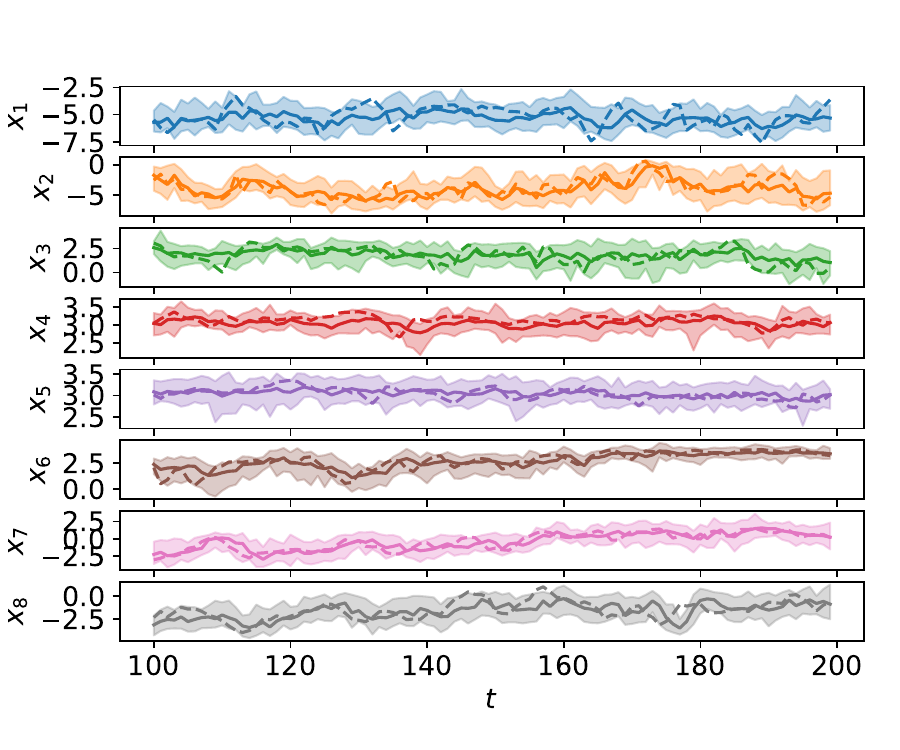}
			\caption{Patched Energy Score (8)}
		\end{subfigure}~
		\begin{subfigure}{0.32\textwidth}
			\centering
			\includegraphics[width=\textwidth]{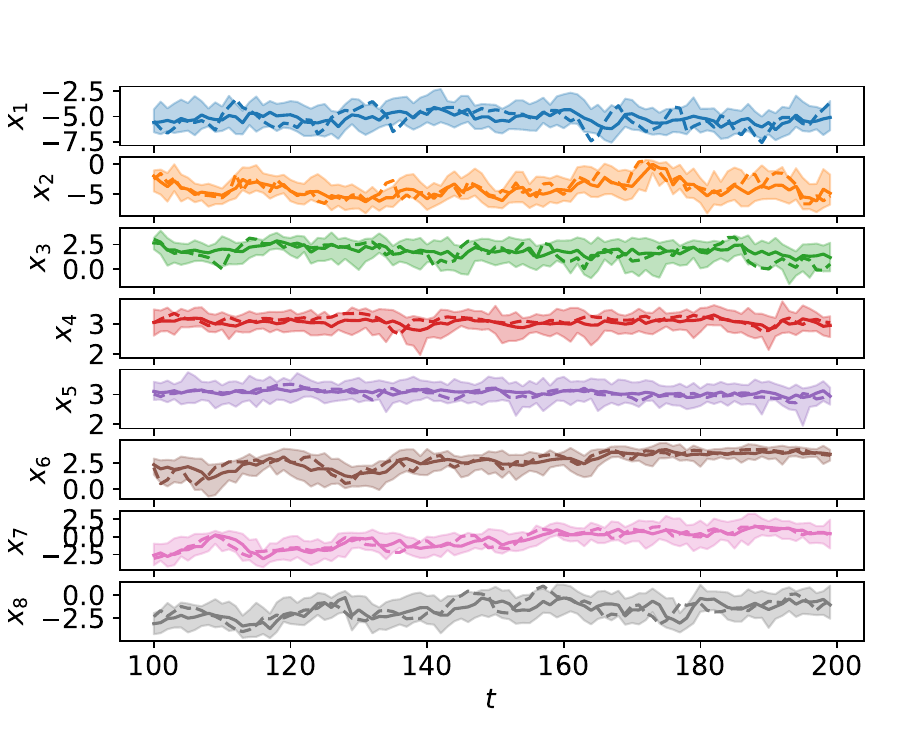}
			\caption{Patched Energy Score (16)}
		\end{subfigure}
		\caption{Results with the Energy-Variogram, Kernel-Variogram and Patched Energy Score (with patch size both 8 and 16) Scores for 8 locations on the WeatherBench grid. The panels show observations (dashed line), median forecast (solid line) and 99\% credible interval (shaded region) for a portion of the test set.}
		\label{fig:weatherbench_timeseries_2}
	\end{figure*}

\begin{figure*}[htb]
		\centering
		\begin{subfigure}{0.32\textwidth}
			\begin{center}
				\includegraphics[width=\textwidth]{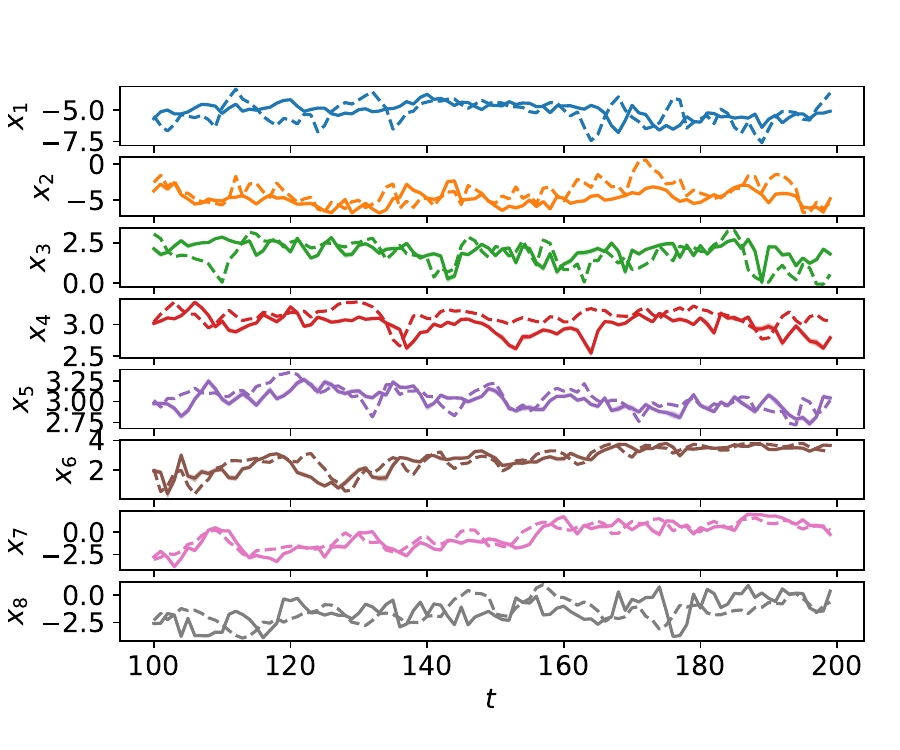}
			\end{center}
			\caption{GAN (1).}
		\end{subfigure}~
		\begin{subfigure}{0.32\textwidth}
			\begin{center}
				\includegraphics[width=\textwidth]{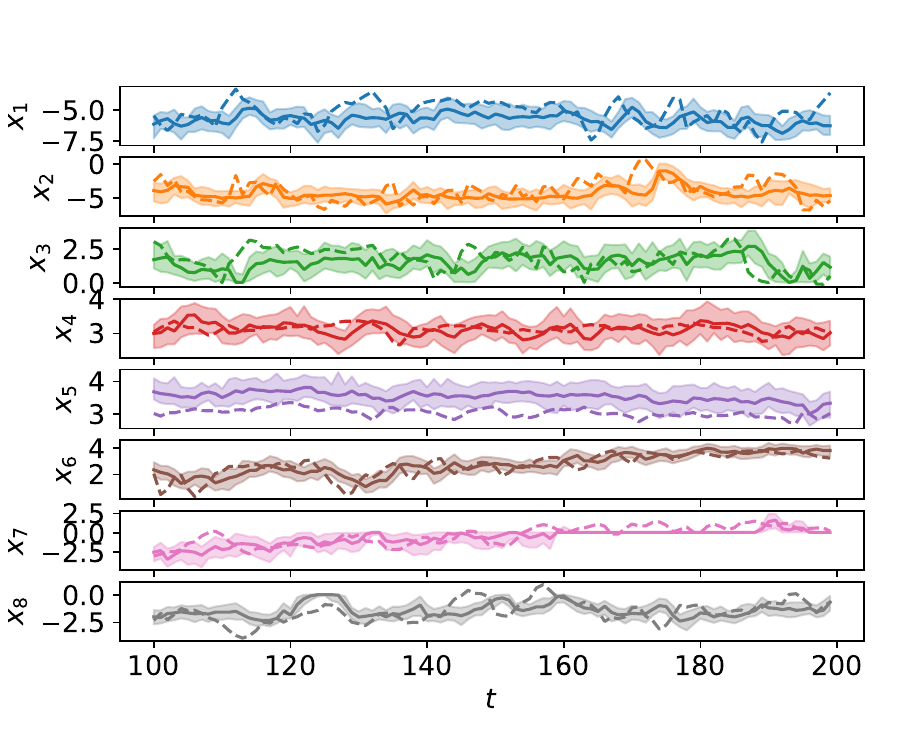}
			\end{center}
			\caption{GAN (2).}
		\end{subfigure}\\
		\begin{subfigure}{0.32\textwidth}
			\centering
			\includegraphics[width=\columnwidth]{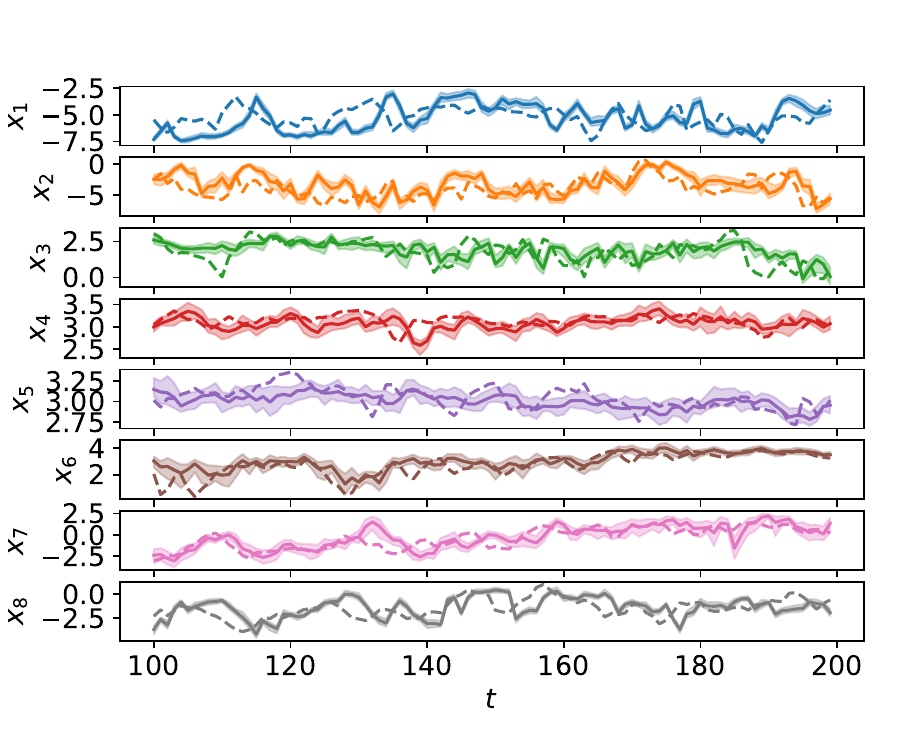}
			\caption{GAN (3).}
		\end{subfigure}~
		\begin{subfigure}{0.32\textwidth}
			\centering
			\includegraphics[width=\textwidth]{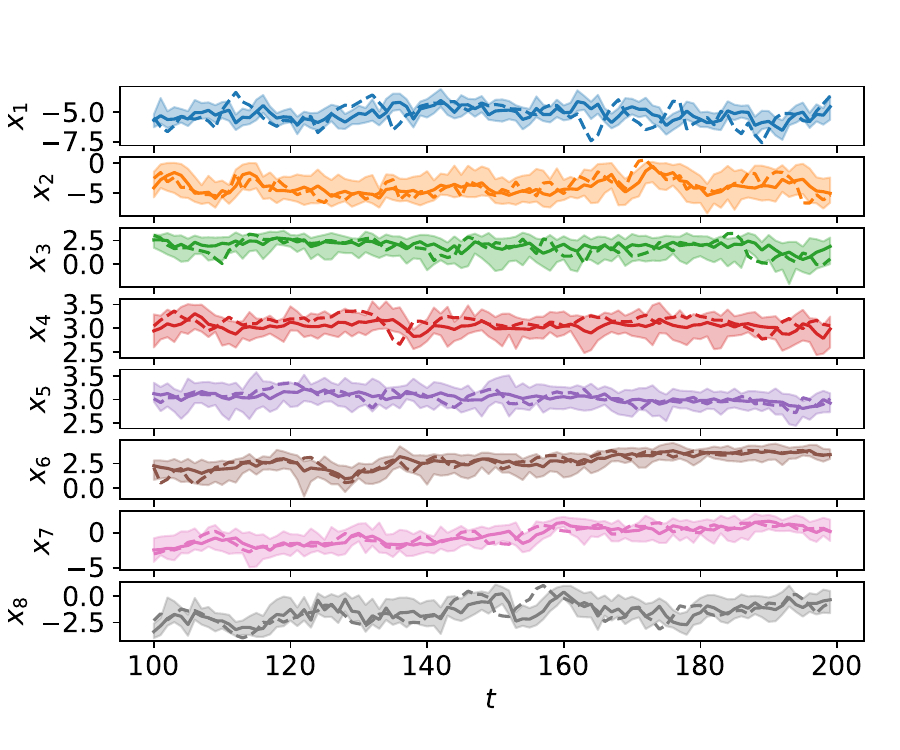}
			\caption{WGAN-GP}
		\end{subfigure}
		\caption{Results with the three considered GAN setups and WGAN-GP Scores for 8 locations on the WeatherBench grid. The panels show observations (dashed line), median forecast (solid line) and 99\% credible interval (shaded region) for a portion of the test set. Notice how the first GAN setup severely underestimates the uncertainty region, while the second one forecasts unpyhsical evolution for some time intervals. }
		\label{fig:weatherbench_timeseries_3}
	\end{figure*}

	In Figures~\ref{fig:weatherbench_timeseries_det}, ~\ref{fig:weatherbench_timeseries}, \ref{fig:weatherbench_timeseries_2} and \ref{fig:weatherbench_timeseries_3}, and show the time series evolution, for a portion of the test period, for 8 randomly selected locations on the WeatherBench grid, for all considered methods (the same locations are shown for all methods). The dashed line represents the true evolution, the solid one the forecast mean, while the shaded region represents 99\% credible intervals.

	\FloatBarrier

	\vskip 0.2in

\end{document}